\documentclass[nonblindrev]{informs3_noinforms}

\usepackage{enumerate}
\usepackage{multirow}

\usepackage{float, bm}

\usepackage[algo2e,linesnumbered,ruled,vlined]{algorithm2e}
\SetKwInput{KwRequire}{Require}
\SetKwInput{KwEnsure}{Ensure}

\DoubleSpacedXI

\usepackage{endnotes}

\newcommand{\RD}{\texttt{DualReflect}}
\newcommand{\PNL}{\mathsf{P}}
\newcommand{\Model}{\mathsf{M}}
\newcommand{\Code}{\mathsf{C}}
\newcommand{\Scenario}{\mathsf{SCE}}
\newcommand{\CoT}{\mathsf{CoT}}
\newcommand{\ModelF}{\mathsf{MF}}
\newcommand{\Data}{\mathsf{PD}}
\newcommand{\DPLM}{\texttt{DPLM}}
\newcommand{\DPBENCH}{\texttt{DP-Bench}}

\renewcommand{\theequation}{\thesection.\arabic{equation}}

\usepackage{natbib}
\bibpunct[,
]{(}{)}{,}{a}{}{,}
\def\bibfont{\small}

\TheoremsNumberedThrough

\ECRepeatTheorems

\EquationsNumberedThrough

\usepackage[most]{tcolorbox}
\newtcolorbox[list
inside=prompt,auto
counter,number
within=section]{prompt}[1][]{ colbacktitle=black!60, coltitle=white, fonttitle=\scriptsize,fontupper=\tiny, boxsep=5pt, enhanced, left=0pt, right=0pt, top=0pt, bottom=0pt, boxrule=1pt, breakable, #1 }

\usepackage{algorithm}

\usepackage{booktabs}
\usepackage{arydshln}
\usepackage{makecell}

\newtcolorbox{promptbox}[1]{
  enhanced,
  breakable,
  colback=blue!3!white,
  colframe=blue!35!black,
  colbacktitle=blue!40!black,
  title=\textbf{#1},
  fonttitle=\bfseries\footnotesize,
  fontupper=\scriptsize,
  sharp corners,
  boxrule=0.8pt,
  left=5pt, right=5pt, top=5pt, bottom=5pt,
  parbox=false
}

\usepackage{fvextra}
\DefineVerbatimEnvironment{dplmoutput}{Verbatim}{
    fontsize=\scriptsize,
    frame=single,
    breaklines=true,
    breaksymbolleft={},
    xleftmargin=2pt,
    xrightmargin=2pt
}

\begin{document}

\RUNTITLE{Auto-Formulating Dynamic Programming Problems with Large Language Models}
\TITLE{Auto-Formulating Dynamic Programming Problems with Large Language Models}

\ARTICLEAUTHORS{
\AUTHOR{\bf Chenyu Zhou\thanks{Co-first authors with equal contribution.}}
\AFF{Antai College of Economics and Management, Shanghai Jiao Tong University, Shanghai, China, chenyuzhou@sjtu.edu.cn}
\AUTHOR{\bf Jingyuan Yang\footnotemark[1]}
\AFF{Booth School of Business, University of Chicago, Chicago, IL, jyang21@chicagobooth.edu}
\AUTHOR{\bf Linwei Xin\thanks{Corresponding author.}}
\AFF{School of Operations Research and Information Engineering, Cornell University, Ithaca, NY, lx267@cornell.edu}
\AUTHOR{\bf Yitian Chen}
\AFF{Cardinal Operations, Beijing, China, chenyitian@shanshu.ai}
\AUTHOR{\bf Ziyan He}
\AFF{Shanghai University of Finance and Economics, Shanghai, China, heziyan@stu.sufe.edu.cn}
\AUTHOR{\bf Dongdong Ge\footnotemark[2]}
\AFF{Antai College of Economics and Management, Shanghai Jiao Tong University, Shanghai, China, ddge@sjtu.edu.cn}

\RUNAUTHOR{Zhou et al.} }

\ABSTRACT{Dynamic programming (DP) is a fundamental method in operations research. Formulating DP models has traditionally required expert knowledge of both the problem context and DP techniques, and Large Language Models (LLMs) offer the potential to automate this process. However, DP problems pose unique challenges due to their inherently stochastic transitions and the limited availability of training data. These factors make it difficult to directly apply existing LLM-based models or frameworks developed for other optimization problems, such as linear or integer programming.
We introduce $\DPBENCH$, the first benchmark covering a wide range of textbook-level DP problems to enable systematic evaluation. We present Dynamic Programming Language Model ($\DPLM$), a 7B-parameter specialized model that achieves performance comparable to state-of-the-art LLMs like DeepSeek-R1, and surpasses them on hard problems. Central to $\DPLM$'s effectiveness is $\RD$, our novel synthetic data generation pipeline, designed to scale up training data from a limited set of initial examples. $\RD$ combines forward generation for diversity and backward generation for reliability. Our results reveal a new insight: backward generation is favored in low-data regimes for its strong correctness guarantees, while forward generation, though lacking such guarantees, becomes increasingly valuable at scale for introducing diverse formulations. This trade-off highlights the complementary strengths of both approaches and the importance of combining them.}

\HISTORY{\today{}}

\maketitle

\section{Introduction}
Automating the formulation of decision-making problems represents a major step toward fully autonomous decision-support systems. Traditionally, solving such problems involves two sequential stages: first, translating real-world scenarios into well-defined mathematical models--an essential skill emphasized in operations research education--and second, applying computational tools to find optimal or near-optimal solutions. While substantial research in recent decades has primarily focused on the second stage (i.e., enhancing algorithms and improving solver efficiency), the resulting advancements span a wide range, from foundational developments such as reinforcement learning (RL) frameworks (e.g., \citealp{sutton2018reinforcement}) and approximate dynamic programming techniques (e.g., \citealp{powell2007approximate}), to powerful solvers like COPT, CPLEX, and Gurobi. Such innovations coupled with increasing computational power have led to high-impact real-world applications, exemplified by AlphaGo, which leveraged deep learning and RL to solve complex, large-scale decision-making problems (\citealp{silver2016mastering}). That said, while these advancements have shifted many computational tasks to automated software, the initial problem formulation step has largely remained manual and dependent on expert knowledge.

The recent rapid progress in large language models (LLMs) provides a promising opportunity to automate this crucial first step. LLMs excel in natural language processing and have demonstrated significant potential for effectively automating the formulation of mathematical models directly from plain English descriptions. Leveraging LLMs can substantially reduce the human expertise required, simplify the problem formulation process, and make advanced optimization methods accessible to a broader audience.

Among various optimization problems, dynamic programming (DP) represents a particularly important yet challenging category for formulation automation. Despite recent advances in leveraging LLMs to auto-formulate optimization problems such as LP and MIP, (e.g., \citealp{huang2024orlm,lu2025optmath}), directly applying existing techniques to DP problems proves insufficient for several reasons. First, DP addresses multi-period decision-making with sequential transitions, often requiring careful reasoning about the order of events, such as what occurs first, which costs are incurred immediately, and how the system evolves over time. This complexity is further amplified when transitions are stochastic, which is often the primary challenge even for human modelers. In contrast, LP and MIP problems are deterministic and rarely involve such intricate transitions, focusing instead on constraint formulation.

Second, beyond the technical complexity of modeling DPs, which even human experts may struggle with, LLMs face an additional challenge that is especially salient in the OR/OM domain: accurately interpreting natural-language problem descriptions. Unlike typical DP problems studied in computer science, which are formally structured, OR/OM problems are often embedded in text-rich, business-oriented scenarios that require understanding implicit assumptions, domain-specific terminology, and contextual nuances. Unlike domain experts, who naturally infer implicit constraints and contextual assumptions from subtle cues in wording or context, LLMs often struggle to recognize these subtleties. This limitation makes DP tasks particularly challenging to automate using LLMs.
Let us illustrate this point through an example.

\begin{example}\label{eg-2}
\textit{A warehouse has an end-of-period capacity of 3 units. During a production period, a \$4 setup cost is incurred, and a holding cost of \$1 per unit applies to the period's ending inventory. Variable production costs are \$1 per unit. Demand each period is equally likely to be either 1 or 2 units, and all demand must be fulfilled immediately. The discount factor is $\beta = 0.8$. The objective is to minimize the expected discounted costs over an infinite horizon, starting with zero initial inventory.}
\end{example}

Example \ref{eg-2} represents a toy inventory problem with random demand and a subtle distinction: the term ``end-of-period capacity of 3 units" differs from simply having a ``capacity" of 3 units. Since demand is at least one unit per period, the effective order-up-to level at the start of each period can be 4 units. This nuance is often missed. Most state-of-the-art (SOTA) LLMs (e.g., o4-mini-high, o3, DeepSeek-R1, and GPT-4o) fail to distinguish between the two notions, often leading to incorrect answers.
While this may initially appear to be a simple word-understanding issue, which could also arise in LP or other deterministic optimization problems, we find that the challenge runs deeper. When we modify the same problem to use deterministic demand (always 2 units per period), strong models like o4-mini-high, o3, and DeepSeek-R1 are able to correctly interpret the meaning of ``end-of-period'' capacity. This suggests that the misunderstanding arises not purely from language, but rather from the added complexity introduced by stochastic transitions. It is the combination of subtle language cues and the need to reason over stochastic dynamics that makes DP problems especially difficult for LLMs.

Beyond the already significant challenges LLMs face in automating DP formulations, an even greater obstacle is the lack of a standardized test set for evaluating their performance on DP problems, making it difficult to analyze or compare models in a consistent and rigorous manner. To fill this gap, we introduce $\DPBENCH$, the very first standardized benchmark\footnote{The term benchmark, in the context of datasets, refers to standardized and widely recognized collections used to evaluate and compare the performance of machine learning models and algorithms.} dataset tailored to DP problems.

DP-Bench consists of 132 carefully selected and augmented textbook-level DP problems, each featuring a closed-ended, specific numerical answer, similar in spirit to benchmarks such as the celebrated Humanity's Last Exam by Scale AI (\citealp{phan2025humanity}). Proof-based problems are explicitly excluded. The goal is to assess LLMs’ ability to translate problem descriptions from plain English into mathematical models.
While we evaluate this end-to-end (i.e., having the LLM both formulate and solve each problem via code), the final answer serves as a proxy for formulation accuracy. This is because, for textbook-level problems, a correct solution typically follows directly from an accurate formulation using standard textbook methods. Occasional coding errors may occur, but these are less indicative of core understanding than errors in problem formulation.

Testing SOTA reasoning LLMs on $\DPBENCH$ reveals significant performance gaps. The best-performing model, DeepSeek-R1, achieves only 59.1\% average accuracy (see Table \ref{tab:main_results_sub} in Section \ref{Benchmark}), indicating substantial room for improvement. In contrast, LLMs perform considerably better on benchmarks for other types of optimization problems. For instance, \cite{lu2025optmath} report that DeepSeek-V3 achieves 95.9\% accuracy on NL4OPT, a popular benchmark in operations research featuring textbook-level LP problems, created by \cite{ramamonjison2023nl4opt}.
We also evaluate ORLM (\citealp{huang2024orlm}), an LLM specifically designed for auto-formulating generic optimization problems. On $\DPBENCH$, ORLM achieves only 0.8\% accuracy; even when allowing pass@10, i.e., measuring whether at least one of 10 sampled outputs yields a correct answer, its accuracy remains as low as 8.3\%. These results highlight the inherent challenges LLMs face in auto-formulating DP problems compared to other optimization problems, and the need for developing more capable models tailored specifically for DP problems.

In this paper, we aim to improve the ability of open-source LLMs to auto-formulate DP problems. There are three common approaches to leveraging existing commercial or open-source LLMs to enhance task performance: (1) inference-time enhancement techniques, including prompt engineering, retrieval-augmented generation (RAG), and agentic workflows, (2) commercial API-based fine-tuning, and (3) fine-tuning open-source models. The first two approaches come with notable limitations. Both require sending potentially sensitive data to third-party servers, raising privacy and security concerns. Moreover, inference-time methods do not change the base model's parameters, so performance remains constrained by the model’s existing capabilities. For example, GPT-4o, with a trivial prompt, achieves only 47.8\% accuracy on the easy problems and 19.0\% on the hard problems of $\DPBENCH$. To improve performance, we develop a carefully designed LLM-based agentic workflow that integrates step-by-step reasoning and label-based RAG.
With these enhancements, the agentic workflow achieves 58.9\% on the easy problems and 38.1\% on the hard problems (see Table \ref{tab:rag} below for details). Although this represents a meaningful improvement, gains begin to plateau, suggesting that inference-time techniques alone remain fundamentally limited by the capabilities of the underlying model.
Commercial API-based fine-tuning faces similar constraints. It offers limited customization, and the fine-tuned model is not owned by the user, resulting in vendor lock-in. This introduces operational risks, such as dependency on service availability, pricing changes, or platform discontinuation. These limitations also reduce deployment flexibility, especially for low-latency or offline use cases.

Given these limitations, we choose to fine-tune open-source models for the specialized task of auto-formulating DP problems. To ensure feasibility under limited computational resources, we focus on small-scale models. A key challenge is the lack of suitable training data. While many DP-related problems exist on platforms like LeetCode, they are typically coding-style puzzles and do not reflect the background-rich, natural-language problem descriptions common in operations research. To address this, we build a complete training pipeline from scratch. We begin with a curated set of high-quality seed problems, then introduce a dual synthetic data generation framework to ensure both diversity and fidelity, and finally apply efficient fine-tuning recipes. As a result, our proposed model, $\DPLM$, achieves performance comparable to SOTA LLMs that are over 100 times larger.

Moreover, inference-time techniques and fine-tuning should not be viewed as competing strategies, but as complementary ones. When combined with simple inference-time aggregation techniques such as majority voting, $\DPLM$’s average accuracy further increases to 71.2\% (see Table \ref{tab:teacher_guidance} below), representing a gain of nearly 10 percentage points. This suggests that inference-time enhancement becomes even more effective when built on a stronger, task-specific fine-tuned model.

\subsection{Our Contributions}
This work is the first to systematically investigate the automatic formulation of DP problems using LLMs. Our contributions are threefold.

First, to support the evaluation of LLMs on DP tasks, we introduce $\DPBENCH$, a curated benchmark of 132 textbook-level DP problems. These problems span diverse domains and cover a variety of problem types, including deterministic and stochastic settings, as well as finite- and infinite-horizon formulations. Each problem is classified as either “easy” or “hard”, corresponding to the expected difficulty level for undergraduate and graduate students, respectively. Designed to produce specific numerical answers, this benchmark offers a standardized and rigorous framework for evaluating LLMs’ ability to auto-formulate DP problems from natural language descriptions.

Second, we present $\DPLM$, a specialized model fine-tuned on open-source architectures and trained entirely from scratch using synthetic data distilled from GPT-4o. Our results show that $\DPLM$ significantly outperforms its teacher model, GPT-4o, on both the easy and hard subsets of $\DPBENCH$. Remarkably, despite being only 7B in size (which is nearly 100x smaller than DeepSeek-R1), $\DPLM$ achieves the second-highest accuracy on easy problems and the highest accuracy on hard problems among all evaluated models, including advanced reasoning models such as o1 and DeepSeek-R1. These results highlight both the potential of LLMs for automating DP formulation and the promise of compact, domain-specific models tailored to specialized problem classes like DP.

Third, we introduce $\RD$, a scalable and principled data generation framework that starts from a limited set of 91 seed examples, and produces a synthetic dataset of 113K problem instances, strategically balancing accuracy and diversity. $\RD$ combines forward generation, which creates problems first and then solves them, with backward generation, which begins from a solution and constructs the corresponding problem. As part of backward generation, we design a reflect-and-refine approach to recover 20.8\% of examples that would otherwise be discarded due to solution inconsistencies. In doing so, $\RD$ mitigates a common limitation in synthetic datasets, which tend to contain mostly accurate but overly simplistic problems. Our empirical findings reveals a new insight: backward generation is preferred when only a small number of examples can be generated, due to its high reliability, while forward generation becomes increasingly valuable at scale for its diversity, highlighting the necessity of combining both.

\subsection{Literature Review}
Our work closely aligns with research that leverages LLMs to translate natural language descriptions into optimization models. Early efforts by \cite{ramamonjison2023nl4opt} introduce the widely used NL4OPT benchmark for LP problems. More recent work include LLM-agent prompting approaches \citep{ahmaditeshnizi2023optimus, bertsimas2024robust, liang_llm_2025} and fine-tuning methods \citep{yang2024optibench, huang2024orlm, lu2025optmath}, which primarily improve LLM performance on deterministic LP, MILP, NLP, and robust optimization.
DP, especially when involving stochastic transitions, remains unexplored.

Among existing works, the closest is \cite{huang2024orlm}. Beyond the fact that auto-formulating DP is fundamentally different from formulating LP, several key differences distinguish our approach. First, ORLM relies on proprietary industrial datasets, whereas we propose a more generalizable framework for emerging domains by systematically curating seed data and constructing benchmarks from textbooks. Second, while ORLM uses only forward generation to expand its dataset, we combine both forward and backward generation to balance diversity and correctness. Finally, ORLM trains its models using supervised fine-tuning (SFT) alone, whereas we further apply RL to improve solution quality and robustness.

Preference-based RL alignment has recently emerged as a popular approach for enhancing reasoning-oriented LLMs on complex tasks. Flagship models such as OpenAI’s o1 rely on variants of reinforcement learning from human feedback (RLHF) to improve factuality and performance. Representative techniques include on-policy RLHF \citep{ouyang2022instructgpt}, offline Direct Preference Optimization (DPO) \citep{rafailov2023dpo}, and the lightweight Group-Relative Policy Optimization (GRPO) \citep{shao2024deepseekmath}. Although the training report of DeepSeek-R1-Zero \citep{guo2025deepseek} highlights the potential of RL-only approaches, relying exclusively on RL without prior SFT often leads to training instability and prohibitive computational costs, especially for smaller models \citep{zheng2023secretsofrlhf}. In this paper, we adopt a two-step approach that combines SFT with RL. Our experiments further demonstrate the importance of SFT as a crucial preliminary step for establishing a strong baseline prior to applying RL.

To address the scarcity of training data in DP auto-formulation, our research aligns with the literature on generating synthetic data to alleviate the lack of human-labeled examples. A general approach is called Self-Instruct, introduced by \cite{wang2022self}. The core idea is to prompt an LLM with a small set of seed examples, guiding it to generate its own instructions and corresponding outputs with minimal human effort. This approach follows the forward generation paradigm, where an instruction or problem is created first, followed by a corresponding solution. However, in more reasoning-intensive domains such as mathematical problem solving, deriving solutions from arbitrary problems can be highly nontrivial, limiting the accuracy of purely forward-generated data. To address this, some studies adopt backward generation, where the process begins from a known solution and constructs a compatible problem around it. A notable use of the combination of forward and backward generation is by \cite{lample_deep_2019}, who generate diverse mathematical functions along with their integrals or derivatives. Subsequently, \cite{alfarano_global_2024} apply this technique to discover Lyapunov functions that ensure the global stability of dynamical systems, where algorithmic solvers exist for certain families of polynomial systems. In both cases, the solutions derived from both forward and backward generation are verifiable. In sharp contrast, in our context, there are no tools capable of deriving or verifying optimal solutions to decision-making problems from the description alone via forward generation. As a result, existing works (e.g., \citealp{yang2024optibench, lu2025optmath}) primarily focus on backward generation to guarantee accuracy. In contrast, we highlight that while forward generation is inherently less accurate, it significantly increases diversity and is especially valuable when generating larger training sets.

To enhance reasoning capabilities without extensive manual annotation, various techniques have been proposed. \cite{wei2023chainofthoughtpromptingelicitsreasoning} demonstrate that providing step-by-step chain-of-thought (CoT) reasoning examples substantially improves performance on math and logic problems. Building upon this, \citet{zelikman2022star} propose the Self-Taught Reasoner (STaR), which iteratively prompts an LLM to generate answer-conditioned rationales and fine-tunes it using only those verified as correct. We adopt a similar idea through Reflected CoT but differ from STaR in two key ways. First, while STaR focuses on problems where answers are informative for reasoning, DP answers (e.g., optimal values) provide little guidance for reconstructing the full model. Second, STaR assumes access to a large-scale dataset with correct problems and answers but missing rationales, whereas our backward-generated problems themselves may contain flaws. Providing the correct solution and prompting the LLM to follow the STaR approach may lead it to fabricate a rationale that aligns with the answer but is logically invalid. In some cases, the generated problems themselves may also be flawed. In contrast, Reflected CoT allows the LLM to solve the problem independently, compare its result to the ground truth, and iteratively refine its reasoning over multiple rounds.

LLMs have recently played an expanding role in OR/OM and related decision-making domains. One stream of work leverages LLMs as intelligent agents to support or augment human decision-making. For example, \citet{wang_large_2024} propose a novel statistical data augmentation method that efficiently integrates LLM-generated data with real data in conjoint analysis. \cite{Yin2026} study the optimal proportion of LLM-generated synthetic data to use in market research and develop a principled framework for deriving a closed-form formula that allows heterogeneous treatment across population segments (e.g., younger versus older respondents).
\citet{ye2025lola} leverage LLMs to cold-start online learning algorithms. \citet{bray_tutorial_2025} demonstrates that LLMs can serve as personalized tutors or teaching assistants in data analytics education. A second line of research investigates bias in LLM-generated decisions and data. For instance, \citet{chen_manager_2025} show that ChatGPT can replicate well-known human decision biases, such as loss aversion. \citet{li2025llmgeneratedpersonapromise} find that LLM-generated personas exhibit systematic demographic and ideological biases, introducing distortion even before downstream tasks such as election forecasting or opinion polling. The third direction applies operations research methods to optimize LLM inference efficiency. For instance, \citet{li2025throughputoptimal} develop the queuing fundamentals for LLM inference and prove that a large class of work-conserving scheduling algorithms can achieve maximum throughput for individual inference LLM engine. \citet{ao2025optimizingllminferencefluidguided} model LLM inference as a multi-stage online scheduling problem with Key-Value (KV) cache constraints and establish asymptotic optimality under heavy traffic. \citet{jaillet2025onlineschedulingllminference} also incorporate KV cache constraints to minimize latency and propose a batching and scheduling algorithm with performance guarantees.

\section{Problem Setup}\label{sec-background}

Discrete-time stochastic DP provides a general framework for modeling sequential decision-making under uncertainty over time. A decision-maker seeks to influence the evolution of a probabilistic system over time by selecting actions that optimize a long-term performance criterion. Since each choice impacts not only immediate outcomes but also future states, optimal decisions must anticipate downstream consequences rather than focusing only on immediate rewards. Formally, a DP problem can be mathematically represented as
\begin{equation}\label{eq-MF}
    \ModelF=\left\{T, \mathcal{S}, \mathcal{A}, p_t(\cdot\mid \boldsymbol{s},\boldsymbol{a}), r_t(\boldsymbol{s},\boldsymbol{a}),\gamma\right\},
\end{equation}
which specifies the planning horizon $T$, state and action spaces $\mathcal{S}$ and $\mathcal{A}$, transition probabilities $p_t$, reward functions $r_t$, and discount factor $\gamma$.
Deterministic DP problems arise as a special case in which the transition probabilities degenerate to point masses, yielding a uniquely determined next state for each state–action pair.

To formalize the modeling process of a multi-period decision-making problem, we introduce several key concepts. Let $\PNL$ denote the \emph{problem} natural language description, which describes the decision-making context, objectives, constraints, and sources of uncertainty in plain language. It also contains the central question posed by the decision-maker, such as: ``What is the minimum expected discounted cost over the next 12 months starting with a state 2 machine?'' Given a $\PNL$, we expect an LLM to auto-formulate the mathematical model underlying the problem and then develop an executable code. The formal mathematical representation of the problem is called a \emph{model}, denoted by $\Model$. This model explicitly defines state space, action space, system transitions, reward functions, and boundary conditions. We denote by $\Code$ the corresponding \emph{code implementation}, typically written in Python, that translates the model $\Model$ into a computational solution. The code numerically solves for the optimal value function and policy using standard DP algorithms such as backward induction, value iteration, or policy iteration. The numerical answer computed by $\Code$ is denoted by $y$, and directly addresses the decision question posed in $\PNL$.

In addition to this core pipeline, we also prompt the LLM to generate a chain-of-thought reasoning process, denoted by $\CoT$, as an intermediate step to enhance model formulation and code generation. Chain-of-thought, proposed by \cite{wei2023chainofthoughtpromptingelicitsreasoning}, has been shown to significantly improve LLMs’ ability to perform complex reasoning tasks by encouraging step-by-step intermediate reasoning. Inspired by this technique, we incorporate $\CoT$ generation before solving the problem to help the LLM better structure its understanding and improve solution accuracy.
In summary, we define a complete \emph{solution} to a problem $\PNL$ as the triplet $r = (\CoT, \Model, \Code)$. We use $r^*$ and $y^*$ specifically to denote the solution and answer we consider correct.

It is important to distinguish $\Model$ from $\ModelF$, as defined in \eqref{eq-MF}, which represents the abstract DP structure independent of any specific problem data. In contrast, $\Model$ instantiates $\ModelF$ with concrete problem data $\Data$ extracted from the $\PNL$. Thus, $\Model$ can be viewed as the combination of the abstract model structure $\ModelF$ and the specific data $\Data$ that together fully define the decision problem.

\section{Benchmark}\label{Benchmark}
To systematically evaluate the capability of LLMs to auto-formulate DP problems, we introduce $\DPBENCH$, the first benchmark dedicated to DP. It consists of 90 easy and 42 hard textbook-level problems, all derived and augmented from standard textbooks. The easy problems are primarily adapted from deterministic and probabilistic DP exercises in \cite{winston2004operations}, while the hard problems are drawn from more advanced exercises in \cite{puterman2005markov}. We aim to preserve the original problems from the textbooks while making some necessary modifications to ensure clarity and consistency. We omit the adjustment details here due to space constraints. Each problem in the benchmark includes a ``Question'' and an ``Answer'' obtained by executing the solution. The solutions for the easy benchmark problems are generally available online, while the solutions to the hard problems are derived and verified by ourselves.

\begin{figure}
	\begin{minipage}[t]{0.48\linewidth}
		\centering
		\includegraphics[width=\textwidth]{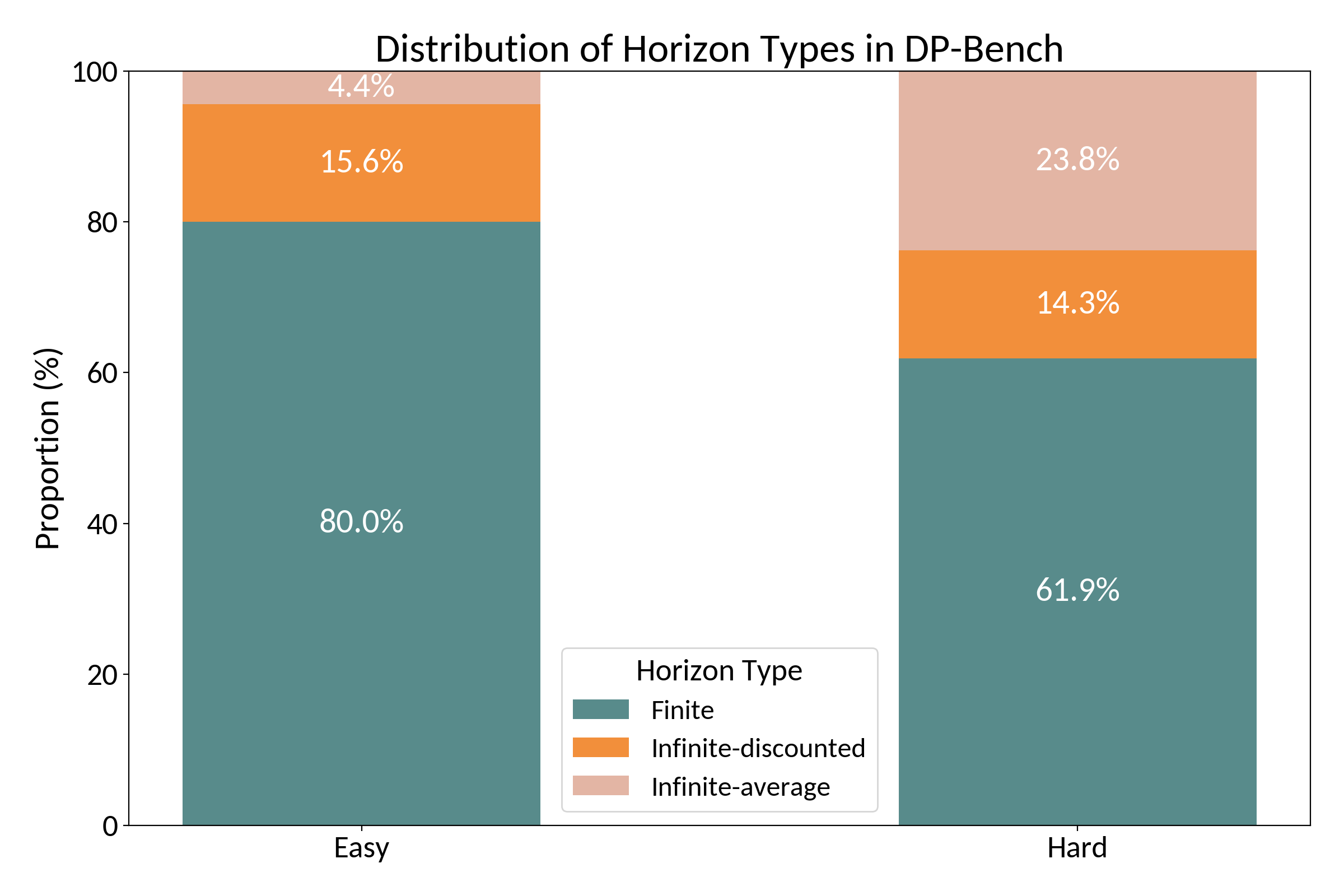}
	\end{minipage}
	\begin{minipage}[t]{0.48\linewidth}
		\centering
		\includegraphics[width=\textwidth]{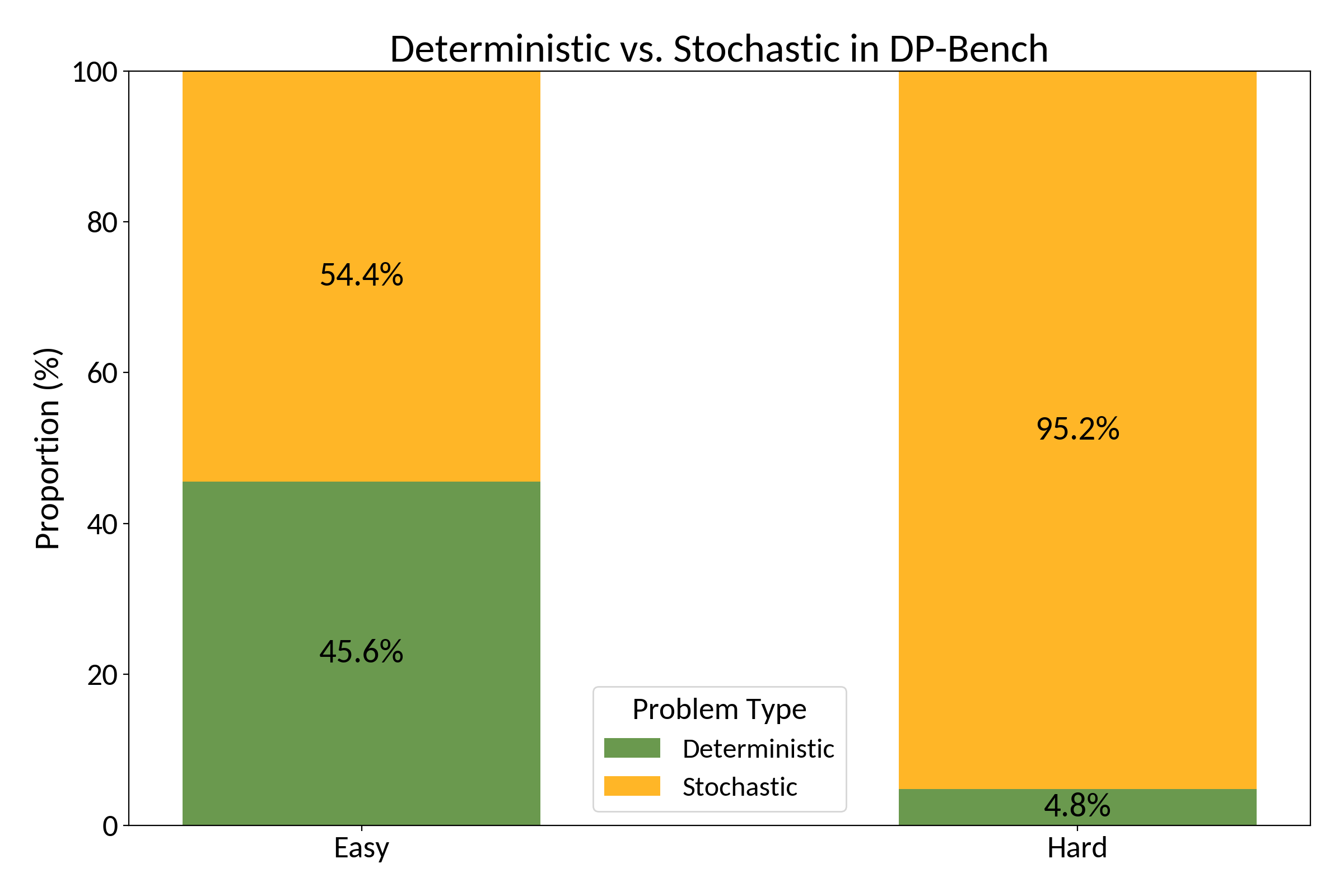}
	\end{minipage}
	\caption{Composition of $\DPBENCH$ problems by horizon type (left) and problem formulation: deterministic vs. stochastic (right).}
	\label{fig-bench_label}
\end{figure}

We annotate each problem with both its application domain and structural characteristics. Although the domain distributions are broadly similar between the easy and hard sets, Figure \ref{fig-bench_label} reveals significant structural differences. The hard benchmark contains 20\% fewer finite-horizon problems and a comparable increase in infinite-average-horizon problems compared to the easy benchmark. Deterministic problems are rare in the hard problem set, only 4.8\%, versus 45.6\% in the easy problem set. Additional structural labels capturing unique modeling formulations or increased complexity, such as action-dependent transitions and time-dependent state spaces, exhibit similar divergence across the two sets. While a comprehensive list of labels and definitions is provided in our supplementary materials, we omit the details due to space constraints, and the differences summarized above already support the validity of our easy-vs-hard benchmark classification.

{\color{blue}
\begin{table}[t]
\centering
\caption{Performance of representative LLMs on $\DPBENCH$. Results for additional models are reported in Table \ref{tab:main_results}. Note that we use the \texttt{gpt-4o-2024-08-06} version of GPT-4o to match the version used for generating our synthetic training data, while all other models use their originally released versions.}
\label{tab:main_results_sub}
\begin{tabular}{clrrrrr}
\toprule & \textbf{Model} & \textbf{Parameters} & \textbf{Easy(\%)} & \textbf{Hard(\%)} & \textbf{Micro(\%)} & \textbf{Macro(\%)} \\
\midrule
& DeepSeek-R1         & 671B              & 73.3 & 28.6 & 59.1 & 51.0 \\
& o1                    & 300B & 48.9 & 31.0 & 43.2 & 39.9\\
 & GPT-4o              & 200B & 47.8 & 19.0 & 38.6 & 33.4 \\
 & Qwen-2.5-7B-Instruct& 7B                & 11.1 & 4.8  & 9.1  & 7.9  \\
\bottomrule
\end{tabular}

\end{table}
}

The performance of representative SOTA LLMs on $\DPBENCH$ is reported in Table \ref{tab:main_results_sub}, with additional large-scale baselines reported later in Table \ref{tab:main_results}. We also compute micro-average and macro-average accuracies: the micro-average aggregates all examples across difficulty levels into a single pool, ensuring each problem contributes equally to the final score, while the macro-average takes the mean accuracy across the easy and hard sets, offering a balanced view of performance across difficulty levels.
Among the large-scale baselines, the best accuracies on the easy and hard benchmarks are 73.3\% and 35.7\%, respectively. In contrast, small-scale models (with fewer than 10B parameters), which we fine-tune in this study, achieve at most 11.1\% and 4.8\%, respectively. These results highlight the difficulty of DP problems for LLMs, especially when compared to their strong performance on textbook-level LP benchmarks.

\section{Framework of Synthetic Data Generation}\label{sec-synthetic}
Fine-tuning LLMs typically requires substantial amounts of high-quality, task-specific training data. Unlike LP and MIP, which benefit from decades of well-curated problem-instance libraries and standardized benchmarks, fine-tuning LLMs for DP auto-formulation faces a key challenge: the lack of high-quality, domain-specific data. While public programming platforms like LeetCode host numerous DP exercises, these are typically framed from a computer science perspective and differ significantly from the real-world, operations-focused DP problems. To bridge this gap, we propose a two-direction data synthesis framework, $\RD$, tailored for scenarios with limited high-quality seed data. $\RD$ strategically balances accuracy and diversity, enabling the creation of large-scale synthetic datasets suitable for DP auto-formulation.

$\RD$ begins with high-quality seeds sourced from multiple domains. The seed data includes 91 problems sourced from textbooks, primarily from worked examples and selected exercises, and modified in a manner consistent with the benchmark problems. Unlike benchmarks that include only the problem description $\PNL$ and final answer $y^*$, we augment each seed example with a correct solution $r^*=(\CoT^*,\Model^*,\Code^*)$. These curated seed data points serve as the foundation for creating synthetic exercises, reasoning tasks, and corresponding code solutions tailored to DP problems.

From these seeds, our goal is to generate a large-scale synthetic data consisting of $(\PNL,r^*,y^*)$. The key challenge lies in ensuring quality while maintaining diversity: even advanced LLMs struggle to reliably solve DP problems. Simply prompting an LLM to solve new problems often yields a low proportion of correct solutions, with no reliable way to verify their correctness.

To address this challenge, we synthesize data along two complementary directions: forward generation and backward generation. Both approaches expand seed problems through scenario expansion, but they differ in the order of construction: whether the problem or the solution is generated first. Forward generation begins with the problem, allowing greater variability but offering limited guarantees of solution correctness. In contrast, backward generation starts from a known solution to ensure correctness, but its variability is inherently limited by the original seed problems.

\begin{figure}
\centering
\includegraphics[width=\textwidth]{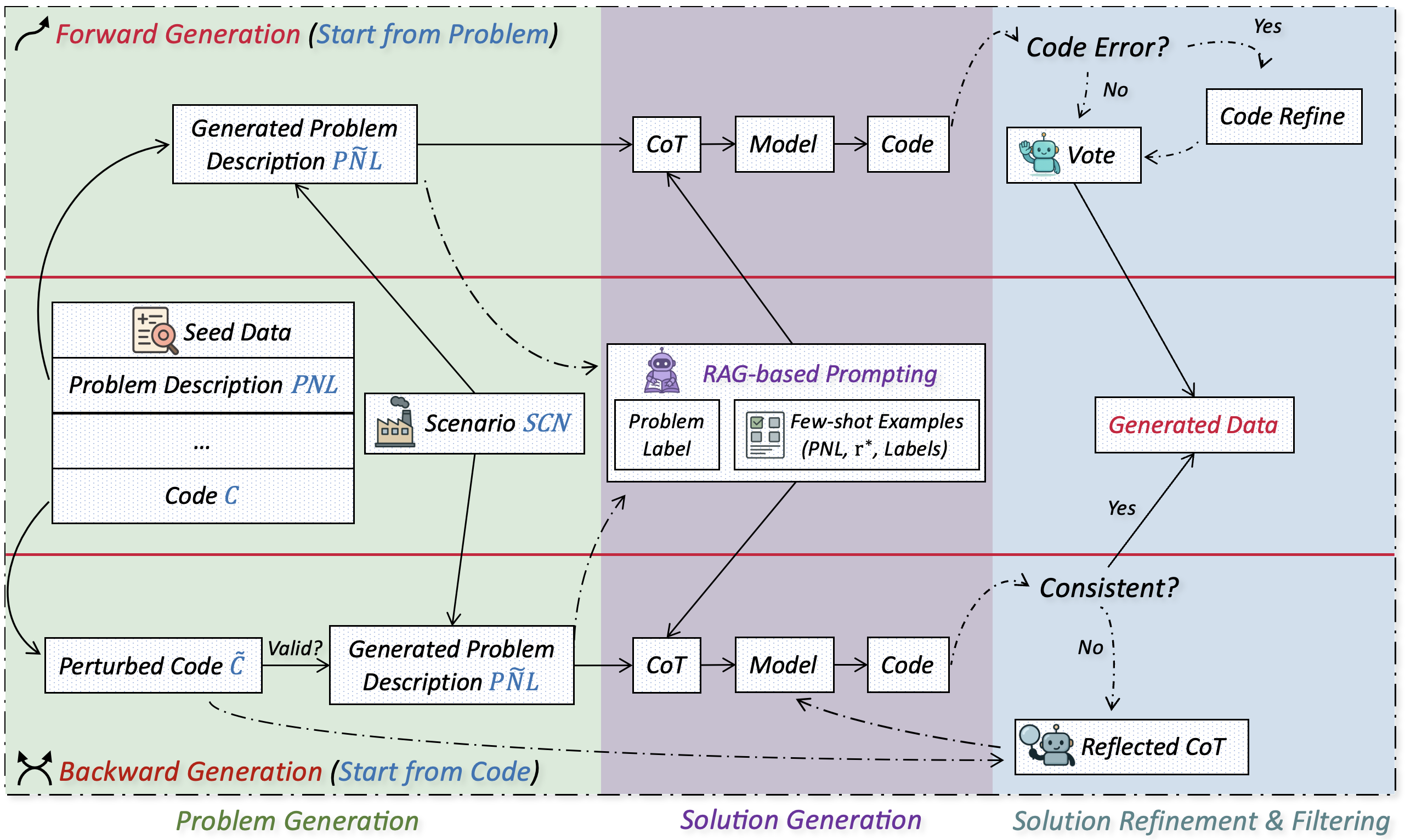}
\caption{
Overview of the $\RD$ synthetic data generation framework.
}
\label{fig-processflow_diagram}
\end{figure}

In what follows, we introduce how we construct a broad range of realistic application scenarios in Section \ref{subsec-scenario}. Since both generation methods require solving DP problems, Section \ref{subsec-RAG} introduces our RAG-based prompting approach, which enhances LLM performance as a solution generator. Sections \ref{subsec-forward} and \ref{subsec-backward} describe the forward and backward generation procedures, respectively. Figure \ref{fig-processflow_diagram} illustrates the overall $\RD$ data generation framework. All data are generated using GPT-4o, and the resulting synthetic dataset is discussed in Section \ref{sec:training-method}.

\subsection{Scenario Expansion}\label{subsec-scenario}
To ensure diversity and realism in our synthetic data, we begin by constructing a broad set of application scenarios that reflect real-world contexts where DP is typically applied. These scenarios do not specify complete problem instances on their own; rather, they serve as contextual templates for reinterpreting and modifying our seed data. Specifically, each new problem is generated by prompting the LLM to adapt a seed-derived component, either the original problem description $\PNL$ in forward generation or the perturbed code $\tilde{\Code}$ (derived from seed $\Code$) in backward generation, to fit a given DP scenario. We group our scenarios into six high-level application categories: (i) Manufacturing and Inventory, (ii) Transportation and Logistics, (iii) Investment and Risk, (iv) Game Strategy, (v) Resource Allocation, and (vi) Others. For each category, we prompt the LLM to generate dozens of scenarios that involve clear trade-offs in decision-making, such as balancing resource usage and profit or managing risk versus reward. Embedding these trade-offs directly into the scenario setup helps guide downstream problem generation toward realistic operational challenges.

\subsection{RAG-Based Solution Generator}\label{subsec-RAG}
Even advanced models like GPT-4o achieve only 38.6\% accuracy on $\DPBENCH$ (Table \ref{tab:main_results_sub}), highlighting the difficulty of generating correct DP solutions directly. To improve the quality and efficiency of synthetic solution generation, we implement a RAG-based few-shot prompting approach. Specifically, given a newly generated problem $\PNL$, we retrieve similar seed examples to guide the LLM in generating its solution $r^*$. However, defining ``similarity'' for DP problems based on $\PNL$ is nontrivial. While semantic similarity (e.g., matching an inventory problem with another inventory example) may seem intuitive, even minor wording changes can imply fundamentally different solution approaches. For example, finite-horizon and infinite-average-horizon inventory problems may have nearly identical $\PNL$s, yet require entirely different algorithms in $\Code$. In such cases, retrieving examples with similar context but mismatched model structures can mislead the LLM and sometimes perform worse than providing no example at all.

To address this challenge, we prioritize structural similarity over surface-level semantics. Each problem is annotated with both its application domain and structural characteristics, such as \textit{infinite-discounted} and \textit{optimal stopping problem}, to capture key modeling and solution characteristics, consistent with the benchmark problems as discussed in Section \ref{Benchmark}.

We manually annotate structural labels for seed examples and then use GPT-4o to automatically label the generated $\PNL$s. These labels are used to narrow the pool of candidate few-shot examples for each $\PNL$. Label-based filtering is then combined with semantic similarity search to retrieve the most relevant examples for each generated $\PNL$. The generation process is decomposed into sequential steps: first generating $\CoT$, then using the problem and $\CoT$ to generate $\Model$, and finally producing $\Code$. At each step, we tailor the prompt using similar examples corresponding to the specific component to guide the LLM. Finally, we execute the generated code $\Code$ to obtain the final answer $y$. Prompts for labeling, and generating CoT-M-C are provided in Appendix Section \ref{appendix_prompts}.

Table~\ref{tab:rag} reports the performance of our RAG-based solution generator under different prompting and retrieval settings. We compare it against: (i) a zero-shot prompt; (ii) Solution Generator (no retrieval), which uses the same structured solution-generation pipeline with detailed step-by-step instructions but no retrieved examples; and (iii)  Solution Generator (fixed retrieval), which uses the same pipeline with fixed examples, where the same four templates (finite deterministic, finite stochastic, infinite average, and infinite discounted) are provided for all problems. Settings (ii), (iii), and the RAG-based solution generator use exactly the same prompting structure and decomposition steps ($\CoT-\Model-\Code$) as the full RAG-based solution generator; the only difference lies in how (or whether) examples are retrieved.

Compared with the zero-shot prompt, the Solution Generator (no retrieval) improves accuracy by approximately 5\%, demonstrating the benefit of structured step-by-step decomposition for GPT-4o in our setting. We note, however, that the effectiveness of explicit decomposition may depend on the underlying model; for more advanced reasoning-oriented models, the marginal benefit of such scaffolding could be smaller. The full RAG-based solution generator further improves micro accuracy from 43.2\% to 52.3\%, with particularly notable gains on the hard subset, demonstrating the benefit of adaptive retrieval. Interestingly, the fixed-example setting achieves slightly lower accuracy than the no-retrieval setting, suggesting that non-adaptive examples may introduce noise rather than useful guidance. In contrast, problem-specific retrieval based on structural labels yields a clear improvement, highlighting the importance of adaptive example selection.

{\color{blue}
\begin{table}[h]
  \centering
  \caption{Performance of GPT-4o under different prompting and retrieval settings.}
  \begin{tabular}{lrrrr}
    \toprule
    \textbf{Method} & \textbf{Easy(\%)} & \textbf{Hard(\%)} & \textbf{Micro(\%)} & \textbf{Macro(\%)} \\
    \midrule
    Zero-shot prompt & 47.8 & 19.0 & 38.6 & 33.4 \\
    Solution generator (no retrieval)$^\dagger$ & 52.2 & 23.8 & 43.2 & 38.0 \\
    Solution generator (fixed retrieval)$^\dagger$ & 50.0 & 23.8 & 41.7 & 36.9 \\
    Solution generator (RAG-based)$^\dagger$ & 58.9 & 38.1 & 52.3 & 48.5 \\
    \bottomrule
  \end{tabular}
  \footnotesize \\

$\dagger$  ``Solution generator'' refers to the same structured $\CoT–\Model–\Code$ pipeline described in Section~\ref{subsec-RAG}. The three variants differ only in how examples are provided: (i) no retrieved examples, (ii) fixed shared examples, and (iii) problem-specific retrieved examples (RAG).
  \label{tab:rag}
\end{table}
}

\subsection{Forward Generation} \label{subsec-forward}
Forward generation consists of three main steps. First, we generate a new problem description $\tilde{\PNL}$ from a seed problem $\PNL$. Next, we apply the RAG-based solution generator introduced in Section \ref{subsec-RAG} to produce a corresponding solution $r$. Finally, we refine and filter the outputs to eliminate potentially incorrect solutions. As this procedure is relatively standardized, we omit the steps here.

Despite downstream augmentation, refinement, and filtering steps, many forward-generated samples still suffer from two major sources of error: (i) the generated solution may contain logical or computational mistakes even when the problem description is valid, and (ii) in some cases, the generated problem itself may be fundamentally ill-posed or incoherent. Both types of errors introduce noise into the training set, and verifying correctness is both difficult and time-consuming for humans. To address this, we turn to backward generation, introduced in the next section.

\subsection{Backward Generation}\label{subsec-backward}
Backward generation reverses the synthesis process by starting with a correct solution and then constructing a compatible problem description. This approach has been successfully applied in other mathematical domains, such as Lyapunov function discovery, where correctness can be easily verified \citep{alfarano_global_2024}. However, in the context of DP, generating valid and coherent problems from known solutions poses challenges similar to those faced when verifying solutions in forward generation. To overcome this, we design a tailored backward generation framework that leverages our confidence in the correctness of the starting solution. Building on this idea, our backward generation approach complements forward generation in three key ways: (i) it guarantees correctness in both the problem and solution generation, (ii) it produces CoT reasoning-trajectories with self-reflection that are valuable for SFT, and (iii) it retains difficult problems in the synthetic dataset that would otherwise be filtered out. The backward generation pipeline consists of the following three steps.

\paragraph{\textbf{Step 1: Generating the problem from the solution.}}
We begin by constructing a new problem description $\tilde{\PNL}$ from a modified code $\tilde{\Code}$, which is derived from the seed code $\Code$ and serves as the ground-truth solution. Specifically, $\tilde{\Code}$ is obtained by keeping the original mathematical formulation $\ModelF$ from a seed code while perturbing problem data $\Data$, such as the number of time periods, transition probabilities, or cost coefficients, by modifying parameters, boundary conditions, or initial values in the seed code. We start from a perturbed code implementation rather than directly modifying the model for two practical reasons. First, $\Model$ is written in natural language, and key components such as constraints or transitions may not be stated explicitly. In contrast, $\Code$ provides a precise and unambiguous specification of the model. Second, since we ultimately need to execute the code to obtain the final answer, working directly with code helps avoid potential inconsistencies or errors that could arise from adding an extra step, such as regenerating code from a modified model description. To ensure that the modified code corresponds to a meaningful and executable DP model, we apply a code-level filter to discard any $\tilde{\Code}$ instances with runtime errors or semantically invalid outputs. This process gives us high confidence in $\tilde{\Code}$ as a correct solution, which we treat as ground truth in the backward generation process. Note that the specific solution algorithm implemented in $\Code$ is not essential. Only the underlying formulation $\ModelF$ encoded in the code affects the resulting problem description $\tilde{\PNL}$.

After passing this filter, $\tilde{\Code}$ is used to generate a new problem description $\tilde{\PNL}$, guided by a scenario prompt. Interestingly, the textual similarity between the original $\PNL$ and the backward-generated $\tilde{\PNL}$ is typically lower than in forward generation, since $\tilde{\PNL}$ is written independently from the code without referencing the original description.

\paragraph{\textbf{Step 2: Solving and verifying the problem.}}
Although we have already obtained the generated problem $\tilde{\PNL}$ and its corresponding code $\tilde{\Code}$, additional verification is necessary to ensure that $\tilde{\PNL}$ is well-posed and to reduce the risk of propagating flawed problems into the fine-tuning dataset. To this end, we apply our RAG-based solution generator to produce an initial solution $(\CoT^0, \Model^0, \Code^0)$ and a final answer $y^0$. We then verify whether the final answer $y^0$ matches the output of $\tilde{\Code}$. If the outputs match, we accept $(\tilde{\PNL}, \CoT^0, \Model^0, \Code^0)$ as a valid training data sample.

\begin{figure}
\centering
\includegraphics[width=0.8\textwidth]{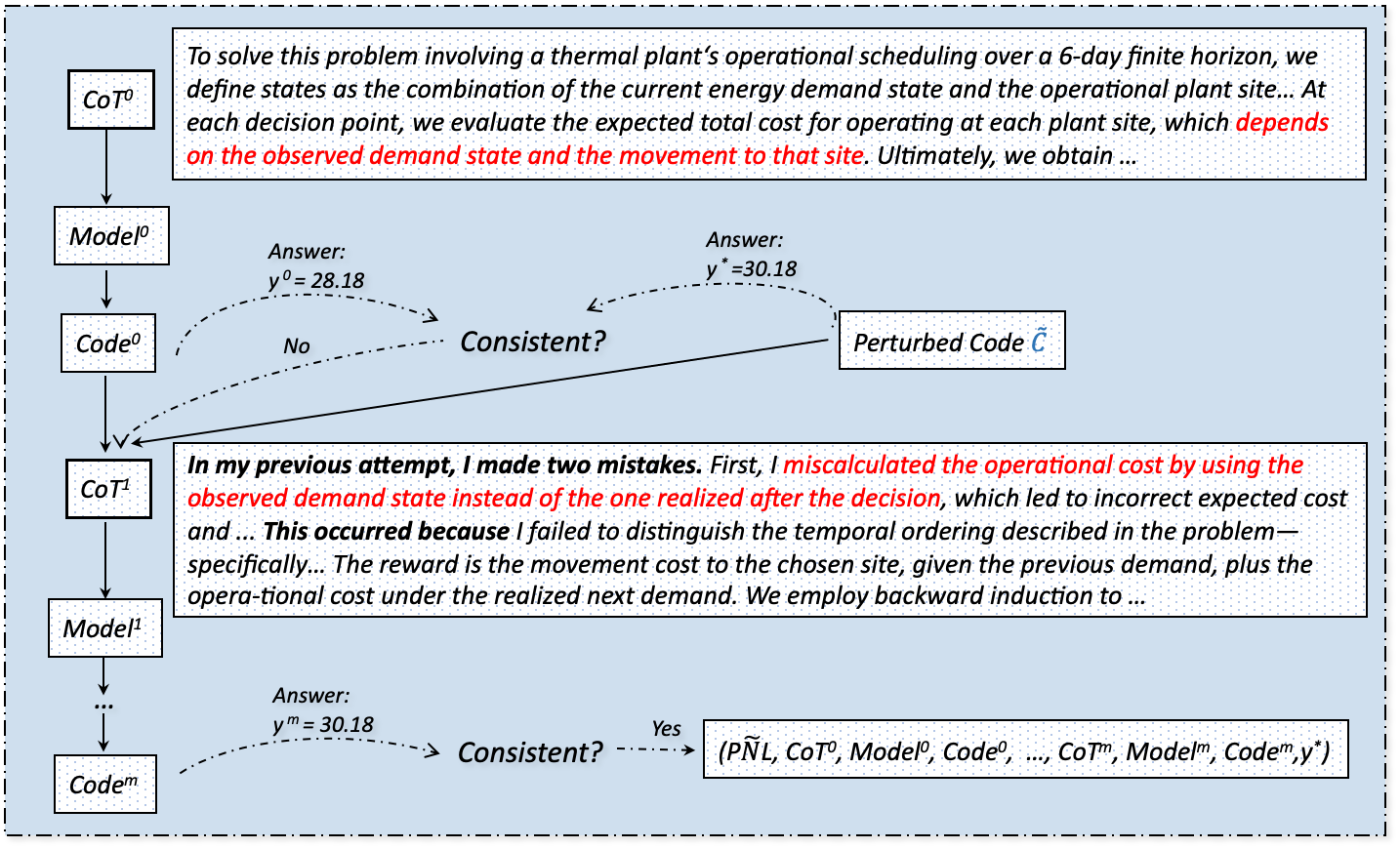}
\caption{
Illustration of the Reflected CoT process and outputs.
}
\label{fig-reflected_cot}
\end{figure}

\paragraph{\textbf{Step 3: Solution trajectory generation and problem recovery via Reflected CoT.}}
To more effectively leverage the ground-truth code $\tilde{\Code}$, beyond its role as a filter for validating generated problems, we introduce a Reflected CoT approach that allows the LLM to learn from the reference solution without merely copying it. This method is inspired by the rationalization technique proposed by \citet{zelikman2022star}, but extends it to a substantially more complex setting. Unlike the short math or commonsense problems considered in their work, our problem descriptions may contain flaws, and the reasoning required to derive a solution is considerably more involved. Specifically, we present the LLM with its own failed code $\Code^0$ alongside the correct reference $\tilde{\Code}$, and prompt it to identify discrepancies and revise its reasoning. Rather than immediately regenerating the full solution, which risks producing a superficially plausible response even when the problem is ill-posed, we first prompt the model to generate a revised chain-of-thought $\CoT^1$ based on its reflection. We then re-run Step 2 using $\CoT^1$ to produce an updated solution tuple $(\CoT^1, \Model^1, \Code^1)$ and check whether $\Code^1$ yields the correct output. If not, the correction process is repeated. This loop continues until the generated code matches the correct answer, with a maximum of six attempts. We set this cap based on empirical observations that each additional round yields diminishing returns. An illustration of the Reflected CoT process is shown in Figure \ref{fig-reflected_cot}.

Reflected CoT naturally produces full reasoning trajectories in the format $(\tilde{\PNL}, \CoT^0, \Model^0, \Code^0, \CoT^1, \Model^1,\Code^1, \dots,y^*)$, providing valuable supervision signals that capture both the final solution and the iterative reasoning process. More importantly, it enables us to recover 19.1\% of samples that would otherwise be discarded, including approximately 20.8\% of new problems that fail to yield a correct solution in any of the initial attempts across all roles. This enrichment introduces more challenging problems into the dataset, which are especially valuable for RL. Further details and the full algorithm are provided in Algorithm \ref{alg-back} in the appendix.

Although backward generation ensures solution correctness, it inherently limits problem diversity. All backward-generated problems, regardless of how many are synthesized or how much the industry context varies, ultimately share the same model formulations $\ModelF$ as the 91 seed problems. Consequently, the variety of underlying problem structures remains constrained by the original seed formulations. To introduce new dynamics, constraints, and decision structures, forward generation is essential: it enables the creation of novel problem formulations by adapting the mathematical structure to new application contexts beyond those captured by the original seeds. Overalll, their roles are complementary, and their distinct effects are reflected in the experimental results discussed in Section \ref{sec:data_scaling}.

\section{Training Method}\label{sec:training-method}
After constructing the synthetic dataset described in the previous section, we next train $\DPLM$ using a two-stage procedure combining SFT and RL. Relying exclusively on RL to equip a model with complex task-solving capabilities often results in training instability, unpredictable outputs, and substantial computational overhead \citep[e.g.,][]{zheng2023secretsofrlhf}. In contrast, “cold‑starting” the model with SFT has proved to be a more efficient way to learn the required output format (e.g., \citealp{ouyang2022instructgpt}). Recent comparative studies highlight that both SFT and RL are essential and play complementary roles: SFT memorizes demonstrations, whereas RL improves out-of-distribution generalization.

\begin{figure}
\centering
\includegraphics[width=\textwidth]{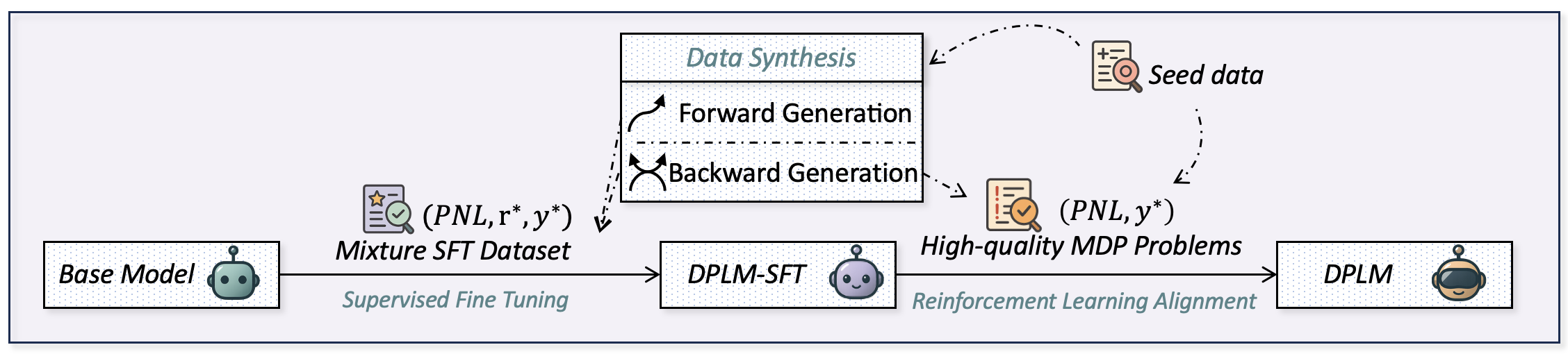}
\caption{
Overview of our training pipeline.
}
\label{fig-training-pipeline}
\end{figure}

Thus, we first fine-tune our model on a dataset containing complex reasoning trajectories and self-reflective CoT, enabling the model to internalize the structure of sophisticated solutions. We then apply RL‑based alignment to further refine the model's capability to consistently produce correct answers. For this RL stage, we compare two representative algorithms: GRPO \citep{shao2024deepseekmath} for online training and DPO \citep{rafailov2023dpo} for offline training. Figure~\ref{fig-training-pipeline} provides an overview of our training pipeline.

\subsection{Stage 1: Supervised Fine-Tuning}\label{subsec:sft}

The SFT stage “cold‑starts” the base model by pulling the initial policy $\pi_{0}$ toward the human instruction distribution via maximum‑likelihood training. This narrows the search space for subsequent RL, mitigates early‑stage mode collapse, and approximates minimizing $\mathrm{KL}(\pi^{\ast}\,\|\,\pi_{0})$, yielding a smoother reward landscape and faster convergence.

\paragraph{Dataset \(\mathcal D_{\mathrm{SFT}}\).}

To support SFT, our $\RD$ data synthesis framework generates a total of 113K training samples using GPT-4o. Among these, 70K are forward-generated samples, and 34K are backward-generated samples that were solved correctly on the first attempt without invoking reflected CoT. Both sets share the standard format $(\PNL,\CoT^*,\Model^*,\Code^*)$. The remaining 8K backward-generated samples are produced via reflected CoT and include a complete reflection–revision trajectory of the form $(\PNL, \CoT^0, \Model^0, \Code^0, \dots, \CoT^*, \Model^*, \Code^*, y^*)$.

\subsection{Stage 2: Reinforcement Learning}

While SFT on synthetic trajectories teaches the model to produce coherent and well-structured answers, it merely approximates the empirical distribution of the demonstrations. RL overcomes this limitation by directly optimizing a reward signal that aligns with the correctness of the final numeric answer. Consequently, RL enables the model to (1) explore reasoning paths beyond those available in the demonstrations and (2) mitigate residual biases inherited from the teacher model. In analogy, SFT provides a structured lesson plan teaching essential skills, whereas RL offers personalized coaching that refines how these skills are practically applied.

The RL alignment process is made possible by our verifiable synthetic dataset, which supplies a reliable reward signal. Specifically, our RL training leverages a curated corpus, $\mathcal{D}_{\text{RL}}$, designed to provide verifiable supervision on challenging instances: (1) approximately 8,000 ``hard-recovered'' problems from the backward-generation pipeline--initially unsolved by the teacher LLM but later passing the reflected CoT consistency filter after iterative correction; (2) the 91 textbook seed problems, each with verified numeric and code solutions. In total, $\mathcal{D}_{\text{RL}}$ contains roughly 8,100 verified problem–solution pairs.

The standard approach to RL alignment is Proximal Policy Optimization (PPO), which involves sampling rollouts\footnote{A rollout is the sequence of states, actions, and rewards generated as an agent interacts with an environment while following a specific policy. It captures the trajectory of the agent's experience.}, estimating advantage functions using a value critic, and maximizing a clipped, KL-regularised surrogate objective. However, on thousand-token reasoning traces, PPO’s value critic significantly increases memory usage, and the long credit-assignment horizon often destabilizes training. To address these issues, we separately adopt two recent, more lightweight alternatives: (1) DPO, which eliminates the need for both the value critic \emph{and} online sampling by training directly on static preference pairs; and (2) GRPO, which removes the value critic by computing baselines internally \emph{within groups} of $k$ rollouts generated from the same prompt.

\paragraph{DPO: Direct preference optimization.}
For DPO, we construct a static preference dataset by repeatedly sampling the \texttt{DPLM-7B-SFT} model on prompts from $\mathcal{D}_{\text{RL}}$. In each sampling round for a prompt $x$, we sample up to 8 trajectories, execute them, and form preference pairs from the correct and incorrect trajectories observed for the same prompt. We repeat this process across prompts and sampling rounds, allowing a single prompt to contribute multiple valid pairs when available. Let $\mathcal{D}_{\text{pref}}$ denote the resulting offline dataset of paired preferences $(x,y_w,y_l)$.
\[
  \mathcal{L}_{\mathrm{DPO}}
  \;=\;
  -\,
   \mathbb{E}_{(x, y_w, y_l) \sim \mathcal{D}_{\text{pref}}}
   \Bigl[
    \log\sigma\!\bigl(
      \beta\bigl(\log\pi_\theta(y_w\mid x)-\log\pi_\theta(y_l\mid x)\bigr)
    \bigr)
  \Bigr],
\]
where $\sigma$ is the sigmoid function and $\beta$ controls the reward–KL trade-off. This objective is equivalent to maximizing the same KL-regularized policy improvement targeted by PPO-based RLHF, but achieves it several times faster while requiring only preference pairs.

\paragraph{GRPO: Group relative policy optimization.}
For a given prompt $x$, we first sample $k$ candidate completions $\{y_i\}_{i=1}^{k}$ from the current policy $\pi_{\text{current}}$ and evaluate them using the reward function, yielding scores $\{r_i\}_{i=1}^k$. We then compute a \emph{group–normalized} advantage function $A_i = \frac{r_i - \operatorname{mean}(r)}{\operatorname{std}(r)}$, and optimize the following clipped surrogate objective:
\[
  \max_{\theta}\;
  \mathbb{E}_{x \sim \mathcal{D},\; y_i \sim \pi_{\text{current}}(\cdot \mid x)}\bigg[
    \min\left(\rho_i A_i,\;
              \operatorname{clip}(\rho_i, 1\pm\varepsilon)A_i\right)
  \bigg]
  - \beta\, D_{\mathrm{KL}}\left(\pi_\theta \,\|\, \pi_{\text{ref}}\right),
\]
where $\rho_i = \frac{\pi_\theta(y_i \mid x)}{\pi_{\text{current}}(y_i \mid x)}$ is the importance weight between the updated policy and current policy, and the operator $\operatorname{clip}(\rho, 1 \pm \varepsilon)=\max\bigl( 1- \varepsilon,\;\min(\rho,\;1+\varepsilon)\bigr)$ constrains the policy ratio to stabilize updates. The reference policy $\pi_{\text{ref}}$ denotes the checkpoint used for KL regularization prior to the start of RL. Because the baseline is computed internally within each group, GRPO eliminates the need for a learned value critic. Compared to PPO, this reduces GPU memory usage by approximately 40\% and leads to faster convergence on long-horizon reasoning tasks \citep{shao2024deepseekmath}.

To facilitate verifiable training, we define the total reward as a composition of two sub-rewards: $r(x, \hat a, a^*) = r_{\text{format}}(x, \hat a) + r_{\text{answer}}(\hat a, a^*)$, where $r_{\text{format}}$ reflects the syntactic and executable validity of the model output $\hat a$, and $r_{\text{answer}}$ measures its semantic correctness against the ground-truth $a^*$:
\[
  r_{\text{format}}(x, \hat a) =
  \begin{cases}
    0.2 & \text{if } \hat a \text{ is executable and well-typed};\\
    0   & \text{if } \hat a = \textsc{null} \text{ or fails execution},
  \end{cases}
  \qquad
  r_{\text{answer}}(\hat a, a^*) =
  \begin{cases}
    0.8 & \text{if } \hat a = a^*;\\
    0   & \text{otherwise}.
  \end{cases}
\]
This hierarchical design encourages the policy to first generate structurally valid formulations before converging toward exact numeric correctness. The reward of $0.2$ for executable-but-incorrect completions provides an intermediate learning signal that helps mitigate the sparse-reward bottleneck of binary schemes, especially in multi-stage tasks such as DP formulation.
However, setting this intermediate reward too high may lead to ``reward hacking'', where the model converges to syntactically plausible but semantically invalid outputs. Following prior work such as \cite{shao2024deepseekmath}, we separate execution feedback from solution accuracy, which has been shown to improve training stability. This shaping design is especially aligned with DP auto-formulation, where outputs must satisfy both structural and computational constraints. For instance, an incorrect formulations may still contain valid subcomponents (e.g., a correct recurrence structure but incorrect indexing), making partial credit a valuable inductive bias.

In conclusion, GRPO is trained online, continually updating the policy using fresh rollouts and their associated reward signals.

\section{Experiments}\label{sec:experiments}

We conduct extensive numerical experiments to evaluate our proposed framework for DP auto-formulation. Specifically, we aim to answer three primary research questions: (1) How effective is our training pipeline, and can our fine-tuned model outperform its much larger teacher model? (2) What are the distinct scaling behaviors of the forward and backward generation methods? (3) Are both SFT and RL necessary components, and what are their individual contributions to the model's final performance?

We begin by presenting our main results against leading baselines in Section \ref{susec-results}. We then analyze the scaling properties of the forward and backward generation methods in Section \ref{sec:data_scaling} and conduct a detailed ablation study in Section \ref{sec:ablation}. Additional analyses on model-size and inference scaling were conducted but are omitted here due to space constraints.

\subsection{Experimental Setup}\label{subsec-setup}

\paragraph{Tasks and Data Splits.}
All experiments are conducted on $\DPBENCH$, our curated benchmark of textbook–level DP problems described in Section~\ref{Benchmark}. The benchmark contains 132 problems, divided into 90 easy and 42 hard instances.

\paragraph{Models and Training Recipes.}
We choose \texttt{Qwen-2.5-7B-Instruct} after a comparison with other 7–10B checkpoints. We initialize $\DPLM$ and employ a two-stage pipeline consisting of SFT followed by RL alignment.

\paragraph{Baselines.}
We benchmark against a diverse suite of open-source and closed-source LLMs, spanning two orders of magnitude in parameter count.

\paragraph{Evaluation Metrics.}
We report pass@1 accuracy, which measures the proportion of problems solved correctly on the first attempt, reflecting the model’s ability to provide direct, accurate answers without retries. We also compute micro-average and macro-average accuracies.

\subsection{Main Results}\label{susec-results}

The results presented in Table \ref{tab:main_results} show that our final model, \texttt{DPLM-7B-SFT-GRPO}, delivers highly competitive performance, achieving the highest scores on the hard problem set (40.5\%) and the macro-average (53.6\%). It is also a top contender on the easy set (66.7\%) and in the micro-average (58.3\%), ranking second only to \texttt{DeepSeek-R1} while outperforming all other baselines, including \texttt{GPT-5}, \texttt{o1}, \texttt{GPT-4o}, and \texttt{DeepSeek-V3}. Importantly, it significantly exceeds \texttt{GPT-4o}, the model initially used for synthetic data generation. This outcome effectively addresses the key question of whether our synthesis and training procedures can produce models that outperform the teacher model.

\begin{table}[ht]
  \small
  \centering
  \caption{Performance comparison across models on DP auto-formulation tasks. For each metric, the top value is highlighted in \textbf{bold}, and the second-best is underlined. Note that we use the \texttt{gpt-4o-2024-08-06} version of \texttt{GPT-4o} to match the version used for generating our synthetic training data, while all other models use their originally released versions.}
  \label{tab:main_results}
  \begin{tabular}{clrrrrr}
    \toprule
    \textbf{Type} & \textbf{Model} & \textbf{Parameters} & \textbf{Easy(\%)} & \textbf{Hard(\%)} & \textbf{Micro(\%)} & \textbf{Macro(\%)} \\
    \midrule
    \multirow{7}{*}{\shortstack{Baseline\\ Large-Scale}}
      & \texttt{DeepSeek-R1} & 671B & \textbf{73.3} & 28.6 & \textbf{59.1} & \underline{51.0} \\
      & \texttt{DeepSeek-V3} & 671B & 51.1 & 26.2 & 43.2 & 38.7 \\
      & \texttt{GPT-5} & $^\ast$300B & 50.0 & \underline{35.7} & 45.5 & 42.9 \\
      & \texttt{o1} & $^\ast$300B & 48.9 & 31.0 & 43.2 & 39.9 \\
      & \texttt{GPT-4o} & $^\ast$200B & 47.8 & 19.0 & 38.6 & 33.4 \\
      & \texttt{Qwen-2.5-72B-Instruct} & 72B & 43.3 & 21.4 & 36.4 & 32.4 \\
      & \texttt{Qwen-2.5-32B-Instruct} & 32B & 37.8 & 21.4 & 32.6 & 29.6 \\
    \hdashline
    \multirow{3}{*}{\shortstack{Baseline\\ Small-Scale}}
      & \texttt{Gemma-2-9B-It} & 9B & 6.7 & 2.4 & 5.3 & 4.5 \\
      & \texttt{Llama-3.1-8B-Instruct} & 8B & 6.7 & 2.4 & 5.3 & 4.5 \\
      & \texttt{Qwen-2.5-7B-Instruct} & 7B & 11.1 & 4.8 & 9.1 & 7.9 \\
    \midrule
    \multirow{2}{*}{\shortstack{Ours}}
      & \texttt{DPLM-7B-SFT} (SFT Only) & 7B & 40.0 & 21.4 & 34.1 & 30.7 \\
      & \texttt{DPLM-7B-SFT-GRPO} & 7B & \underline{66.7} & \textbf{40.5} & \underline{58.3} & \textbf{53.6} \\
    \bottomrule
  \end{tabular}
  \footnotesize \\
  \quad $^\ast$ Parameter counts for \texttt{GPT-5}, \texttt{o1} and \texttt{GPT-4o} are literature-based estimates (e.g., \citealp{BenAbacha2025MEDEC, gpt5_para}), since OpenAI has not publicly disclosed them.
\end{table}

Notably, even solely with SFT (without RL), the \texttt{DPLM-7B-SFT} variant achieves a pass@1 rate of 40.0\% on easy problems and 21.4\% on hard problems, substantially outperforming the base model \texttt{Qwen2.5-7B-Instruct}, which achieves only 11.1\% and 4.8\% respectively. This demonstrates the effectiveness of the $\RD$ data synthesis process and the high quality of the resulting synthetic dataset. Incorporating the GRPO RL stage significantly boosts the performance,

highlighting the benefits of RL-based alignment.

While our model notably outperforms DeepSeek-R1 on the hard problem set, demonstrating particular strength in complex reasoning scenarios, it lags behind on easy problems.  Further inference scaling experiments (see Figure \ref{fig:test-scaling} in Appendix) suggest this shortfall likely comes from limited coverage of specialized domain knowledge within our current dataset. Expanding the initial seed data and scaling up the synthetic data generation process seem promising to overcome this limitation and enhance future performance.

Finally, the relatively poor performance of general-purpose large model baselines on $\DPBENCH$ highlights the inherent challenges of applying such models to specialized domains such as DP. This highlights the need for custom-tailored models explicitly trained for domain-specific tasks.

\subsection{Forward vs. Backward Generation for Data Scaling}\label{sec:data_scaling}

\begin{figure}
\centering
\includegraphics[width=0.7\textwidth]{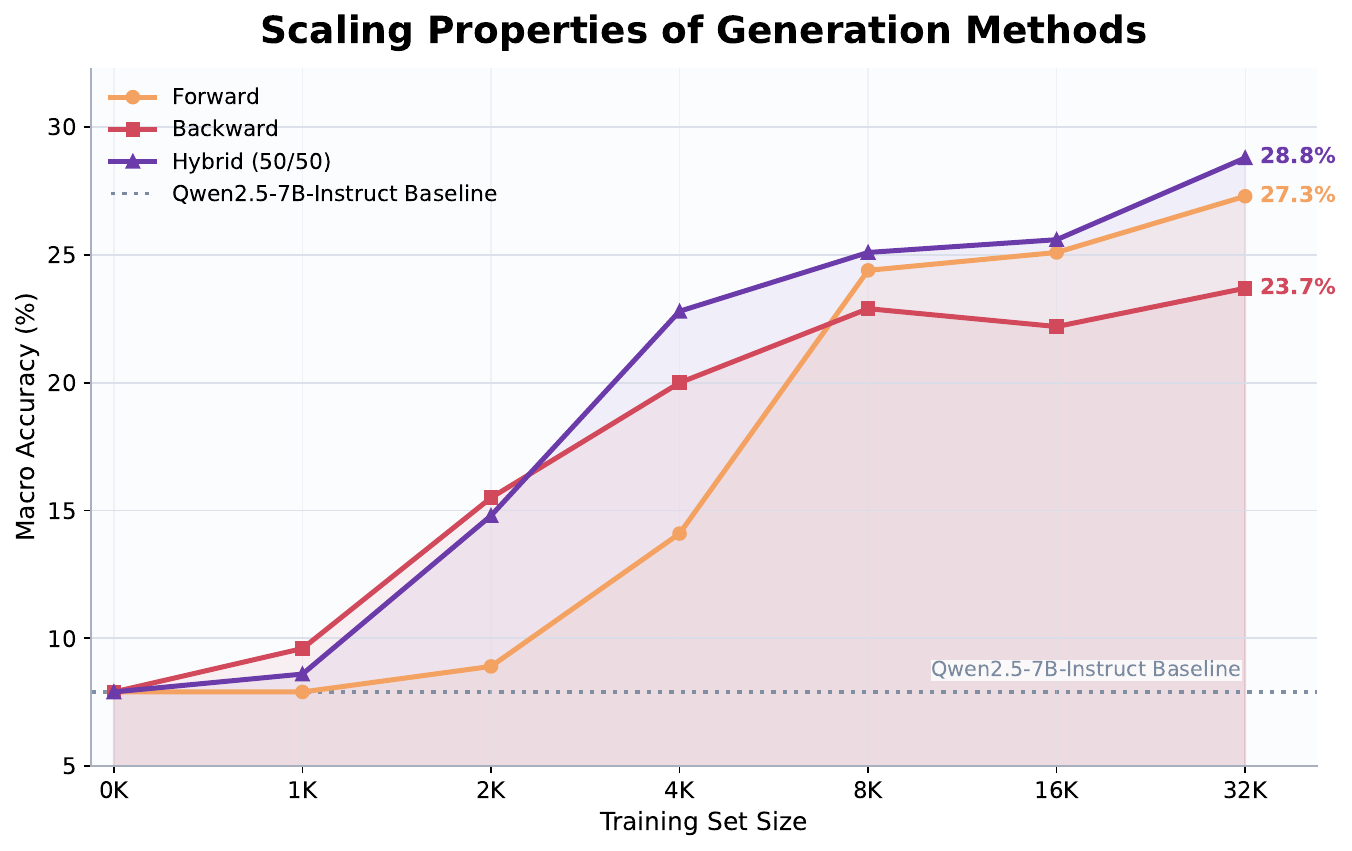}
\caption{Macro-average accuracy of the 7B SFT model under forward, backward, and 50/50 hybrid data generation as the SFT training set size increases.}
\label{forward_vs_backward}
\end{figure}

In this subsection, we evaluate how scaling the SFT training dataset affects the 7B model's macro-average accuracy by comparing forward and backward generation techniques. We randomly sample subsets of varying sizes (1K, 2K, 4K, 8K, 16K, 32K) from a validated dataset and use these subsets for SFT. We also compare both methods to a hybrid approach in which each dataset consists of 50\% forward- and 50\% backward-generated data.

Figure~\ref{forward_vs_backward} demonstrates that all three approaches improve model performance as training size increases, though they do so differently. As discussed in Section \ref{sec-synthetic}, forward generation enriches the dataset with diverse problem formulations. While it offers limited improvement initially, forward generation significantly enhances the model's ability to generalize as the dataset size increases. Conversely, backward generation emphasizes high-quality, verified solutions, which initially provides rapid performance gains even at smaller dataset sizes. However, this approach encounters diminishing returns as the dataset expands, primarily due to its inherent limitation in diversity. Expanding the initial seed dataset can mitigate this constraint. Nonetheless, the backward generation method is valuable for its ability to produce verifiable problems and solutions, a crucial advantage given data scarcity.

The performance of the hybrid approach validates the trade-off between diversity and correctness. At smaller data scales, where correctness is the dominant factor, it is outperformed by the backward method. However, as the dataset grows, the hybrid approach consistently outperforms both pure methods.
Overall, our experiments highlight that both the diversity and quality of training data are essential for maximizing model performance.

\subsection{Ablation Study of Two-Stage Training}\label{sec:ablation}

To isolate and understand the individual contributions of SFT and RL alignment, we train six variants of the base model \texttt{Qwen-2.5-7B-Instruct} and compare their performance. All variants are trained using identical datasets and hyperparameters to ensure comparability.

\begin{table}[htb]
  \centering
  \caption{Pass@1 accuracy across different training recipes. ``Hard" and ``Easy" refer to the 42 difficult and 90 easier problems, respectively, in $\DPBENCH$.}
  \label{tab:ablation}
  \begin{tabular}{lrrrr}
    \toprule
    \textbf{Model}  & \textbf{Easy(\%)} & \textbf{Hard(\%)} & \textbf{Micro(\%)} & \textbf{Macro(\%)} \\
    \midrule
    Base-Model-7B (no further training)     & 11.1 & 4.8 & 9.1 & 7.9 \\
    DPLM-7B (SFT only)                      & 40.0 & 21.4 & 34.1 & 30.7 \\
    DPLM-7B (DPO only)                      & 23.3 & 9.5 & 18.9 & 16.4 \\
    DPLM-7B (GRPO only)                     & 27.8 & 14.3 & 23.5 & 21.1 \\
    DPLM-7B (SFT\,$\rightarrow$\,DPO)       & 48.9 & 21.4 & 40.2 & 35.2 \\
    DPLM-7B (SFT\,$\rightarrow$\,GRPO)      & \textbf{66.7} & \textbf{40.5} & \textbf{58.3} & \textbf{53.6} \\
    \bottomrule
  \end{tabular}
\end{table}

The results in Table \ref{tab:ablation} show that SFT is by far the dominant driver of performance improvement. A single two-epoch SFT pass boosts micro-average accuracy from 9.1\% to 34.1\%, a gain of 25 percentage points. In contrast, the RL-only variants achieve substantially lower performance: 23.5\% with GRPO and 18.9\% with DPO, indicating that reward-driven training alone is not competitive with high-quality supervision. One possible explanation is that a 7B-parameter base model lacks the capacity to reliably discover high-quality reasoning trajectories from scratch; under constrained training budgets, SFT is thus markedly more efficient than RL.

Although RL alone underperforms SFT, it remains a valuable refinement step. When initialized from the SFT checkpoint, GRPO raises accuracy to 58.3\%, while DPO also improves, albeit to a lower 40.2\%. Across all configurations, GRPO consistently outperforms DPO in accuracy. However, this advantage comes with a substantial computational cost: on identical hardware (eight H100-80GB GPUs), GRPO requires approximately eight times more wall time due to the need for on-policy rollout generation and group-normalized advantage computation.

\subsection{Stronger Teacher Supervision as a Scalable Path for $\DPLM$}\label{sec:teacher_guidance}

An interesting question is whether $\DPLM$ can continue to improve as teacher quality improves. To explore this scaling direction, we replace the original teacher with \texttt{DeepSeek-V3.2} and synthesize additional training data for both stages. Specifically, we generate 47K new trajectories to augment the original 113K SFT dataset ($42\%$ increase) and 3K verified instances to augment the original 8K RL dataset (38\% increase). We then repeat the same two-stage training recipes (SFT and SFT$\rightarrow$GRPO) using these augmented datasets.

\begin{table}[ht]
  \small
  \centering
  \caption{Performance impact of augmenting the training set with additional data synthesized by a stronger teacher model (\texttt{DeepSeek-V3.2}). The \textit{Inference} column specifies whether results are reported with pass@1 or inference-time self-consistency ($k=5$).}
  \label{tab:teacher_guidance}
  {\setlength{\tabcolsep}{4pt}
  \begin{tabular}{llcrrrr}
    \toprule
    \multicolumn{1}{c}{\textbf{\shortstack[c]{Training\\Data}}} & \multicolumn{1}{c}{\textbf{\shortstack[c]{Training\\Recipe}}} & \multicolumn{1}{c}{\textbf{\shortstack[c]{Inference}}} & \multicolumn{1}{c}{\textbf{\shortstack[c]{Easy(\%)}}} & \multicolumn{1}{c}{\textbf{\shortstack[c]{Hard(\%)}}} & \multicolumn{1}{c}{\textbf{\shortstack[c]{Micro(\%)}}} & \multicolumn{1}{c}{\textbf{\shortstack[c]{Macro(\%)}}} \\
    \midrule
    Original data       & SFT only                 & pass@1     & 40.0 & 21.4 & 34.1 & 30.7 \\
    Original data       & SFT$\rightarrow$GRPO     & pass@1     & 66.7 & 40.5 & 58.3 & 53.6 \\
    Original + new data & SFT only                 & pass@1     & 53.3 & 23.8 & 43.9 & 38.6 \\
    Original + new data & SFT$\rightarrow$GRPO     & pass@1     & 71.1 & 42.9 & 62.1 & 57.0 \\
    Original + new data & SFT$\rightarrow$GRPO     & SC@$k=5$   & 78.9 & 54.8 & 71.2 & 66.8 \\
    \bottomrule
  \end{tabular}}
\end{table}

Table \ref{tab:teacher_guidance} demonstrates that the enhanced teacher supervision consistently boosts performance across all metrics. SFT-only benefits substantially, with micro-average accuracy rising from 34.1\% to 43.9\%. The subsequent GRPO stage further amplifies this gain to 62.1\%. Most notably, with inference-time majority voting ($k=5$), the model achieves a micro-average accuracy of \textbf{71.2\%} and a macro-average of \textbf{66.8\%}. These results significantly outperform our previous best configurations and even surpass the SOTA baseline \texttt{DeepSeek-R1} reported in Table \ref{tab:main_results}. More importantly, they indicate that $\DPLM$ is not yet saturated in our setting: improvements in teacher quality can be systematically converted into student gains under the same training pipeline, suggesting a practical path for future improvement.

\subsection{Out-of-Domain Performance}
Due to the scarcity of domain-specific DP datasets, both our seed data and benchmark problems are primarily derived from textbooks and therefore exhibit limited contextual diversity. This raises a natural question: how well does $\DPLM$ generalize to out-of-domain contexts?

To investigate this, we construct two small healthcare-based out-of-domain datasets, focusing specifically on clinical decision-making settings. The first (denoted \textbf{Healthcare-OOD1}) evaluates $\DPLM$ on problems where the underlying DP structure remains in-domain, but the application context is shifted to medical treatment. This allows us to isolate the effect of contextual shift, while keeping the mathematical DP model unchanged.
The second dataset (denoted \textbf{Healthcare-OOD2}) is more challenging and consists of four structurally complex medical decision problems derived from classical MDP applications summarized in \citet{Schaefer2004}. Details of two dataset constructions are provided in Appendix~\ref{app:ood_construction}.

\begin{table}[h]
\centering
\caption{Performance on Healthcare-OOD1. ``Underlying Benchmark'' refers to performance on the corresponding original DP-Bench problems.}
\label{tab:ood1}
\begin{tabular}{lrr}
\toprule
\textbf{Dataset} & \textbf{Total} & \textbf{Accuracy(\%)} \\
\midrule
Healthcare-OOD1 (Full Set) & 114 & 58.8 \\
Underlying Benchmark ($\DPBENCH$ overlap) & 114 & 67.5 \\
Healthcare-OOD1 (Underlying solved $\geq$ 3 times) & 80 & 70.3 \\
Healthcare-OOD1 (Underlying solved $\geq$ 5 times) & 77 & 70.4 \\
\bottomrule
\end{tabular}
\end{table}
Table~\ref{tab:ood1} summarizes performance on Healthcare-OOD1. On the full set of 114 problems, $\DPLM$ achieves 58.77\% accuracy, compared to 67.5\% accuracy on the corresponding underlying DP-benchmark problems. The performance gap appears largely attributable to output variability rather than structural misunderstanding. When restricting attention to problems that $\DPLM$ solves consistently on the original benchmark (e.g., three or five successful runs out of five), the accuracy on Healthcare-OOD1 increases from 58.8\% to 70.3\%. Error analysis suggests that the performance gap arises primarily from differences in natural-language presentation rather than from structural misunderstanding. In particular, Healthcare-OOD1 problems are often more implicit and less mathematically explicit, increasing ambiguity in problem interpretation. Importantly, errors do not appear to stem from unfamiliarity with the healthcare context. Additional examples and analysis are provided in Appendix~\ref{app:ood1}.

For Healthcare-OOD2, we conduct a qualitative case study of $\DPLM$. Full problem descriptions, outputs, and analyses are provided in Appendix~\ref{app:OOD2}. Although $\DPLM$ is not yet fully reliable on structurally complex, real-world medical MDPs, it demonstrates meaningful generalization beyond textbook contexts and can serve as a strong draft generator with light human correction. Specifically, $\DPLM$ solves two of the four Healthcare-OOD2 problems completely correctly and produces only minor errors on the remaining two. For example, in one case $\DPLM$ incorrectly states in its reasoning that it applies backward induction for an infinite-horizon problem, while the generated code correctly implements value iteration and produces the correct result. As a baseline, we also evaluate GPT-4o, which fails on all four problems with critical modeling errors.

While $\DPLM$ is largely able to formulate the model and generate executable code for Healthcare-OOD2 problems, we note an important limitation: as problems become more realistic and structurally complex, computational intractability becomes a concern. In the epidemic control problem, even when $\DPLM$ correctly specifies the model and value iteration algorithm, the value iteration becomes difficult to converge due to the enlarged state space. A more practical and promising future direction, therefore, is to use $\DPLM$ primarily as an auto-formulation tool and connect it with more appropriate ADP or RL solvers for large-scale real-world problems.

\section{Conclusion}\label{sec-conclusion}
This paper explores the auto-formulation of DP problems using LLMs, a critical yet unexplored step toward fully automating end-to-end sequential decision-making under uncertainty. We introduce $\DPBENCH$, the first benchmark for this task, featuring 132 textbook-level problems with standardized metrics to evaluate LLM's ability to translate natural language into formal DP models.

We present $\DPLM$, a specialized 7B-parameter LLM trained from scratch on synthetic data distilled from GPT-4o. Despite its smaller size, $\DPLM$ significantly outperforms its teacher model and rivals larger SOTA models including DeepSeek-R1 and OpenAI’s o1, particularly on hard problems. This highlights the promise of domain-specific, small-scale LLMs for DP auto-formulation.

To support model training, we propose $\RD$, a data generation framework that balances diversity and accuracy by combining forward and backward generation. A key component is the Reflected CoT mechanism, which recovers difficult problems that would otherwise be discarded and enables reflect-and-refine solution trajectories. Our empirical results reveal a new insight for data synthesis: backward generation is preferred at small scales, while adding forward generation improves performance as data scales.

Overall, our work demonstrates the potential of LLMs for DP automation and offers a practical roadmap for fine-tuning domain-specific models in data-scarce settings with limited supervision.
We view this research as part of a broader effort to reduce barriers to the adoption of operations research methods, which currently often require substantial expert involvement and problem-specific customization. Automating model formulation, and potentially coupling it with automated solvers and reinforcement learning methods, may expand access to DP based decision support tools for practitioners and small firms without formal training in operations research.

\bibliographystyle{informs2014}
\bibliography{LLM}

@article{bertsimas2024robust,
  title={Robust and Adaptive Optimization under a Large Language Model Lens},
  author={Bertsimas, Dimitris and Margaritis, Georgios},
  journal={arXiv preprint arXiv:2501.00568


        },
  year={2024}
}

@incollection{Schaefer2004,
  author = {Schaefer, Andrew J. and Bailey, Mark D. and Shechter, Steven M. and Roberts, Mark S.},
  title = {Modeling Medical Treatment Using {M}arkov Decision Processes},
  booktitle = {Modeling {M}edical {D}ecision Making},
  year = {2004},
  publisher = {Springer US},
  address = {Boston, MA},
  pages = {593--612},
  isbn = {978-0-387-21787-1},
  url = {https://link.springer.com/chapter/10.1007/1-4020-8066-2_23}
}

@article{guo2025deepseek,
  title={Deepseek-r1: Incentivizing reasoning capability in llms via reinforcement learning},
  author={Guo, Daya and Yang, Dejian and Zhang, Haowei and Song, Junxiao and Zhang, Ruoyu and Xu, Runxin and Zhu, Qihao and Ma, Shirong and Wang, Peiyi and Bi, Xiao and others},
  journal={arXiv:2501.12948


        },
  year={2025}
}

@article{huang2024orlm,
  title={ORLM: A Customizable Framework in Training Large Models for Automated Optimization Modeling},
  author={Huang, Chenyu and Tang, Zhengyang and Ge, Dongdong and Hu, Shixi and Jiang, Ruoqing and Wang, Benyou and Wang, Zizhuo and Zheng, Xin},
  journal={Operations Research},
  year={2025},
  volume={73},
  number={6},
  pages={2986-3009}
}

@article{lu2025optmath,
  title={OptMATH: A Scalable Bidirectional Data Synthesis Framework for Optimization Modeling},
  author={Lu, Hongliang and Xie, Zhonglin and Wu, Yaoyu and Ren, Can and Chen, Yuxuan and Wen, Zaiwen},
  journal={arXiv preprint arXiv:2502.11102},
  year={2025}
}

@InProceedings{ramamonjison2023nl4opt,
  title = 	 {NL4Opt Competition: Formulating Optimization Problems Based on Their Natural Language Descriptions},
  author =       {Ramamonjison, Rindranirina and Yu, Timothy and Li, Raymond and Li, Haley and Carenini, Giuseppe and Ghaddar, Bissan and He, Shiqi and Mostajabdaveh, Mahdi and Banitalebi-Dehkordi, Amin and Zhou, Zirui and Zhang, Yong},
  booktitle = 	 {Proceedings of the NeurIPS 2022 Competitions Track},
  pages = 	 {189--203},
  year = 	 {2022},
  editor = 	 {Ciccone, Marco and Stolovitzky, Gustavo and Albrecht, Jacob},
  volume = 	 {220},
  series = 	 {Proceedings of Machine Learning Research},
  month = 	 {28 Nov--09 Dec},
  publisher =    {PMLR}
}

@article{bray_tutorial_2025,
   title = {A Tutorial on Teaching Data Analytics with Generative {AI}},
   author={Bray, Robert L},
   journal = {{INFORMS} Journal on Applied Analytics},
   year={2025},
   volume={55},
   number={4},
   pages={319-343}
}

@inproceedings{alfarano_global_2024,
	title = {Global Lyapunov functions: a long-standing open problem in mathematics, with symbolic transformers},
	volume = {37},
	pages = {93643--93670},
	booktitle = {Advances in Neural Information Processing Systems},
	publisher = {Curran Associates, Inc.},
	author = {Alfarano, Alberto and Charton, François and Hayat, Amaury},
	year = {2024},
}

@unpublished{lample_deep_2019,
	title = {Deep Learning for Symbolic Mathematics},
	note = {http://arxiv.org/abs/1912.01412},
	author = {Lample, Guillaume and Charton, François},
        year ={2019}
}

@misc{jaillet2025onlineschedulingllminference,
      title={Online Scheduling for LLM Inference with KV Cache Constraints},
      author={Patrick Jaillet and Jiashuo Jiang and Konstantina Mellou and Marco Molinaro and Chara Podimata and Zijie Zhou},
      year={2025},
      eprint={2502.07115},
      archivePrefix={arXiv},
      primaryClass={cs.LG},
      url={https://arxiv.org/abs/2502.07115},
}

@misc{liang_llm_2025,
	location = {Rochester, {NY}},
	title = {{LLM} for Large-Scale Optimization Model Auto-Formulation: A Lightweight Few-Shot Learning Approach},
	url = {https://papers.ssrn.com/abstract=5329027},
	doi = {10.2139/ssrn.5329027},
	shorttitle = {{LLM} for Large-Scale Optimization Model Auto-Formulation},
	number = {5329027},
	publisher = {Social Science Research Network},
	author = {Liang, Kuo and Lu, Yuhang and Mao, Jianming and Sun, Shuyi and Yang, Chunwei and Zeng, Congcong and Jin, Xiao and Qin, Hanzhang and Zhu, Ruihao and Teo, Chung-Piaw},
	urldate = {2025-07-07},
	date = {2025-06-28},
        year={2025},
	langid = {english},
}

@misc{ao2025optimizingllminferencefluidguided,
      title={Optimizing LLM Inference: Fluid-Guided Online Scheduling with Memory Constraints},
      author={Ruicheng Ao and Gan Luo and David Simchi-Levi and Xinshang Wang},
      year={2025},
      eprint={2504.11320},
      archivePrefix={arXiv},
      primaryClass={cs.LG},
      url={https://arxiv.org/abs/2504.11320},
}

@article{chen_manager_2025,
  title={A manager and an AI walk into a bar: does ChatGPT make biased decisions like we do?},
  author={Chen, Yang and Kirshner, Samuel N and Ovchinnikov, Anton and Andiappan, Meena and Jenkin, Tracy},
  journal={Manufacturing \& Service Operations Management},
  year={2025},
  volume = {27},
  pages = {354--368},
  number = {2}
}

@article{wang_large_2024,
	title = {Large Language Models for Market Research: A Data-augmentation Approach},
	author = {Wang, Mengxin and Zhang, Dennis and Zhang, Heng},
    year = {2026},
	note={Forthcoming},
    journal={Marketing Science }
}

@unpublished{yang2024optibench,
    title={OptiBench Meets ReSocratic: Measure and Improve LLMs for Optimization Modeling},
    author={Zhicheng Yang and Yiwei Wang and Yinya Huang and Zhijiang Guo and Wei Shi and Xiongwei Han and Liang Feng and Linqi Song and Xiaodan Liang and Jing Tang},
    year={2024},
    note={arXiv preprint arXiv:2407.09887}
}

@unpublished{li2025throughputoptimal,
    title={Throughput-Optimal Scheduling Algorithms for LLM Inference and AI Agents},
    author={Yueying Li and Jim Dai and Tianyi Peng},
    year={2025},
    note={arXiv preprint arXiv:2504.07347}
}

@article{ye2025lola,
  title={Lola: LLM-assisted online learning algorithm for content experiments},
  author={Ye, Zikun and Yoganarasimhan, Hema and Zheng, Yufeng},
  journal={Marketing Science},
  year={2025},
  volume={44},
  number={5},
  pages={995-1016}
}

@misc{li2025llmgeneratedpersonapromise,
      title={LLM Generated Persona is a Promise with a Catch},
      author={Ang Li and Haozhe Chen and Hongseok Namkoong and Tianyi Peng},
      year={2025},
      eprint={2503.16527},
      archivePrefix={arXiv},
      primaryClass={cs.CL},
      url={https://arxiv.org/abs/2503.16527},
}

@article{ouyang2022instructgpt,
  title  = {Training Language Models to Follow Instructions with Human Feedback},
  author = {Ouyang, Long and Wu, Jeff and Jiang, Xu and Almeida, Diogo and Wainwright, Carroll~L. and Mishkin, Pamela and Zhang, Chong and Agarwal, Sandhini and Slama, Katarina and Ray, Alex and others},
  journal = {arXiv:2203.02155},
  year   = {2022}
}

@article{phan2025humanity,
  title={Humanity's last exam},
  author={Phan, Long and others},
  journal={arXiv preprint arXiv:2501.14249},
  year={2025}
}

@InProceedings{ahmaditeshnizi2023optimus,
  title = 	 {{O}pti{MUS}: Scalable Optimization Modeling with ({MI}){LP} Solvers and Large Language Models},
  author =       {Ahmaditeshnizi, Ali and Gao, Wenzhi and Udell, Madeleine},
  booktitle = 	 {Proceedings of the 41st International Conference on Machine Learning},
  pages = 	 {577--596},
  year = 	 {2024},
  volume = 	 {235},
  series = 	 {Proceedings of Machine Learning Research},
  month = 	 {21--27 Jul},
  publisher =    {PMLR},
}

@book{powell2007approximate,
  title={Approximate Dynamic Programming: Solving the Curses of Dimensionality},
  author={Powell, Warren B},
  year={2011},
  edition={Second},
  publisher={John Wiley \& Sons, Inc.},
  address={Hoboken, NJ}
}

@book{puterman2005markov,
  title={Markov Decision Processes: Discrete Stochastic Dynamic Programming},
  author={Puterman, Martin L},
  year={2005},
  publisher={John Wiley \& Sons, Inc.},
  address={Hoboken, NJ}
}

@article{rafailov2023dpo,
  title  = {Direct Preference Optimization: Your Language Model Is Secretly a Reward Model},
  author = {Rafailov, Rafael and Sharma, Archit and Mitchell, Eric and Ermon, Stefano and Manning, Christopher~D. and Finn, Chelsea},
  journal = {arXiv preprint arXiv:2305.18290},
  year   = {2023}
}

@unpublished{shao2024deepseekmath,
  title  = {DeepSeekMath: Pushing the Limits of Mathematical Reasoning in Open Language Models},
  author = {Shao, Zhihong and Wang, Peiyi and Zhu, Qihao and Xu, Runxin and Song, Junxiao and Bi, Xiao and Zhang, Haowei and Zhang, Mingchuan and Li, Y.~K. and Wu, Y. and others},
  note = {arXiv preprint arXiv:2402.03300},
  year   = {2024}
}

@article{silver2016mastering,
  title={Mastering the game of Go with deep neural networks and tree search},
  author={Silver, David and Huang, Aja and Maddison, Chris J and Guez, Arthur and Sifre, Laurent and Van Den Driessche, George and Schrittwieser, Julian and Antonoglou, Ioannis and Panneershelvam, Veda and Lanctot, Marc and others},
  journal={Nature},
  volume={529},
  number={7587},
  pages={484--489},
  year={2016}
}

@book{sutton2018reinforcement,
  title={Reinforcement Learning: An Introduction},
  author={Sutton, Richard S. and Barto, Andrew G},
  publisher={The MIT Press},
  edition={Second},
  year={2018},
  address={Cambridge, Massachusetts and London, England}
}

@article{wang2022self,
  title={Self-instruct: Aligning language models with self-generated instructions},
  author={Wang, Yizhong and Kordi, Yeganeh and Mishra, Swaroop and Liu, Alisa and Smith, Noah A and Khashabi, Daniel and Hajishirzi, Hannaneh},
  journal={arXiv preprint arXiv:2212.10560},
  year={2022}
}

@misc{wei2023chainofthoughtpromptingelicitsreasoning,
      title={Chain-of-Thought Prompting Elicits Reasoning in Large Language Models},
      author={Jason Wei and Xuezhi Wang and Dale Schuurmans and Maarten Bosma and Brian Ichter and Fei Xia and Ed Chi and Quoc Le and Denny Zhou},
      year={2023},
      eprint={2201.11903},
      archivePrefix={arXiv},
      primaryClass={cs.CL},
      url={https://arxiv.org/abs/2201.11903},
}

@book{winston2004operations,
  title={Operations Research: Applications and Algorithm},
  author={Winston, Wayne L},
  year={2004},
  publisher={Thomson Learning, Inc.},
  edition={Fourth},
  address={Belmont, CA}
}

@article{zelikman2022star,
  title={Star: Bootstrapping reasoning with reasoning},
  author={Zelikman, Eric and Wu, Yuhuai and Mu, Jesse and Goodman, Noah},
  journal={Advances in Neural Information Processing Systems},
  volume={35},
  pages={15476--15488},
  year={2022}
}

@article{zheng2023secretsofrlhf,
  title  = {Secrets of RLHF in Large Language Models Part I: PPO},
  author = {Zheng, Rui and Dou, Shihan and Gao, Songyang and Hua, Yuan and Shen, Wei and Wang, Binghai and Liu, Yan and Jin, Senjie and Liu, Qin and Zhou, Yuhao and others},
  journal = {arXiv preprint arXiv:2307.04964},
  year   = {2023}
}

@article{BenAbacha2025MEDEC,
  title         = {MEDEC: A Benchmark for Medical Error Detection and Correction in Clinical Notes},
  author        = {Ben Abacha, Asma and Yim, Wen-wai and Fu, Yujuan and Sun, Zhaoyi and Yetisgen, Meliha and Xia, Fei and Lin, Thomas},
  journal       = {arXiv preprint arXiv:2412.19260},
  year          = {2025}
}

@article{Yin2026,
  title={Synthetic but Not Infinite: How Much LLM-Generated Data to Use in Market Research},
  author={Yin, Qichuan and Xin, Linwei},
  journal={Available at SSRN 6078686},
  year={2026}
}

@inproceedings{lin2004rouge,
    title = "{ROUGE}: A Package for Automatic Evaluation of Summaries",
    author = "Lin, Chin-Yew",
    booktitle = "Text Summarization Branches Out",
    month = jul,
    year = "2004",
    address = "Barcelona, Spain",
    publisher = "Association for Computational Linguistics",
    url = "https://aclanthology.org/W04-1013/",
    pages = "74--81"
}

@inproceedings{lin2004lcs,
    title = "Automatic Evaluation of Machine Translation Quality Using Longest Common Subsequence and Skip-Bigram Statistics",
    author = "Lin, Chin-Yew  and
      Och, Franz Josef",
    booktitle = "Proceedings of the 42nd Annual Meeting of the Association for Computational Linguistics ({ACL}-04)",
    month = jul,
    year = "2004",
    address = "Barcelona, Spain",
    url = "https://aclanthology.org/P04-1077/",
    doi = "10.3115/1218955.1219032",
    pages = "605--612"
}

@misc{brown2020fewshot,
      title={Language Models are Few-Shot Learners},
      author={Tom B. Brown and Benjamin Mann and Nick Ryder and Melanie Subbiah and Jared Kaplan and Prafulla Dhariwal and Arvind Neelakantan and Pranav Shyam and Girish Sastry and Amanda Askell and Sandhini Agarwal and Ariel Herbert-Voss and Gretchen Krueger and Tom Henighan and Rewon Child and Aditya Ramesh and Daniel M. Ziegler and Jeffrey Wu and Clemens Winter and Christopher Hesse and Mark Chen and Eric Sigler and Mateusz Litwin and Scott Gray and Benjamin Chess and Jack Clark and Christopher Berner and Sam McCandlish and Alec Radford and Ilya Sutskever and Dario Amodei},
      year={2020},
      eprint={2005.14165},
      archivePrefix={arXiv},
      primaryClass={cs.CL},
      url={https://arxiv.org/abs/2005.14165},
}

@inproceedings{deng2024contamination,
    title = "Investigating Data Contamination in Modern Benchmarks for Large Language Models",
    author = "Deng, Chunyuan  and
      Zhao, Yilun  and
      Tang, Xiangru  and
      Gerstein, Mark  and
      Cohan, Arman",
    editor = "Duh, Kevin  and
      Gomez, Helena  and
      Bethard, Steven",
    booktitle = "Proceedings of the 2024 Conference of the North American Chapter of the Association for Computational Linguistics: Human Language Technologies (Volume 1: Long Papers)",
    month = jun,
    year = "2024",
    address = "Mexico City, Mexico",
    publisher = "Association for Computational Linguistics",
    url = "https://aclanthology.org/2024.naacl-long.482/",
    doi = "10.18653/v1/2024.naacl-long.482",
    pages = "8706--8719"
}

@inproceedings{li2024contaminationreport,
    title = "An Open-Source Data Contamination Report for Large Language Models",
    author = "Li, Yucheng  and
      Guo, Yunhao  and
      Guerin, Frank  and
      Lin, Chenghua",
    editor = "Al-Onaizan, Yaser  and
      Bansal, Mohit  and
      Chen, Yun-Nung",
    booktitle = "Findings of the Association for Computational Linguistics: EMNLP 2024",
    month = nov,
    year = "2024",
    address = "Miami, Florida, USA",
    publisher = "Association for Computational Linguistics",
    url = "https://aclanthology.org/2024.findings-emnlp.30/",
    doi = "10.18653/v1/2024.findings-emnlp.30",
    pages = "528--541"
}

@inproceedings{lee2022deduplicating,
    title = "Deduplicating Training Data Makes Language Models Better",
    author = "Lee, Katherine  and
      Ippolito, Daphne  and
      Nystrom, Andrew  and
      Zhang, Chiyuan  and
      Eck, Douglas  and
      Callison-Burch, Chris  and
      Carlini, Nicholas",
    editor = "Muresan, Smaranda  and
      Nakov, Preslav  and
      Villavicencio, Aline",
    booktitle = "Proceedings of the 60th Annual Meeting of the Association for Computational Linguistics (Volume 1: Long Papers)",
    month = may,
    year = "2022",
    address = "Dublin, Ireland",
    publisher = "Association for Computational Linguistics",
    url = "https://aclanthology.org/2022.acl-long.577/",
    doi = "10.18653/v1/2022.acl-long.577",
    pages = "8424--8445"
}

@inproceedings{reimers2019sbert,
    title = "Sentence-{BERT}: Sentence Embeddings using {S}iamese {BERT}-Networks",
    author = "Reimers, Nils  and
      Gurevych, Iryna",
    editor = "Inui, Kentaro  and
      Jiang, Jing  and
      Ng, Vincent  and
      Wan, Xiaojun",
    booktitle = "Proceedings of the 2019 Conference on Empirical Methods in Natural Language Processing and the 9th International Joint Conference on Natural Language Processing (EMNLP-IJCNLP)",
    month = nov,
    year = "2019",
    address = "Hong Kong, China",
    publisher = "Association for Computational Linguistics",
    url = "https://aclanthology.org/D19-1410/",
    doi = "10.18653/v1/D19-1410",
    pages = "3982--3992"
}

@misc{abbas2023semdedup,
      title={SemDeDup: Data-efficient learning at web-scale through semantic deduplication},
      author={Amro Abbas and Kushal Tirumala and Dániel Simig and Surya Ganguli and Ari S. Morcos},
      year={2023},
      eprint={2303.09540},
      archivePrefix={arXiv},
      primaryClass={cs.LG},
      url={https://arxiv.org/abs/2303.09540},
}

@inproceedings{wang2020minilm,
author = {Wang, Wenhui and Wei, Furu and Dong, Li and Bao, Hangbo and Yang, Nan and Zhou, Ming},
title = {MINILM: deep self-attention distillation for task-agnostic compression of pre-trained transformers},
year = {2020},
isbn = {9781713829546},
publisher = {Curran Associates Inc.},
address = {Red Hook, NY, USA},
booktitle = {Proceedings of the 34th International Conference on Neural Information Processing Systems},
articleno = {485},
numpages = {13},
location = {Vancouver, BC, Canada},
series = {NIPS '20}
}

@misc{gpt5_para,
  author       = {Thompson, Alan D.},
  title        = {{GPT-5 (2025)}},
  year         = {2025},
  howpublished = {LifeArchitect.ai},
  url          = {https://lifearchitect.ai/gpt-5/},
  note         = {Accessed: 2026-03-25}
}

\ECSwitch

\ECHead{Electronic Companion for ``Auto-Formulating Dynamic Programming Problems with Large Language Models"}

\section{Examples}\label{appx-examples}
\subsection*{Example: Computer Science–Style DP Problem}
The following example is taken directly from a Leetcode problem\footnote{Link: \url{https://leetcode.com/problems/longest-valid-parentheses/description/}} categorized as ``Hard'' in their Dynamic Programming section. It illustrates a typical computer science-style DP problem: concise in structure, formally specified, and minimally ambiguous. In contrast, OR/OM problems often lack such clarity and involve implicit assumptions.

\begin{example}\label{eg-cs}
Given a string containing just the characters `(' and `)', return the length of the longest valid (well-formed) parentheses substring.
\end{example}
\subsection*{Example: Category, Scenario and Characteristics}\label{appx-scenario-example}
\begin{example}\label{eg-example}
\textbf{Category:} Thermal Plant Operational Scheduling. \\ \textbf{Scenario:} Scheduling operations for coal plants during peak demand in India. \\ \textbf{Characteristics:} Focuses on minimizing operational costs while meeting energy demands.
\end{example}
\section{Mathematical Formulation of DP Problems}\label{appx-dp_background}
We describe each component in Equation \eqref{eq-MF} in more detail. The planning horizon is denoted by $T$, which may be either finite ($T < \infty$) or infinite ($T = \infty$). $\mathcal{S}$ denotes the set of all possible states, and $\mathcal{A}$ denotes the set of all possible actions across all decision periods. At each decision period $t=1,\dots,T$, the system occupies a state $\boldsymbol{s}_t$ and we denote by $S_t \subseteq \mathcal{S}$ the subset of states that are reachable at decision period $t$. Similarly, when the decision-maker observes the system in state $\boldsymbol{s}_t \in S_t$, the decision-maker selects an action $\boldsymbol{a}_t$ from the set $A_{\boldsymbol{s}_t,t}$. So $\mathcal A = \bigcup_{t=1,\dots,T}\bigcup_{\boldsymbol{s}_t\in S_t} A_{\boldsymbol{s}_t,t} $. After a result of choosing action $\boldsymbol{a}_t\in A_{\boldsymbol{s}_t,t}$, the decision-maker receives an immediate reward $r_t(\boldsymbol{s}_t,\boldsymbol{a}_t)$, and the system state at the next decision period is described by the probability distribution $p_t(\cdot|\boldsymbol{s}_t,\boldsymbol{a}_t)$. Finally, $\gamma$ represents the discount factor.

A admissible policy $\pi = \{d_1, d_2, \ldots, d_{N-1}\}$ is a sequence of decision rules, where each decision rule $d_t$ specifies which action to take given the observed state $\boldsymbol{s}_t$, i.e., $\boldsymbol{a}_t = d_t(\boldsymbol{s}_t) \in A_{\boldsymbol{s}_t,t}$. In infinite-horizon settings, stationary policies, where $d_t = d$ for all $t$, are often of interest. The set of all admissible policies is denoted by $\Pi$

The sequence of events at each decision period $t$ occurs as follows. First, the decision-maker observes the current state $\boldsymbol{s}_t$. Then, according to the chosen policy, the decision-maker selects an action $\boldsymbol{a}_t = d_t(\boldsymbol{s}_t)$. After the action is selected, the random state transition is realized according to $p_t(\cdot|\boldsymbol{s}_t,\boldsymbol{a}_t)$, resulting in a new state $\boldsymbol{s}_{t+1}$. Concurrently, an immediate reward $r_t(\boldsymbol{s}_t,\boldsymbol{a}_t)$ is incurred.

For finite-horizon problems, the performance of a policy can be evaluated via the Bellman optimality equation. The value function $v_t(\boldsymbol{s}_t)$, which represents the maximum expected cumulative reward from time $t$ onward, satisfies the optimality equation
$$v_t(\boldsymbol{s}_t) = \sup_{\boldsymbol{a} \in A_{\boldsymbol{s}_t,t}}\left[r_t(\boldsymbol{s}_t,\boldsymbol{a})+ \sum_{\boldsymbol{j}\in S_{t+1}}p_t(\boldsymbol{j}|\boldsymbol{s}_t,\boldsymbol{a})v_{t+1}(\boldsymbol{j}) \right]\quad \text{for $t = 0, 1, \ldots, T-1$},$$
with the boundary condition $v_T(\boldsymbol{s}_T)=r_T(\boldsymbol{s}_T)$, and $r_T(\boldsymbol{s}_T)$ is the terminal value. The goal of the DP formulation is to identify an optimal policy $\pi^*$ such that maximizes the initial-stage reward $v_1(\boldsymbol{s}_1)$ among the set of all admissible policies $\Pi$.

In infinite-horizon problems, we focus primarily on stationary policies. Accordingly, we slightly abuse notation by omitting the time index and working with action sets $A_{\boldsymbol{s}}$ and state space $S$ instead of $A_{\boldsymbol{s}_t,t}$ and $S_t$. In the discounted infinite-horizon setting with discount factor $\gamma \in (0,1)$, we seek a stationary value function $v$ that maximizes the expected discounted total reward, i.e.,
$$v(\boldsymbol{s}) = \sup_{\boldsymbol{a} \in A_{\boldsymbol{s}}} \left[r(\boldsymbol{s},\boldsymbol{a})+ \gamma\sum_{\boldsymbol{j}\in S}p(\boldsymbol{j}|\boldsymbol{s},\boldsymbol{a})v(\boldsymbol{j}) \right],$$
where $r(\boldsymbol{s},\boldsymbol{a})$ is the one-stage reward function.

For the long-run average infinite-horizon cost problem, the performance of a policy $\pi$ is evaluated by the long-run average cost:
$$
v^\pi(\boldsymbol{s}_1) = \limsup_{T \to \infty} \frac{1}{T} \, \mathbb{E}\left[ \sum_{t=1}^{T} r(\boldsymbol{s}_t^\pi,\boldsymbol{a}_t^\pi) \right].
$$
The objective in this setting is to determine an optimal policy $\pi^*$ that maximizes the average reward for a given initial state $s_1 \in \mathcal{S}$.

Deterministic DP problems arise as a special case of the above formulation when the state transition is deterministic. Specifically, for each state–action pair $(\boldsymbol{s}_t,\boldsymbol{a}_t)$, the next state is uniquely determined by a transition function $f_t$, i.e.,
$s_{t+1} = f_t(\boldsymbol{s}_t,\boldsymbol{a}_t)$.
Equivalently, the transition probability $p_t(\cdot \mid \boldsymbol{s}_t,\boldsymbol{a}_t)$ degenerates to a point mass, and expectations over next states are no longer required.

\section{Benchmark Details}
\subsection{Guidelines for Modifying Benchmark Textbook Questions}\label{appx-benchmark-guide}
(a) In the textbooks, some problems are first introduced as examples to illustrate modeling and then appear in exercises with specific parameter values or additional assumptions. We merged these versions, incorporating modifications to focus on the main problem and eliminate unnecessary modeling details provided as examples.
\\
(b) While optimal policy is sometimes more important then the optimal value in many DP problems, for benchmarking purposes we need clear evaluation criteria. Therefore, we modified problems that originally required non-scalar answers to focus on computing the optimal value, a scalar metric, rather than deriving the more complex optimal policy.
\\
(c) It is common in exercises to present different variants of the same problem with only slight parameter changes. In such cases, we retained versions where the modeling differed and removed those with only minor parameter adjustments.
\\
(d) There are some ambiguities in certain problems, particularly regarding the time sequence of events and the timing of cost computations. To resolve these ambiguities, we clarify the required time sequence.
\\
(e) A few problems may include a hint, typically regarding how to model the space and actions. For example, in Problem 4.33 in \cite{puterman2005markov}, the hint is written as follows: ``(\emph{Hint: Write a system equation that describes the system state at each review period as a function of the system state from the previous review period and the daily demand for each day between reviews. Note further that the two-day demand can be found by computing the convolution of the demand distribution with itself. Solving this problem involves writing a computer program.})" We retain any provided hints but will not add new hints to any problems.

\subsection{Benchmark Annotation and Labeling Details}\label{appx-benchmark-label}

\begin{figure}
\centering
\includegraphics[width=15cm]{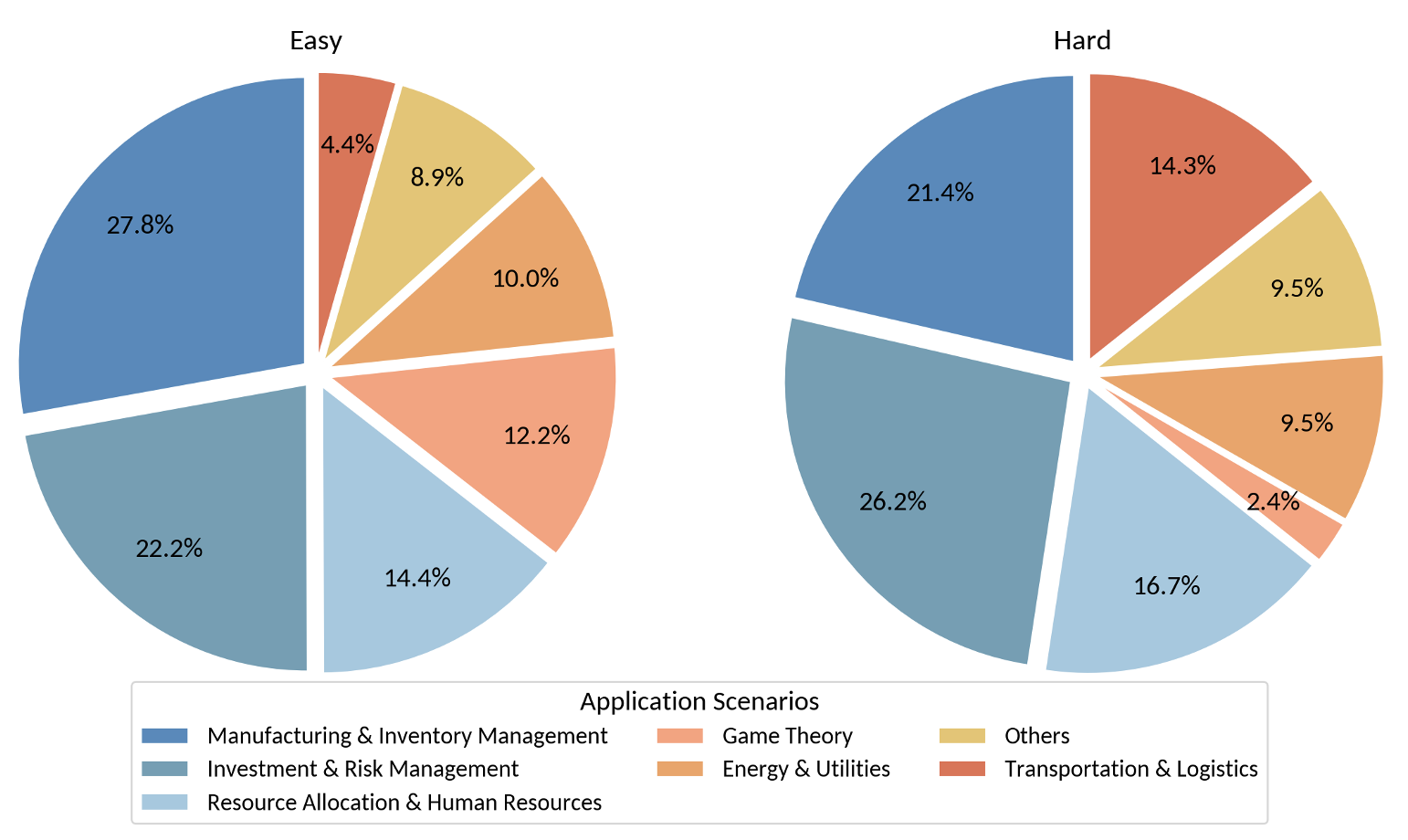}
\caption{Application scenario proportions: Easy vs. Hard.}
\label{fig-bench_industry}
\end{figure}

\begin{table}[htbp]
\centering
\caption{Proportion of $\DPBENCH$ problems with additional features.}
\begin{tabular}{p{8cm}cc}
\toprule
\textbf{Label} & \textbf{Easy(\%)} & \textbf{Hard(\%)} \\
\midrule
action-dependent transition probability & 22.22 & 66.67 \\
optimal stopping problem & 4.44 & 14.29 \\
truncation-required state space & 4.44 & 11.90 \\
time-dependent state space & 4.44 & 7.14 \\
continuous or non-integer state space & 2.22 & 7.14 \\
\bottomrule
\end{tabular}
\label{tab-bench_label}
\end{table}

Figure \ref{fig-bench_industry} shows the distribution of problems across common application scenarios in our easy and hard benchmarks. Overall, the two sets share similar top-level coverage. Both include over one-fourth of problems in manufacturing and inventory management and another quarter in investment and risk management, reflecting the typical emphasis in OR/OM curricula and textbooks. The remaining categories show varying proportions, with small shifts in areas such as transportation and logistics, and game theory.

Furthermore, as shown in Table~\ref{tab-bench_label}, the hard benchmark contains a higher proportion of problems with structural labels indicating either unique modeling formulations or increased complexity, such as \textit{action-dependent transition probability}, \textit{optimal stopping problem}, \textit{truncation-required state space}, \textit{time-dependent state space}, and \textit{continuous or non-integer state space}.

Table~\ref{tab:accuracy_comparison} reports the performance of \texttt{GPT-4o} and $\DPLM$ on deterministic and stochastic finite-horizon problems (excluding infinite-horizon cases for a fair comparison). \texttt{GPT-4o} exhibits a substantial performance gap between deterministic and stochastic settings, indicating that stochastic transitions introduce significantly greater difficulty. A similar gap remains for $\DPLM$, although the magnitude is smaller, primarily due to improved performance on stochastic problems.

\begin{table}[htbp]
\centering
\caption{Accuracy Comparison by Problem Type for Finite-Horizon Problems.}
\label{tab:accuracy_comparison}
\begin{tabular}{lcc}
\toprule
\textbf{Problem Type} & \textbf{DPLM(\%)} & \textbf{GPT-4o(\%)} \\
\midrule
Deterministic (Total=39) & 76.92 & 76.92 \\
Stochastic (Total=61)    & 50.82 & 19.67 \\
\bottomrule
\end{tabular}
\end{table}

\section{Data Generation Details}
\subsection{Implementation Details of Forward Generation}\label{appx-forward}
\paragraph{\textbf{Step 1: Generating the problem from the seed problem.}}
The new problem description $\tilde \PNL$ is generated by prompting an LLM to adapt the original $\PNL$ from a seed problem to a new scenario, as defined in Section \ref{subsec-scenario}. The prompt instructs the model to preserve core elements while making context-specific adjustments. As a result, the model formulation $\tilde \ModelF$ in the adapted problem $\tilde{\PNL}$ often differs from that of the seed. This approach is effective when the seed problem's domain aligns with the target scenario (e.g., adapting an inventory problem to a manufacturing context). In contrast, mapping to a conceptually distant domain, such as investment, is more likely to produce incoherent or ill-posed problem descriptions. To ensure novelty and correctness, we apply filter steps to remove $\PNL$s that are either too similar to their seed source or exhibit clear logical flaws.

\paragraph{\textbf{Step 2: Solution generation.}}
For each $\tilde{\PNL}$, we generate the solution $r=(\CoT, \Model, \Code)$ along with the final answer $y$ using the RAG-based method introduced in Section \ref{subsec-RAG}.

\paragraph{\textbf{Step 3: Solution refinement and filtering.}}
Since we cannot directly verify the correctness of DP solutions, we apply two techniques to improve reliability: code refinement and majority voting. Most DP implementations rely on simple loop-based logic rather than specialized solvers, but edge cases, such as unbounded state spaces or indexing errors, can still trigger execution failures. When such errors occur, we prompt the LLMs to reflect on common coding mistakes and regenerate its output. This error-aware refinement step not only improves execution success but also expands the candidate pool for majority voting. For each $\PNL$, we run the RAG-based solution generator five times, each with a distinct expert role (e.g., a DP expert or an OR student; see details in Section \ref{sec-generation}). A sample is retained only if the answer computed by its code matches the majority of reasonable answers across the five generations. This ensemble approach not only improve the reliability of the forward-generated training data samples but also provides a valuable signal for downstream filtering: the distribution of answers across agents serves as a proxy for confidence estimation and difficulty assessment of the data sample.

\subsection{Algorithms}\label{sec-generation}
This section presents the pseudocode for the forward and backward generation algorithms in Algorithms \ref{alg-for} and \ref{alg-back}, respectively. Both rely on a shared subroutine, SolveDP, provided in Algorithm \ref{alg-solve0}, to generate solutions. We also introduce the hyperparameters associated with these procedures.

For self-consistency, we generate solutions in both the forward and backward processes using $K = 5$ role-based agents, each simulating a different reasoning perspective: a DP expert (balanced across problem types), an OR professor (produces more instructive chain-of-thought reasoning), an OR student (writes more detailed models and code, mimicking homework-style solutions), a researcher in decision-making under uncertainty (well-suited for stochastic DPs), and an MDP specialist (effective on infinite-horizon problems), indexed by $k=1,\ldots,5$. These agents perform majority voting to enhance solution accuracy and robustness, correcting errors and encouraging diverse reasoning paths that converge on the correct answer.

We generate over 400 detailed scenarios, denoted by $\mathbb{S}$. Each seed sample from $\mathcal D^{seed}$ is paired with a scenario to create a new problem instance. To further expand the dataset, this pairing process is repeated $R$ times using a temperature of 0.7 to encourage diverse outputs.

In the backward generation process, when producing reflected CoT reasoning, we set an upper bound $P$ on the number of times the LLM compares its newly generated solution with the provided reference solution. This is motivated by two observations. First, inconsistencies often arise not from the LLM's inability to solve the problem, but from flaws in the generated problem statement.  Second, we empirically evaluate solution correctness on seed data under different numbers of reflection attempts. As shown in Table \ref{tab:try_distribution}, repeated reflections exhibit diminishing returns. To balance effectiveness and cost, we set $P = 6$.

\begin{table}[htbp]
\centering
\caption{Correctness distribution across reflection attempts in Reflect CoT.}
\label{tab:try_distribution}
\begin{tabular}{ccc}
\toprule
Reflection Attempt & Number of Instances & Fraction Correct \\
\midrule
1  & 48 & 52.75\% \\
2  & 7  & 7.69\%  \\
3  & 5  & 5.49\%  \\
4  & 3  & 3.30\%  \\
6  & 1  & 1.10\%  \\
8  & 1  & 1.10\%  \\
9  & 2  & 2.20\%  \\
$\geq$ 9 & 25 & 26.37\% \\
\bottomrule
\end{tabular}
\end{table}

\begin{algorithm}[t]
\caption{SolveDP}
\KwRequire{
  Problem description $\PNL$,
  Chain‐of‐thought $\CoT$,
  few-shot examples $\mathcal{E}_{fs} = \left\{(\PNL_{fs},\CoT_{fs},\Model_{fs},\Code_{fs},\mathit{labels}_{fs})\right\}$,
  role index $k$
}
\KwEnsure{$(\Model,\Code,y)$}
$\Model\leftarrow\text{promptM}^{(k)}(\PNL,\CoT,\mathcal{E}_{\mathit{fs}})$\\
$\Code\leftarrow\text{promptC}^{(k)}(\PNL,\Model,\mathcal{E}_{\mathit{fs}})$\\
$y\leftarrow\text{execute}(\Code)$
\label{alg-solve0}
\end{algorithm}

\setcounter{AlgoLine}{0}
\begin{algorithm}[t]
\caption{Forward Generation of Synthetic DP Data}
\label{alg-for}
\KwRequire{
  Seed set
    $\mathcal{D}^{\rm seed}=\{(\PNL_i,\ModelF_i,\Code_i)\}_{i=1}^n$,
  Seed examples
    $\mathcal{E}=\{(\PNL_e,\CoT_e,\Model_e,\Code_e,\mathit{labels}_e)\}$,\\
  scenarios $\mathbb S=\{\Scenario_j\}_{j=1}^m$,
  repeats $R$, $K$ roles
}
\KwEnsure{Forward‐generated dataset $\mathcal{D}^{\rm forward}$}

\ForEach{$(\PNL_i,\ModelF_i,\Code_i)\in\mathcal{D}^{\rm seed}$}{
  \ForEach{$\Scenario_j\in\mathbb S$}{
    \For{$r=1$ \KwTo $R$}{
      $\tilde\PNL_{ijr}\leftarrow\text{promptForwardP}(\PNL_i,\Scenario_j)$\\
      \If{$\tilde\PNL_{ijr}$ passes similarity and validity checks}{
      $\mathit{labels}\leftarrow\text{promptLabelAssign}(\tilde\PNL_{ijr})$\\
$\mathcal{E}_{{ijr}}\leftarrow\text{ExamplesRetrieve}(\tilde \PNL_{ijr},\mathcal{E},\mathit{labels})$\\
        \For{role $k=1$ \KwTo $K$}{

$\CoT_{ijr}^{(k)}\leftarrow\text{promptCoT}^{(k)}(\tilde\PNL_{ijr},\mathcal{E}_{{ijr}})$\\
          $(\Model_{ijr}^{(k)},\Code_{ijr}^{(k)},y_{ijr}^{(k)})\leftarrow
            \mathrm{SolveDP}(\tilde\PNL_{ijr},\CoT_{ijr}^{(k)},\mathcal{E}_{{ijr}},\text{role } k)$\\
            \If{$\Code_{ijr}^{(k)}$ executable with error}{
  $\Code_{ijr}^{(k)}\leftarrow\text{promptRefineCode}(\PNL_{ijr}^{(k)},\Code_{ijr}^{(k)},\text{Code Error})$\\
   $y_{ijr}^{(k)}\leftarrow\text{execute}(\Code_{ijr}^{(k)})$
}

        }
        $y_{ijr}^* \leftarrow \mathrm{MajorityVote}(\{y_{ijr}^{(k)}\}_{k=1}^K)$\\
        \If{$y_{ijr}^*$ is valid}{
          \For{$k=1$ \KwTo $K$}{
            \If{$y_{ijr}^{(k)} = y_{ijr}^*$}{
              add $(\tilde\PNL_{ijr},\CoT_{ijr}^{(k)},\Model_{ijr}^{(k)},\Code_{ijr}^{(k)},y_{ijr}^{(k)})$
                  to $\mathcal{D}^{\rm forward}$
            }

          }
        }
      }
    }
  }
}
\end{algorithm}

\setcounter{AlgoLine}{0}
\begin{algorithm*}
\caption{Backward Generation of Synthetic DP Data}
\label{alg-back}
\KwRequire{
  Seed set
    $\mathcal{D}^{\rm seed}=\{(\PNL_i,\ModelF_i,\Code_i)\}_{i=1}^n$,
  Seed examples
    $\mathcal{E}=\{(\PNL_e,\CoT_e,\Model_e,\Code_e,\mathit{labels}_e)\}$,
  scenarios $\mathbb S=\{\Scenario_j\}_{j=1}^m$,
  repeats $R$, $K$ roles, $P$ reflection attempts
}
\KwEnsure{Backward-generated dataset $\mathcal{D}^{\rm backward}$}

\ForEach{$(\PNL_i,\ModelF_i,\Code_i)\in\mathcal{D}^{\rm seed}$}{
  \ForEach{$\Scenario_j\in\mathbb S$}{
    \For{$r=1$ \KwTo $R$}{

      $\tilde\Code_{ijr}\leftarrow\text{Perturb}(\Code_i)$;
      $y^*_{ijr}\leftarrow\text{execute}(\tilde\Code_{ijr})$\\
      \If{$y^*_{ijr}$ is valid}{

        $\tilde\PNL_{ijr}\leftarrow\text{promptBackwardP}(\tilde\Code_{ijr},\Scenario_j)$\\
        \If{$\tilde{\PNL}_{ijr}$ passes similarity and validity checks}{
          $\mathit{labels}\leftarrow\text{promptLabelAssign}(\tilde\PNL_{ijr})$\\
$\mathcal{E}_{{ijr}}\leftarrow\text{ExamplesRetrieve}(\tilde \PNL_{ijr},\mathcal{E},\mathit{labels})$\\
        \For{role $k=1$ \KwTo $K$}{

$\CoT_{ijr}^{0,(k)}\leftarrow\text{promptCoT}^{(k)}(\tilde\PNL_{ijr},\mathcal{E}_{{ijr}})$\\
          $(\Model_{ijr}^{0,(k)},\Code_{ijr}^{0,(k)},y_{ijr}^{0,(k)})\leftarrow
            \mathrm{SolveDP}(\tilde\PNL_{ijr},\CoT_{ijr}^{0,(k)},\mathcal{E}_{ijr},\text{role } k)$

          \If{$y_{ijr}^{0,(k)}=y_{ijr}^{*}$}{
            add $(\tilde\PNL_{ijr},\CoT_{ijr}^{0,(k)},\Model_{ijr}^{0,(k)},\Code_{ijr}^{0,(k)},y_{ijr}^*)$
                to $\mathcal{D}^{\rm backward}$
          }
          \Else{
          \For{$p = 1$ \KwTo $P$}{
          $\CoT^{p,(k)}_{ijr} \leftarrow\text{promptCoTReflect}(\tilde\PNL_{ijr},C^{p-1,(k)}_{ijr},\tilde C_{ijr},\mathcal{E}_{ijr})$\\
$(\Model_{ijr}^{p,(k)},\Code_{ijr}^{p,(k)},y_{ijr}^{p,(k)})\leftarrow
            \mathrm{SolveDP}(\tilde\PNL_{ijr},\CoT_{ijr}^{p,(k)},\mathcal{E}_{ijr},\text{role } k)$\\
          \If{$y_{ijr}^{p,(k)}=y_{ijr}^{*}$}{
            add $(\tilde\PNL_{ijr},\CoT_{ijr}^{0,(k)},\Model_{ijr}^{0,(k)},\Code_{ijr}^{0,(k)},\dots, \CoT_{ijr}^{p,(k)},\Model_{ijr}^{p,(k)},\Code_{ijr}^{p,(k)},y_{ijr}^*)$
                to $\mathcal{D}^{\rm backward}$; \textbf{break}

        }
          }
          }

          }
        }
      }
    }
  }
}
\end{algorithm*}

\section{Data Contamination Analysis}\label{ec:contamination}

To assess potential data leakage between the human-curated seed set underlying our training pipeline and the evaluation benchmark ($\DPBENCH$), we conducted a decontamination analysis using both lexical and semantic metrics.

\paragraph{Lexical Overlap Analysis.}
We adopted a strict Longest Common Subsequence (LCS) heuristic to detect potential memorization or direct copying, which is closely related to ROUGE-L style sequence overlap metrics \citep{lin2004rouge,lin2004lcs}.
A benchmark problem $p_{test} \in \DPBENCH$ is flagged as contaminated if there exists any seed problem $p_{seed} \in \mathcal{D}^{seed}$ such that:
\begin{equation}
|LCS(p_{test}, p_{seed})| \ge 13 \quad \text{and} \quad \frac{|LCS(p_{test}, p_{seed})|}{\min\left(|p_{test}|, |p_{seed}|\right)} \ge 0.6,
\end{equation}
where $|s|$ denotes the token length of string $s$.
Our minimum-overlap design (13 tokens) is consistent with widely-used token-level n-gram / substring based decontamination heuristics in LLM evaluation (e.g., GPT-3's 13-gram collision rule and subsequent benchmark contamination studies) \citep{brown2020fewshot,deng2024contamination,li2024contaminationreport}.

\begin{table}[h]
\centering
\caption{Data contamination analysis results on $\DPBENCH$.}
\label{tab:leakage}
\begin{tabular}{lccc}
\toprule
\textbf{Subset} & \textbf{Total Problems} & \textbf{Flagged Overlaps} & \textbf{Contamination Rate} \\
\midrule
Easy & 90 & 6 & 6.7\% \\
Hard & 42 & 0 & 0.0\% \\
\textbf{Total} & \textbf{132} & \textbf{6} & \textbf{4.5\%} \\
\bottomrule
\end{tabular}
\end{table}

As shown in Table \ref{tab:leakage}, only 6 out of 132 problems were flagged.
This low flagged rate is comparable to the train--test near-duplicate overlap observed in standard web-scale NLP datasets \citep{lee2022deduplicating}.

\paragraph{Semantic Similarity Analysis.}
We computed semantic similarity using sentence embeddings from Sentence-BERT, where cosine similarity in the embedding space is designed to reflect semantic closeness \citep{reimers2019sbert}.
Embedding-based similarity is also a standard tool for semantic near-duplicate detection and semantic deduplication at scale \citep{abbas2023semdedup}.
We used \texttt{all-MiniLM-L6-v2}, a lightweight encoder based on the MiniLM family \citep{wang2020minilm}.
The average cosine similarity score (0.6062) indicates topical alignment (expected for in-domain synthetic training) while remaining far from the high-similarity regime typically associated with semantic duplicates \citep{abbas2023semdedup,reimers2019sbert}.

\section{Visualizing Data Distributions with t-SNE}\label{app:tsne}
\begin{figure}
    \centering
    \includegraphics[width=0.6\textwidth]{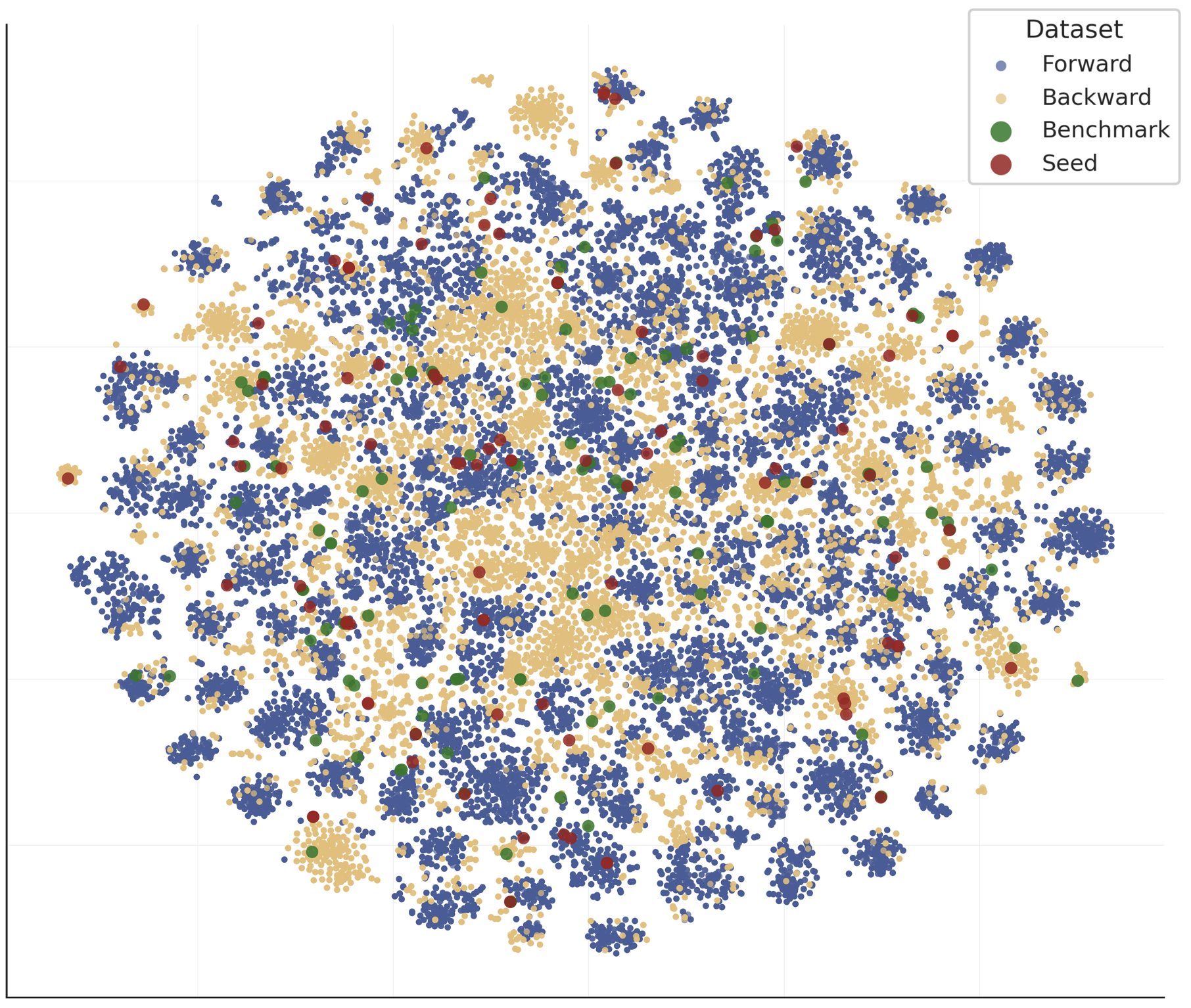}
    \caption{t-SNE visualization of the problem description distributions across datasets.}
    \label{fig:tsne_appendix}
\end{figure}

To illustrate the semantic relationships among the synthetic data generated by our $\RD$ framework, the seed data, and the benchmark, we conduct a t-SNE (t-distributed Stochastic Neighbor Embedding) visualization, as shown in Figure~\ref{fig:tsne_appendix}. Each problem description is embedded into a high-dimensional vector using a pre-trained language model and then projected into two dimensions using t-SNE, which places semantically similar problems closer together.

The visualization confirms that the synthetic data cloud (forward and backward points) effectively envelops the original seed and evaluation benchmark points. This demonstrates that our generated data is highly relevant to the core tasks and successfully spans their semantic space.

A key insight from the plot is the differing distributions of the two synthetic generation methods. The backward-generated data (orange) tends to form denser clusters as these problems, though varied in parameters, are constrained by the fixed model structures of the original seeds. In contrast, the forward-generated data (blue) is more dispersed, filling the space between clusters and extending toward the periphery. This pattern highlights the forward method's strength in creating more new problem formulations beyond the initial seeds, thereby enhancing the overall diversity of the training set. This visual evidence reinforces the value of our dual-generation approach: backward generation ensures fidelity, while forward generation drives diversity.

\section{Base Model Selection}\label{appx:base-model}
Before committing to a full training run, we conduct a small-scale study to identify a strong yet computationally affordable starting checkpoint. Given our hardware budget and the need for rapid iteration, we limit the search to open-source models in the 7–10B parameter range. Specifically, we compare \texttt{Gemma-2-9B-It}, \texttt{Llama-3.1-8B-Instruct}, and \texttt{Qwen-2.5-7B-Instruct}. Each candidate is evaluated {(i)} in its raw, pre-trained form, and {(ii)} after two epochs of SFT on our 113K synthetic DP trajectories.\footnote{The training hyperparameters mirrored those reported in Section~\ref{subsec:sft}.} Performance is measured by the micro-average pass@1 accuracy on $\DPBENCH$. As summarized in Table~\ref{tab:pilot}, \texttt{Qwen-2.5-7B-Instruct} emerges as the best performer in both settings, and we adopt it as the backbone for all subsequent experiments.

\begin{table}
  \centering
  \caption{Preliminary screening of candidate foundation models. Values report micro-average pass@1 (\%) on $\DPBENCH$. For each metric, the highest value is highlighted in bold, and the second-highest is underlined.}
  \label{tab:pilot}
  \begin{tabular}{lcrrr}
    \toprule
    \textbf{Model} & \textbf{Parameters} & \textbf{Base} & \textbf{+SFT} & $\boldsymbol{\Delta}$ \\
    \midrule
    Gemma-2-9B-It         & 9B & {\underline{5.3}} & 27.2 & {21.9} \\
    Llama-3.1-8B-Instruct & 8B & {\underline{5.3}} & \underline{28.8} & {\underline{23.5}} \\
    Qwen-2.5-7B-Instruct  & 7B & \textbf{{9.1}} & \textbf{{34.1}} & \textbf{{25.0}} \\
    \bottomrule
  \end{tabular}
\end{table}

\section{ORLM Baseline on $\DPBENCH$}\label{sec-orlm}
A natural question arises: Can a general optimization-focused LLM, such as ORLM \citep{huang2024orlm}, be directly repurposed to achieve strong performance on DP auto-formulation tasks? To explore this, we evaluate the recently released ORLM (i.e., the \texttt{ORLM-Llama-3-8B} version), one of the best-performing vertically trained models for optimization auto-formulation, on $\DPBENCH$.

Despite its impressive performance on LP and MIP tasks as reported by \citet{huang2024orlm}, ORLM performs poorly on DP tasks. As shown in Table \ref{tab:orlm_vs_ours}, it achieves only 0.8\% pass@1 and 8.3\% pass@10 micro-average accuracy. This underperformance likely comes from ORLM’s training paradigm, which is heavily based on LP/MIP problems and encourages translating optimization tasks (including DP) into LP formulations to be solved by external solvers. However, this rigid pipeline results in performance that lags behind even general-purpose LLMs. The sharp drop highlights a critical gap: techniques that excel in algebraic optimization do not seamlessly transfer to sequential, stochastic decision-making problems, where success depends on navigating a larger search space and correctly formulating Bellman recursions and state transitions.

\begin{table}[ht]
  \centering
  \caption{ORLM performance on $\DPBENCH$.}
  \label{tab:orlm_vs_ours}
  \begin{tabular}{lrrrrr}
    \toprule
    \textbf{Model} & \textbf{Parameters} & \textbf{Easy(\%)} & \textbf{Hard(\%)} & \textbf{Micro(\%)} & \textbf{Macro(\%)} \\
    \midrule
    ORLM-LLaMA-3-8B (pass@1)    & 8B   & 1.1  & 0.0  & 0.8  & 0.6 \\
    ORLM-LLaMA-3-8B (pass@10)   & 8B   & 11.1 & 2.4  & 8.3  & 6.8 \\
    \midrule
    DPLM-7B-SFT-GRPO (pass@1)         & 7B   & \textbf{66.7} & \textbf{40.5} & \textbf{58.3} & \textbf{53.6} \\
    \bottomrule
  \end{tabular}
\end{table}

\section{Additional Experimental Details}\label{ec:exp_details}

\subsection{Training Hyperparameters}\label{ec:hyper}

\textit{(1) SFT.} The model is fine-tuned for two epochs on 113K synthetic trajectories generated by $\RD$ using a batch size of 256, a learning rate of $1 \times 10^{-5}$, a weight decay of 0.1, and a total of two epochs.

\noindent\textit{(2) RL alignment.} We explore two alignment methods: GRPO and DPO. For GRPO, we generate $k=4$ rollouts per prompt from $\mathcal{D}_{\text{RL}}$, set the KL coefficient $\beta=0.01$, and conduct 5K training updates. For DPO, we train for three epochs on an offline preference dataset constructed by repeated sampling on prompts from $\mathcal{D}_{\text{RL}}$, with $\beta=0.1$. Subsequent ablation studies maintain these training parameters as consistently as possible across methods. For larger models (14B and 32B parameters), we reduce the learning rate to $2 \times 10^{-6}$ to achieve more stable training dynamics. Additional details are provided in their respective sections.

\subsection{Details on $\DPLM$ Performance and Failure Analysis}\label{sec-failure}

We complement pass@1 accuracy with executability rates and a finer-grained failure analysis to better disentangle formulation errors from implementation issues.

Table \ref{tab:failure_types} reports (i) executability rates across all problems, (ii) the share of incorrect answers attributable to coding errors, and (iii) the share of numerically correct answers that are logically incorrect upon manual inspection. Coding errors refer to cases where the generated program is syntactically invalid or fails to execute despite a correct model formulation; numerically correct but logically incorrect cases are those where the code executes successfully and returns the correct value, but the underlying formulation (e.g., state definition or transition) is incorrect.

\begin{table}[htbp]
\centering

\caption{Executability and Diagnostic Breakdown.}
\label{tab:failure_types}
\begin{tabular}{lrr}
\toprule
Failure Type / Metric & DPLM & GPT-4o \\
\midrule
Executability rate (all problems)          & 85.7\% & 89.4\% \\
\midrule

Coding errors (among incorrect answers) & 7.4\% & 2.6\% \\
Logically incorrect despite correct numeric answer (among correct answers) & 8.0\% & 2.4\% \\
\bottomrule
\end{tabular}
\end{table}

For both models, only a minority of incorrect answers are attributable to standalone coding errors (7.4\% for $\DPLM$ and 2.6\% for GPT-4o). Thus, most remaining incorrect answers are formulation-related rather than standalone coding bugs, often through downstream code errors induced by incorrect modeling.
Likewise, only a small fraction of numerically correct answers are logically incorrect upon manual inspection (8.0\% for $\DPLM$ and 2.4\% for GPT-4o). Because $\DPLM$ solves substantially more hard problems than \texttt{GPT-4o}, these conditional rates are affected by residual case difficulty and should be interpreted as within-model diagnostics rather than as direct cross-model reliability comparisons. Taken together, these results suggest that pass@1 accuracy largely reflects whether the model correctly formulates the dynamic programming problem rather than downstream implementation artifacts, supporting the use of end-to-end evaluation as a proxy for autoformulation ability in this setting.

To further characterize the limitations of $\DPLM$, we summarize common failure patterns observed on DP-Bench.

First, regarding coding errors, $\DPLM$ is more prone to pure syntax compared with GPT-4o, likely due to its relatively weaker base coding capability. In addition, $\DPLM$ sometimes struggles with implementing correctly formulated models. Common issues include: (i) handling infinite state/action spaces that require truncation. $\DPLM$ often introduces indexing errors or truncates the space too aggressively; (ii) difficulty handling non-integer state spaces; and (iii) selecting inappropriate algorithms for certain infinite-horizon settings (e.g., applying value iteration to problems with average-reward constraints).

In terms of modeling errors, the failure modes are more varied. One major type occurs when $\DPLM$ fails to recognize that a problem is infinite-horizon because this is not explicitly stated in the description. Another common issue arises in modeling transition dynamics.

Example~\ref{ex-4reviewcycle} illustrates a representative failure case from the hard subset of $\DPBENCH$. The problem is finite-horizon, but the transition structure is nontrivial. Inventory is reviewed every four days rather than daily, and orders arrive with a lead time of two days. Although $\DPLM$ correctly reasons that the state should be defined as the inventory level at the end of each four-day review period and even revises its initial reasoning, it fails to correctly compute the multi-day transition dynamics. Instead, it reverts to modeling the system on a daily review basis, leading to an incorrect formulation.

\begin{example}\label{ex-4reviewcycle}
(Canadian Tire inventory control) Daily demand for paint brushes at a particular store follows the demand distribution:
\begin{center}
\smallskip
\begin{tabular}{lccccc}
\hline
Demand & 0 & 1 & 2 & 3 & 4 \\
\hline
Probability & 0.70 & 0.15 & 0.10 & 0.04 & 0.01 \\
\hline
\end{tabular}
\smallskip
\end{center}
The stock level is reviewed in the evening every four days and when warranted an order is placed at the central warehouse to augment stock. Orders arrive two days later (a two day lead time) and are available to meet demand on the morning of the third day following the review. Demand not satisfed from stock on hand is never filled. Management imposes a penalty to account for this. Find the minimizes expected total ordering, holding and shortage costs under the assumption that the fixed cost for placing an order is \$$ 0.20$, the daily per unit holding cost is \$$ 0.01$ and the per unit penalty cost for unfilled orders is \$$ 0.50$. Daily costs are incurred after the demand for the day is realized. Assume a planning horizon of 84 days and no salvage value for each unit on hand at the end of this period. The initial inventory is 0.
\end{example}
\textbf{DPLM Output}

\begin{dplmoutput}
<reasoning>
To solve this inventory control problem for Canadian Tire with a finite horizon of 84 days, we consider the states as the inventory levels at the end of each four-day review period. The actions involve placing orders with the central warehouse to augment stock, considering the lead time of two days. The objective is to minimize the expected total costs, which include ordering costs, holding costs, and shortage penalties. The fixed cost for placing an order is \$0.20, the daily per unit holding cost is \$0.01, and the per unit penalty cost for unfilled orders is \$0.50. Daily costs are incurred after demand realization, and the demand follows a specific probability distribution: 0 brushes with probability 0.7, 1 brush with 0.15, 2 brushes with 0.1, 3 brushes with 0.04, and 4 brushes with 0.01. The reward function is defined by the costs associated with each state-action pair, where costs include holding costs for leftover inventory and penalty costs for unmet demand. State transitions are determined by the inventory carried over to the next period, which equals the current inventory plus ordered quantities minus demand, accounting for the two-day lead time. The initial state is an inventory of 0 brushes, and the terminal state is at the end of 84 days, with no salvage value for leftover inventory. We employ backward induction to solve the problem, starting from the last review period and moving backward to the first. At each stage, we calculate the expected costs for all possible state-action combinations by considering the demand distribution and cost components. By iterating this process, we determine the optimal order quantities that minimize the expected total costs over the 84-day horizon. Ultimately, we attain the minimum expected total cost by evaluating the value function at the initial state with zero inventory.
</reasoning>
Now let’s translate this logic into Python code:

import numpy as np

# Define the probability distribution of the demand
demand_probs = {0: 0.7, 1: 0.15, 2: 0.1, 3: 0.04, 4: 0.01}

# Define the convolution of the demand distribution for two days
two_day_demand_probs = {
    0: 0.49,
    1: 0.255,
    2: 0.17,
    3: 0.054,
    4: 0.01,
    5: 0.015,
    6: 0.004
}

# Parameters
days = 84
review_interval = 4
lead_time = 2
max_inventory = 20  # Assumed maximum inventory level for truncation
order_cost = 0.20
holding_cost = 0.01
shortage_cost = 0.50

# Initialize value function
V = np.zeros((review_interval + 1, max_inventory + 1))

# Backward induction
for t in range(review_interval, -1, -1):
    for s in range(max_inventory + 1):
        min_cost = float('inf')
        for a in range(max_inventory - s + 1):  # Order between 0 and max_inventory - s units
            order_cost_total = order_cost * a
            expected_cost = 0
            for d, prob in two_day_demand_probs.items():
                sales = min(s + a, d)
                holding_inventory = max(0, s + a - d)
                unsatisfied_demand = max(0, d - (s + a))
                daily_cost = holding_cost * holding_inventory + shortage_cost * unsatisfied_demand
                next_inventory = holding_inventory
                expected_cost += prob * (daily_cost + V[t, min(next_inventory, max_inventory)])
            total_cost = order_cost_total + expected_cost
            min_cost = min(min_cost, total_cost)
        V[t, s] = min_cost

# Output the minimum expected total cost starting from initial inventory of 0
print(V[0, 0])


Looking back, that wasn't the right approach. Let me correct it.

I made the following mistakes: First, I misunderstood the problem's state transition dynamics by incorrectly truncating the inventory level, which led to inaccurate calculations of future states. This happened because I did not fully grasp the importance of accurately modeling the inventory over the entire planning horizon. Second, I failed to correctly implement the demand convolution for the two-day lead time, which skewed the expected demand calculations. To avoid such errors in the future, I should ensure a thorough understanding of the problem's constraints and dynamics, visualize state transitions, and verify the correctness of probability distributions before coding.

To solve this inventory control problem with a finite horizon of 84 days, we consider states as the inventory levels at the end of each four-day review period. The actions are the order quantities placed to augment stock, constrained by the maximum inventory capacity. Our objective is to minimize the expected total costs, which include ordering costs, holding costs, and shortage penalties. The ordering cost is a fixed \$0.20 per order, the holding cost is \$0.01 per unit per day, and the shortage penalty is \$0.50 per unit of unmet demand. The daily demand follows a specified probability distribution, and the two-day demand is computed by convolving the one-day demand distribution with itself, reflecting the cumulative probability of demands over two days. State transitions are determined by the inventory carried over to the next review period, calculated as the current inventory plus ordered quantities minus realized demand, adjusted for the two-day lead time. The boundary conditions include an initial inventory of 0 units and a terminal state at the end of 84 days with no salvage value for leftover inventory. We employ backward induction to solve the problem, iterating from the last review period back to the first. At each stage, we compute the expected costs for all possible state-action combinations by considering the demand probabilities and cost components. By iterating this process, we determine the optimal order quantities that minimize the expected total costs over the planning horizon. Ultimately, we attain the minimum expected total cost starting from an initial inventory of 0, ensuring efficient inventory management throughout the 84-day period.Now, based on the above reasoning, here is the Python implementation:
```python
import numpy as np

# Define the demand probabilities
demand_prob = {0: 0.7, 1: 0.15, 2: 0.1, 3: 0.04, 4: 0.01}
demands = list(demand_prob.keys())
probabilities = list(demand_prob.values())

# Parameters
days = 84
order_cost = 0.20
holding_cost = 0.01
penalty_cost = 0.50
max_inventory = 50  # Assumption for maximum inventory level
max_order = 10  # Assumption for maximum order quantity

# Initialize cost-to-go function
V = np.zeros((days + 1, max_inventory + 1))

# Boundary condition: No future costs at the end of the planning horizon
# V[days, s] = 0 for all s (already initialized to zero by np.zeros)

# Perform backward induction
for t in range(days - 1, -1, -1):
    for s in range(max_inventory + 1):
        min_cost = float('inf')
        for a in range(max_order + 1):  # Order between 0 and max_order units
            if s + a > max_inventory:
                continue  # Skip actions leading to inventory beyond max_inventory
            order_cost_total = order_cost if a > 0 else 0
            expected_cost = order_cost_total
            for d, prob in demand_prob.items():
                sales = min(s + a, d)
                unsatisfied_demand = max(d - (s + a), 0)
                holding_inventory = max(s + a - d, 0)
                penalty = unsatisfied_demand * penalty_cost
                holding = holding_inventory * holding_cost
                next_inventory = holding_inventory
                future_cost = V[t + 1, min(next_inventory, max_inventory)]  # Avoid index out of bounds
                period_cost = penalty + holding + future_cost
                expected_cost += prob * period_cost
            min_cost = min(min_cost, expected_cost)
        V[t, s] = min_cost

# Initial inventory is zero
initial_inventory = 0
min_expected_cost = V[0, initial_inventory]

# Print the final result
print(min_expected_cost)
```
\end{dplmoutput}

\subsection{Sensitivity Analysis of GRPO Reward Configuration}\label{ec:grpo_reward_sensitivity}

To justify our choice of the reward decomposition $r(x, \hat a, a^*) = r_{\text{format}}(x, \hat a) + r_{\text{answer}}(\hat a, a^*)$, we conduct a sensitivity analysis by varying the weights assigned to the format reward ($w_f$) and the answer correctness reward ($w_a$), such that $w_f + w_a = 1$. The results are summarized in Table~\ref{tab:grpo_reward_sensitivity}.

We observe a distinct trade-off in performance as the format reward weight $w_f$ increases.

\textit{(1) Sparse Reward Issue ($w_f=0$):} When relying solely on answer correctness ($w_f=0, w_a=1$), the model achieves a Micro-average accuracy of 41.7\%, which is a significant improvement over the SFT baseline (34.1\%) but lower than our optimal setting. This suggests that without the intermediate "partial credit" for generating executable code, the reward signal is too sparse, making it harder for the policy to navigate the early stages of optimization.

\textit{(2) Reward Hacking ($w_f \ge 0.5$):} Conversely, setting the format reward too high (e.g., $w_f=0.5$) degrades performance to 48.5\%. In this regime, the optimization landscape encourages "reward hacking," where the model prioritizes generating simple, executable code to secure the easy format reward, rather than engaging in the complex reasoning required to solve the problem correctly.

\textit{(3) Optimal Balance ($w_f=0.2$):} The configuration ($w_f=0.2, w_a=0.8$) achieves the highest performance across all metrics. This setting provides a sufficient dense signal to guide the model toward valid program generation while ensuring that the dominant training signal remains tied to semantic correctness.

\begin{table}[htb]
  \centering\small
  \caption{Sensitivity analysis of GRPO performance under different reward decompositions between formatting validity ($w_f$) and answer correctness ($w_a$). All experiments use the DPLM-7B-SFT checkpoint as the starting policy.}
  \label{tab:grpo_reward_sensitivity}
  \begin{tabular}{@{}p{6cm}rrrr@{}}
    \toprule
    \textbf{Configuration} & \textbf{Easy(\%)} & \textbf{Hard(\%)} & \textbf{Micro(\%)} & \textbf{Macro(\%)} \\
    \midrule
    Baseline (SFT Only) & 40.0 & 21.4 & 34.1 & 30.7 \\
    \midrule
    $w_f=0.0, \quad w_a=1.0$ & 46.7 & 31.0 & 41.7 & 38.8 \\
    $w_f=0.05, \; w_a=0.95$ & 61.1 & 33.3 & 52.3 & 47.2 \\
    $w_f=0.2, \quad w_a=0.8$ \textbf{(Ours)} & \textbf{66.7} & \textbf{40.5} & \textbf{58.3} & \textbf{53.6} \\
    $w_f=0.3, \quad w_a=0.7$ & 62.2 & 35.7 & 53.8 & 49.0 \\
    $w_f=0.5, \quad w_a=0.5$ & 53.3 & 38.1 & 48.5 & 45.7 \\
    \bottomrule
  \end{tabular}
\end{table}

\subsection{Baseline Model Catalog}\label{ec:baseline}
We compare $\DPLM$ against the following SOTA language models.\\
\noindent\textit{(1) Open-source baselines.}
(a) DeepSeek Series: \texttt{DeepSeek-R1 671B}
and \texttt{DeepSeek-V3 671B}.
(b) Qwen Series: models ranging from 0.5B to 72B parameters.
(c) Other leading models: \texttt{Llama-3.1-8B-Instruct} and \texttt{Gemma-2-9B-It}.\\
\noindent\textit{(2) Closed-source baselines.}
(a) ChatGPT (GPT-4o) and (b) o1, a proprietary 300B-parameter reasoning model from OpenAI.

\subsection{Hardware and Decoding}\label{ec:hardware}

Both training and inference are conducted on a single node equipped with eight NVIDIA H100 GPUs (80GB each), interconnected via NVLink. During inference, we use nucleus sampling with top-\(p=0.95\), temperature \(T=0.7\), and a maximum generation length of 8,192 tokens. Because local deployment is not feasible for closed-source large models, we access these models remotely via the provider’s cloud-based API. To ensure a fair comparison, we match the decoding parameters for open-source models. However, since closed-source models run remotely, all other real-time engineering and system-level configurations follow the provider’s default settings.

\section{Model-Size Scaling}\label{ec:model_size_scaling}

Table~\ref{tab:size_scaling} reports the pass@1 accuracy of \texttt{Qwen-2.5} models ranging from 0.5B to 32B parameters, evaluated before and after two epochs of SFT on our 113K textbook–trajectory corpus. Overall, performance improves sharply up to 7B parameters, with more gradual gains beyond that point.

According to established scaling laws, model quality improves predictably when model size, dataset size, and compute are scaled in balance; if any one of these factors lags, it constrains the benefits of the others. Our results suggest that model capacity is the primary bottleneck below 7B parameters. However, between 7B and 32B, the limiting factor shifts to the volume of training data. While additional data can still improve performance in this regime, the model's ability to effectively leverage it becomes the limiting factor. Thus, a more effective path to further accuracy gains is to co-scale model size and training data in tandem.

\begin{table}[htb]
  \centering\small
  \caption{``Base" refers to the raw pre-trained model; ``\textsc{+SFT}" denotes two epochs of SFT (113K trajectories). $\Delta$ denotes the improvement from SFT.}
  \label{tab:size_scaling}

  \begin{tabular}{@{}c
                  rrr
                  rrr
                  rrr
                  rrr@{}}
    \toprule
    \multirow{2}{*}{\textbf{Parameters}}
        & \multicolumn{3}{c}{\textbf{Easy(\%)}}
        & \multicolumn{3}{c}{\textbf{Hard(\%)}}
        & \multicolumn{3}{c}{\textbf{Micro(\%)}}
        & \multicolumn{3}{c}{\textbf{Macro(\%)}} \\
    \cmidrule(lr){2-4}\cmidrule(lr){5-7}\cmidrule(lr){8-10}\cmidrule(lr){11-13}
        & Base & +SFT & {\(\Delta\)}
        & Base & +SFT & {\(\Delta\)}
        & Base & +SFT & {\(\Delta\)}
        & Base & +SFT & {\(\Delta\)}\\
    \midrule
    0.5 B & 0.0 & 8.9 & {+8.9}
          & 0.0 & 0.0 & {0.0}
          & 0.0 & 6.1 & {+6.1}
          & 0.0 & 4.5 & {+4.5} \\

    1.5 B & 3.3 & 15.6 & {+12.3}
          & 0.0 & 2.4  & {+2.4}
          & 2.3 & 11.4 & {+9.1}
          & 1.7 & 9.0 & {+7.3} \\

    3 B   & 4.4 & 25.6 & {+21.2}
          & 0.0 & 4.8  & {+4.8}
          & 3.0 & 18.9 & {+15.9}
          & 2.2 & 15.2 & {+13.0} \\

    7 B   & {11.1} & {40.0} & \textbf{{+28.9}}
          & {4.8}  & 21.4 & \textbf{{+16.6}}
          & {9.1}  & {34.1} & \textbf{{+25.0}}
          & {7.9}  & {30.7} & \textbf{{+22.8}} \\

    14 B  & 24.4 & 48.9 & {+24.5}
          & 9.5  & 23.8 & {+14.3}
          & 19.7 & 40.9 & {+21.2}
          & 17.0 & 36.4 & {+19.4} \\

    32 B  & {37.8} & 55.6 & {+17.8}
          & {21.4} & 28.6 & {+7.2}
          & {32.6} & 47.0 & {+14.4}
          & {29.6} & 42.1 & {+12.5} \\
    \bottomrule
  \end{tabular}
\end{table}

\section{Ablation Study on Forward-Backward Data Mixture Ratios}\label{appx-mixture-ablation}
To determine the optimal composition of the training dataset, we conducted an ablation study varying the proportion of forward-generated versus backward-generated data.

For these experiments, we fixed the total dataset size at 16,000 instances. To create each mixture subset, we performed independent random sampling without replacement from the full Forward and Backward data pools until the target counts were met. By re-sampling for each configuration rather than using a fixed nested subset, we mitigate potential selection bias associated with any specific group of instances, ensuring that the observed performance trends are attributable to the mixture ratios rather than the idiosyncrasies of a particular data sample.

It is important to note that while we balance by instance count, the two data sources differ in information density. Backward-generated instances, especially those containing Reflected CoT trajectories, typically possess higher token counts and represent "harder" verified reasoning paths. In contrast, Forward-generated instances contribute primarily to the semantic diversity of problem formulations. As shown in Table~\ref{tab:mixture_ablation}, the performance does not scale linearly with the ratio. Instead, we observe an empirical "sweet spot" in the 40\%--60\% range. This indicates that despite differences in token consumption and individual data value, a roughly balanced mix (by count) effectively allows the model to leverage the complementary strengths of diversity (Forward) and rigorous correctness (Backward).

\begin{table}[htbp]
  \centering
  \caption{Macro-Average Accuracy (\%) of DPLM (7B, 14B, 32B) under different ratios of Forward synthetic data.}
  \label{tab:mixture_ablation}

  \begin{tabular}{crrr|rrr|rrr}
    \toprule
    \multirow{2}{*}{\textbf{Forward(\%)}} & \multicolumn{3}{c|}{\textbf{7B}} & \multicolumn{3}{c|}{\textbf{14B}} & \multicolumn{3}{c}{\textbf{32B}} \\
    \cmidrule(lr){2-4} \cmidrule(lr){5-7} \cmidrule(lr){8-10}
     & Easy & Hard & Macro & Easy & Hard & Macro & Easy & Hard & Macro \\
    \midrule
    10\% & 28.9 & 11.9 & 20.4 & 35.6 & 16.7 & 26.1 & 48.9 & 21.4 & 35.2 \\
    20\% & 25.6 & 14.3 & 19.9 & 34.4 & 16.7 & 25.6 & 50.0 & 21.4 & 35.7 \\
    30\% & 27.8 & 16.7 & 22.2 & 33.3 & 21.4 & 27.4 & 52.2 & 23.8 & 38.0 \\
    40\% & 32.2 & 16.7 & 24.4 & 47.8 & 21.4 & 34.6 & 52.2 & 26.2 & \textbf{39.2} \\
    50\% & 34.4 & 16.7 & 25.6 & 45.6 & 23.8 & \textbf{34.7} & 53.3 & 23.8 & 38.6 \\
    60\% & 36.7 & 14.3 & \textbf{25.5} & 44.4 & 21.4 & 32.9 & 51.1 & 21.4 & 36.3 \\
    70\% & 27.8 & 16.7 & 22.2 & 42.2 & 16.7 & 29.4 & 53.3 & 23.8 & 38.6 \\
    80\% & 30.0 & 14.3 & 22.1 & 35.6 & 14.3 & 24.9 & 48.9 & 21.4 & 35.2 \\
    90\% & 26.7 & 14.3 & 20.5 & 36.7 & 16.7 & 26.7 & 46.7 & 19.0 & 32.9 \\
    \bottomrule
  \end{tabular}

\end{table}

\section{Inference Scaling Analysis}\label{ec:scaling}
Sampling multiple candidate solutions at inference time and selecting the best-performing one has proved effective for improving accuracy (e.g., \citealp{wei2023chainofthoughtpromptingelicitsreasoning,guo2025deepseek,huang2024orlm}). Understanding how accuracy scales with the number of generated samples helps practitioners allocate computational resources more efficiently.

\begin{figure}[ht]
      \centering
      \includegraphics[width=\linewidth]{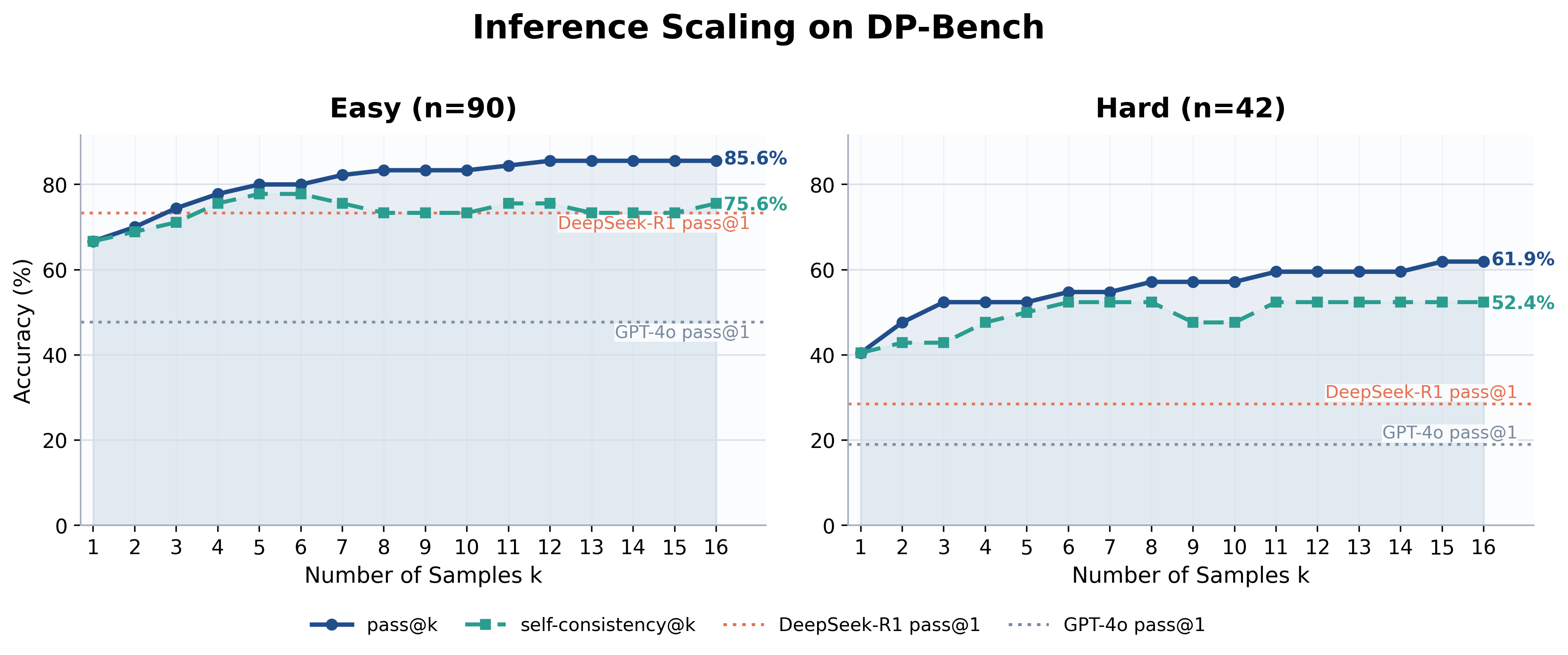}
        \caption{Test-time scaling for our $\DPLM$ model on $\DPBENCH$ Easy (left) and Hard (right): solid lines represent $\DPLM$ pass@k, dashed lines represent $\DPLM$ self-consistency@k, and horizontal dotted lines indicate the pass@1 accuracy of DeepSeek-R1 and GPT-4o for reference. The x-axis denotes the number of sampled completions $k$, and the y-axis reports accuracy.}
      \label{fig:test-scaling}
    \end{figure}

\paragraph{Experimental Setup.}
For each test instance, we generate $k\in\{1,...,16\}$ independent completions from $\DPLM$ using nucleus sampling  ($p\!=\!0.95$) and a moderate temperature ($T\!=\!0.9$). Each generated completion is executed, and failures are recorded as \textsc{null}. Figure \ref{fig:test-scaling} reports two evaluation metrics: pass@$k$, which measures whether at least one of the $k$ completions is correct, and self-consistency@$k$, which uses majority voting over the $k$ generated answers.

\paragraph{Accuracy Scaling on the Easy Problem Set.}

For the 90 easy problems, pass@$k$ increases from 66.7\% at $k=1$ to 85.6\% at $k=16$. Sampling four candidate solutions already captures 58.8\% of the total gain attainable at $k=16$, and the curve begins to flatten after roughly twelve samples. Self-consistency@$k$ also improves initially, but remains below pass@$k$ throughout most of the range and trails it by 10.0 percentage points at $k=16$. This suggests that even on Easy problems, execution-based selection benefits from diverse valid reasoning paths that majority voting does not fully exploit.

\paragraph{Accuracy Scaling on the Hard Problem Set.}

The 42 Hard problems exhibit a similarly strong but more gradual scaling pattern. Pass@$k$ rises from 40.5\% at $k=1$ to 61.9\% at $k=16$, with more than half of the total gain already realized by $k=4$. Even four samples are sufficient to comfortably surpass both DeepSeek-R1 and GPT-4o on the hard subset, and by $k=12$ the model reaches 59.5\% accuracy. In contrast, self-consistency@$k$ plateaus at 52.4\%, leaving a 9.5-point gap at $k=16$. This again indicates that Hard problems benefit from exploring diverse solution approaches, and that execution-based selection is crucial for fully leveraging this diversity.
\section{Additional Case Studies on Out-of-Domain Datasets}
\label{app:ood}
\subsection{Dataset Construction}
\label{app:ood_construction}

We construct two healthcare-based out-of-domain datasets, Healthcare-OOD1 and Healthcare-OOD2, to evaluate contextual and structural generalization of $\DPLM$.

\paragraph{Healthcare-OOD1.}
Healthcare-OOD1 is designed to isolate contextual generalization while keeping the underlying DP model structure unchanged. The dataset is generated using the backward generation pipeline described in Section~\ref{subsec-backward}.

Specifically, we start from problems in $\DPBENCH$, exclude instances that are highly domain-specific and unsuitable for healthcare adaptation, and perturb the problem data $\Data$ (e.g., transition probabilities, rewards, or horizon length). We then regenerate the problem descriptions under medical treatment contexts, including epidemic control, transplantation, and disease progression. Note that although $\DPBENCH$ contains five healthcare-themed problems, they are all related to resource allocation or scheduling (e.g., workforce assignment, capacity planning, inventory control), rather than clinical decision optimization such as treatment selection or patient-level intervention policies. Textbook DP problems rarely emphasize such clinical treatment contexts.

To ensure data quality, we retain only those instances for which at least one solution generator produces the correct output.

\paragraph{Healthcare-OOD2.}
Healthcare-OOD2 is constructed to evaluate performance on more realistic and structurally complex medical decision problems. The dataset is based on classical healthcare applications of Markov decision processes summarized in \citet{Schaefer2004}, Section 23.4.

Among the seven applications discussed, four are formulated as standard MDPs (epidemic control, kidney transplantation, spherocytosis treatment, and liver transplantation), while the remaining three involve POMDPs and are excluded. For each selected application, we construct problem instances following the descriptions in \citet{Schaefer2004} and their corresponding original papers.

The problem data $\Data$ are manually specified based on reported numerical experiments and domain knowledge. While these data may not be fully realistic, we observe that auto-formulation performance is largely insensitive to specific parameter values, as $\DPLM$ typically abstracts numerical inputs into symbolic variables during model construction and code implementation.

\subsection{Healthcare-OOD1: Context Transfer Analysis}\label{app:ood1}

In this section, we provide paired examples to illustrate DPLM’s performance under out-of-context but in-domain model transfer between $\DPBENCH$ and Healthcare-OOD1. Each pair includes (i) the original $\DPBENCH$ problem with DPLM’s output, and (ii) the corresponding healthcare-context adaptation with DPLM’s output. We present two types of cases: one in which $\DPLM$ correctly solves both the original and adapted problems, and another in which it fails on both, thereby highlighting consistent success and consistent failure patterns under contextual shifts.
\subsubsection{Case 1: Consistent Success}
\

In this case, Example~\ref{ex-case1b} is backward-generated from Example~\ref{ex-case1a} in $\DPBENCH$, under the liver transplantation category with the characteristic description: “Balances the timing of surgical intervention against patient health progression and transplant-related risks to improve long-term survival outcomes under uncertainty.”

The underlying model structure remains the same, with only parameter perturbations introduced during generation. Correspondingly, $\DPLM$’s outputs for the two problems are highly similar, reflecting consistent modeling and solution behavior across the original and context-adapted versions.
\paragraph{(A) Original $\DPBENCH$ Problem}
\begin{example}\label{ex-case1a}
A decision maker observes a discrete-time system which moves between states
$\{s_1, s_2, s_3, s_4\}$ according to the following transition probability matrix:
\[
P =
\begin{bmatrix}
0.3 & 0.4 & 0.2 & 0.1 \\
0.2 & 0.3 & 0.5 & 0.0 \\
0.1 & 0.0 & 0.8 & 0.1 \\
0.4 & 0.0 & 0.0 & 0.6
\end{bmatrix}.
\]
At each point in time, the decision maker may leave the system and receive a reward of \( R = 20 \) units, or alternatively remain in the system and receive a reward of \( r(s_i) \) units if the system occupies state \( s_i \). The reward function is defined as \( r(s_i) = i \). If the decision maker decides to remain in the system, its state at the next decision epoch is determined by \( P \). Assume a discount rate of \( \lambda = 0.9 \). The objective is to maximize the expected total discounted reward. Determine the discounted optimal value of state \( s_1 \). Round your answer to four decimal places.
\end{example}
\textbf{DPLM Output}

\begin{dplmoutput}
<reasoning>
To solve this infinite horizon Markov Decision Process (MDP) problem, we consider the states as \\( \\{s_1, s_2, s_3, s_4\\} \\), with the decision maker able to choose between leaving the system for a fixed reward or remaining to earn state-dependent rewards. The objective is to maximize the expected total discounted reward, where the reward for leaving is \\( R = 20 \\) and the reward for remaining in state \\( s_i \\) is \\( r(s_i) = i \\). The transition dynamics are governed by the matrix \\( P \\), which dictates the probability of moving from one state to another if the decision maker chooses to remain. The methodology involves value iteration, where we iteratively update the value function \\( V(s) \\) for each state \\( s \\) by calculating the maximum of the immediate reward and the expected future rewards discounted by \\( \\lambda = 0.9 \\). We initialize the value function with zeros and update it until convergence within a tolerance of \\( 1 \\times 10^{-6} \\). The result is the discounted optimal value of state \\( s_1 \\), which is computed and rounded to four decimal places for precision. This approach correctly applies the reward and transition mechanisms, ensuring the decision maker's optimal decision at each state is accurately captured.
</reasoning>
```python
import numpy as np

# Define parameters
states = [1, 2, 3, 4]  # State space
actions = ["Leave", "Remain"]  # Action space
R = 20  # Reward for leaving
lambda_ = 0.9  # Discount factor

# Transition probability matrix
P = np.array([
    [0.3, 0.4, 0.2, 0.1],
    [0.2, 0.3, 0.5, 0.0],
    [0.1, 0.0, 0.8, 0.1],
    [0.4, 0.0, 0.0, 0.6]
])

# Reward for remaining in each state
r = [1, 2, 3, 4]

# Initialize value function
V = np.zeros(len(states))

# Value iteration
epsilon = 1e-6  # Convergence threshold
while True:
    delta = 0
    V_new = np.copy(V)
    for s in range(len(states)):
        # Calculate the value of leaving
        value_leave = R

        # Calculate the value of remaining
        value_remain = r[s] + lambda_ * np.dot(P[s], V)

        # Update the value function with the maximum of the two options
        V_new[s] = max(value_leave, value_remain)

    # Calculate the maximum change in the value function
    delta = np.max(np.abs(V_new - V))
    V = V_new

    # Check for convergence
    if delta < epsilon:
        break

# Print the optimal value of state s_1
print(round(V[0], 4))
```
\end{dplmoutput}
\paragraph{(B) Healthcare-OOD1 Adaptation}
\begin{example}\label{ex-case1b}
A patient with severe liver disease is being considered for a living-donor liver transplant. The patient’s health is classified into four states (1 through 4), with state 1 being the healthiest and state 4 the most critical. At the start of each month, the physician must decide whether to perform the transplant immediately or to wait one more month. If the transplant is performed, the patient receives a fixed long‑term survival benefit equivalent to 25 quality‑adjusted life years (QALYs). If the physician waits, the patient receives a near‑term QALY equal to the current health state index (i.e., in state \(i\) the patient accrues \(i\) QALYs for that month) and then transitions to a new health state at the beginning of the next month according to the following probability matrix:
\[
P = \begin{bmatrix}
0.2 & 0.5 & 0.2 & 0.1\\
0.1 & 0.4 & 0.4 & 0.1\\
0.0 & 0.1 & 0.7 & 0.2\\
0.3 & 0.1 & 0.1 & 0.5
\end{bmatrix}.
\]
Here, \(P_{ij}\) is the probability of moving from state \(i\) to state \(j\) in one month. Future QALYs are discounted by a factor of \(0.85\) per month. The physician’s objective is to maximize the total expected discounted QALYs from the current time onward.

What is the maximal total expected discounted QALYs for a patient who starts in health state 1?
\end{example}
\textbf{DPLM Output}
\begin{dplmoutput}
<reasoning>
To solve this liver transplant decision problem with stochastic health state transitions over an infinite horizon, we consider the states as the patient's health conditions, ranging from 1 (healthiest) to 4 (most critical). The actions available to the physician at each state are either to perform the transplant immediately, yielding a fixed reward of 25 QALYs, or to wait, allowing the patient to accrue QALYs equal to their current health state index (ranging from 1 to 4) and transition to a new health state based on the given probability matrix \( P \). The objective is to maximize the total expected discounted QALYs, accounting for a monthly discount factor of 0.85. The reward function is straightforward: if the transplant is performed, the immediate reward is 25 QALYs; if the physician waits, the reward is the current health state value. The transition dynamics are governed by the matrix \( P \), where each entry \( P_{ij} \) represents the probability of moving from state \( i \) to state \( j \) in the next month. We employ the value iteration method due to the infinite horizon, iteratively updating the value function for each state by comparing the immediate reward of performing the transplant against the expected discounted future rewards of waiting, considering the probabilistic transitions. Starting from an initial state of 1, we compute the optimal policy by evaluating the expected QALYs for each action and selecting the one that maximizes the total discounted reward. Ultimately, we determine the maximal total expected discounted QALYs for a patient initially in health state 1 by converging the value function to its optimal values through iterative updates until stability is achieved.
</reasoning>
```python
import numpy as np

# Define the states
states = [1, 2, 3, 4]
n_states = len(states)

# Define the action space
actions = ['transplant', 'wait']

# Discount factor
alpha = 0.85

# Transition probabilities
P = np.array([
    [0.2, 0.5, 0.2, 0.1],
    [0.1, 0.4, 0.4, 0.1],
    [0.0, 0.1, 0.7, 0.2],
    [0.3, 0.1, 0.1, 0.5]
])

# Immediate rewards
R_transplant = 25
R_wait = [1, 2, 3, 4]  # QALYs for each state if waiting

# Initialize value function
V = np.zeros(n_states)

# Value iteration parameters
tolerance = 1e-6
max_iterations = 1000
delta = tolerance

# Value iteration
for _ in range(max_iterations):
    delta = 0
    V_new = np.zeros(n_states)
    for s in range(n_states):
        # Calculate expected value of waiting
        expected_value_wait = sum(P[s, s_prime] * V[s_prime] for s_prime in range(n_states))
        # Bellman equation: max of transplant or wait
        V_new[s] = max(R_transplant + alpha * 0, R_wait[s] + alpha * expected_value_wait)
    # Check for convergence
    delta = np.max(np.abs(V_new - V))
    V = V_new
    if delta < tolerance:
        break

# The maximal total expected discounted QALYs for state 1
print(V[0])
```
\end{dplmoutput}
\subsubsection{Case 2: Consistent Failure}
\

In this case, Example~\ref{ex-case2b} is backward-generated from Example~\ref{ex-4reviewcycle} in $\DPBENCH$, which was previously discussed in Section~\ref{sec-failure}. The adapted problem falls under the kidney transplantation category with the characteristic description: “Supports sequential acceptance decisions by weighing potential health outcomes against future treatment opportunities under uncertainty.”

As discussed in Section~\ref{sec-failure}, $\DPLM$ fails to correctly handle the four-day review cycle in the original problem. This issue persists in the perturbed Healthcare-OOD1 version.

\paragraph{Healthcare-OOD1 Adaptation}
\begin{example}\label{ex-case2b}
In response to a moderately infectious public health event, the Metropolitan Health Commission has established a temporary observation facility for isolating and monitoring patients. Every day, the number of patients who meet the recovery criteria and get discharged is stochastic, following this probability distribution:
\begin{center}
\smallskip
\begin{tabular}{lccccc}
\hline
Daily Recoveries & 0 & 1 & 2 & 3 & 4 \\
\hline
Probability & 0.60 & 0.20 & 0.10 & 0.06 & 0.04 \\
\hline
\end{tabular}\smallskip
\end{center}

To maintain social distancing protocols and optimize logistical efficiency, the facility reviews its status every four days and decide the number of newly controlled infections, $Q$, to be transferred from community screening sites. Due to the severity of the regional outbreak, the pool of individuals requiring isolation far exceeds the facility's capacity, ensuring that any chosen quantity $Q$ is always available for immediate transfer. However, the transfer process involves a two-day lead time due to the complexities of community coordination and medical handovers; thus, the quantity $Q$ arrives on the morning of the third day and is available to occupy beds starting then. The facility operates under a cost structure where each executed transfer instruction ($Q > 0$) incurs a fixed administrative fee of 0.35. Furthermore, the facility accounts for the daily burden of care by charging 0.02 for each patient remaining in the facility at the end of the day. To measure the inefficiency of underutilized medical infrastructure, a penalty of 0.75 is levied for each unoccupied bed capacity whenever the daily recovery potential exceeds the actual patient count on hand. In instances where daily occupancy exceeds the facility's physical limit of 40 beds, excess patients are redirected to regional hospitals and are no longer tracked in the facility's state. Assume a planning horizon of 60 days, calculate the minimum total expected cost to operate the facility under an optimal transfer policy when the initial occupancy is 0.
\end{example}
\textbf{DPLM Output}
\begin{dplmoutput}
<reasoning>
To address the challenge of optimizing the operation of a temporary observation facility during a public health event, we classify this as a finite-horizon dynamic programming problem with stochastic daily recoveries. The states are defined by the number of patients currently occupying the facility, ranging from 0 to 40 beds, reflecting the facility's capacity constraints. The actions involve deciding the number of newly controlled infections, \( Q \), to be transferred from community screening sites every four days, considering the two-day lead time for transfers. Our objective is to minimize the total expected cost over the 60-day horizon, which includes administrative fees for transfers, daily care costs for patients, and penalties for unoccupied bed capacity when recovery potential exceeds patient count. The reward function is structured such that each transfer incurs a fixed cost of 0.35, while each patient in the facility incurs a daily care cost of 0.02. Additionally, a penalty of 0.75 is applied for each unoccupied bed if the recovery potential surpasses the actual patient count. Transitions between states are governed by the stochastic nature of daily recoveries, with probabilities specified for 0 to 4 recoveries per day. The boundary conditions include an initial state of zero occupancy and a terminal state at the end of the 60-day period. We employ backward induction to solve this problem, calculating the expected cost for each state-action pair from the final day back to the first. This involves evaluating all possible transfer quantities and their associated costs, considering the probabilistic recovery outcomes. Ultimately, this approach yields the optimal transfer policy that minimizes the total expected cost, starting from an initial occupancy of zero, ensuring efficient use of the facility's resources while adhering to social distancing protocols.
</reasoning>

```python
import numpy as np

# Define constants
max_beds = 40
admin_fee = 0.35
daily_cost_per_patient = 0.02
penalty_per_unoccupied_bed = 0.75
days = 60
review_interval = 4
lead_time = 2

# Recovery probabilities
recovery_probs = [0.60, 0.20, 0.10, 0.06, 0.04]
recovery_values = [0, 1, 2, 3, 4]

# Initialize value function
V = np.zeros((days + 1, max_beds + 1))

# Boundary condition at the end of the planning horizon
for s in range(max_beds + 1):
    V[days, s] = 0

# Backward induction
for t in range(days - 1, -1, -1):
    for s in range(max_beds + 1):
        min_cost = float('inf')
        # Determine feasible actions based on current state
        for Q in range(0, max_beds - s + 1):
            transfer_cost = admin_fee if Q > 0 else 0
            expected_cost = transfer_cost
            for r, prob in zip(recovery_values, recovery_probs):
                next_state = max(0, s + Q - r)
                daily_cost = daily_cost_per_patient * (s + Q)
                penalty = penalty_per_unoccupied_bed * max(0, r - (s + Q))
                future_cost = V[t + 1, next_state]
                period_cost = daily_cost + penalty + future_cost
                expected_cost += prob * period_cost
            min_cost = min(min_cost, expected_cost)
        V[t, s] = min_cost

# Initial state with zero occupancy
initial_occupancy = 0
min_total_expected_cost = V[0, initial_occupancy]

# Print the final result
print(min_total_expected_cost)
```

This code implements the dynamic programming solution for the given problem, following the specified model. It calculates the minimum total expected cost to operate the facility over a 60-day horizon, starting with an initial occupancy of zero. The solution uses backward induction to evaluate the expected costs for each state-action pair, considering the stochastic nature of daily recoveries and the associated costs.
\end{dplmoutput}

\subsection{Healthcare-OOD2: More Complex Medical MDPs}\label{app:OOD2}

Healthcare-OOD2 consists of four classical medical-treatment MDP applications summarized in \citet{Schaefer2004}. These problems are structurally more complex than textbook-style DP problems, featuring longer narratives (437 tokens on average versus 236 in $\DPBENCH$), richer transition structures (all include action-dependent transitions), and more involved modeling assumptions.

We conduct a qualitative case study comparing $\DPLM$ and GPT-4o on these problems. To accommodate this increased complexity, we evaluate our larger model, $\texttt{DPLM-32B-SFT}$. Although $\texttt{DPLM-32B-SFT}$ is less competitive than $\texttt{DPLM-7B-SFT-GRPO}$ in our main experiment, we find that $\texttt{DPLM-32B-SFT}$ produces more stable outputs on Healthcare-OOD2. $\DPLM$ solves two of the four problems completely correctly. On the remaining two, $\DPLM$ produces minor errors (e.g., boundary indexing or algorithm-description mismatch). GPT-4o fails on all four problems, often with substantial modeling errors.

\subsubsection{Problem 1: Epidemic Control}
\begin{example}
A county public-health department is managing an infectious-disease outbreak in a closed population of $N=100$ individuals. Time is divided into weekly periods and the goal is to manage the outbreak over an indefinitely long horizon, with future outcomes discounted by a factor $0.99$ per week. Aside from infection and recovery, there are no births, non-disease deaths, or migration.

At the start of each week, the department reviews surveillance reports indicating how many people are currently infected and then commits to two intervention intensities for the coming week: one that restricts contacts in the community (three possible levels, including no restrictions and two progressively stronger restrictions), and one that governs the clinical resources devoted to treating infected individuals (three possible levels, from standard care to intensive care). Stronger contact restrictions reduce the chance that uninfected people become infected during the week, while more intensive treatment increases the chance that infected individuals recover during the week.

During the week, infections can arise from two sources. Imported infections may occur: in any given week there is a $0.05$ probability that exactly one infected case is introduced from outside the community, provided the entire population is not already infected. In addition, disease transmission occurs through contact between infected and uninfected individuals. If the week begins with $i$ infected individuals and the department applies contact restriction level $q\in\{0,0.25,0.5\}$ for that week, then each uninfected individual becomes infected during that week with probability $0.002(1-q)i$, independently across uninfected individuals.

At the same time, infected individuals may recover during the week. If the department applies clinical intensity level $m\in\{0,1,2\}$ for that week, then each infected individual independently recovers with probability $0.10$ when $m=0$, $0.20$ when $m=1$, and $0.35$ when $m=2$. Recovered individuals rejoin the uninfected population and may become infected again in later weeks.

At the end of each week, the number of infected individuals is updated by adding imported and internally generated infections and subtracting recoveries, subject to the natural bounds that the infected count cannot be less than 0 or greater than 100.

Each week generates a total societal cost made up of three parts: (i) an infection burden that is proportional to the number of infected individuals at the beginning of the week, (ii) a disruption cost proportional to the number of people whose contacts are restricted during the week, and (iii) a treatment cost proportional to the number of infected people treated and to the chosen clinical intensity. Specifically, if a week begins with $i$ infected individuals and the chosen intervention intensities are $(q,m)$, then the week's total cost equals $10i + 2(100q) + 1.5mi$.

The outbreak begins with 10 infected individuals at the start of week 0. The department wishes to choose intervention intensities week by week to minimize the expected discounted sum of weekly costs over an infinite horizon. Please compute the optimal expected discounted cost for this community.
\end{example}
\textbf{DPLM Output}
\begin{dplmoutput}
<reasoning>
To address this dynamic programming problem involving the management of an infectious disease outbreak over an indefinite time horizon, we define the states as the number of infected individuals at the beginning of each week, ranging from 0 to 100. The actions available to the public health department consist of selecting two intervention intensities: a contact restriction level \( q \in \{0, 0.25, 0.5\} \) and a clinical treatment intensity level \( m \in \{0, 1, 2\} \). These interventions influence the transition dynamics by altering infection probabilities and recovery rates, respectively. The objective is to minimize the expected discounted sum of weekly societal costs, which include the infection burden, disruption costs from contact restrictions, and treatment costs. The cost function for a week starting with \( i \) infected individuals and interventions \( (q, m) \) is calculated as \( 10i + 2(100q) + 1.5mi \).

State transitions are governed by the probabilities of new infections due to both imported cases and internal transmissions, as well as recoveries. Specifically, the probability of an uninfected individual becoming infected is \( 0.002(1-q)i \), while the recovery probability for infected individuals is determined by the treatment level \( m \). Additionally, there is a 0.05 probability of an imported infection each week if not all individuals are infected. The process begins with 10 infected individuals at week 0, and the boundary conditions ensure the number of infected individuals remains within the 0 to 100 range.

We employ a value iteration method to solve this problem, given the infinite horizon and discount factor of 0.99. This involves iteratively updating the value function for each state by considering the immediate costs and the expected discounted future costs, based on the transition probabilities resulting from each possible action. By evaluating all potential intervention strategies, we identify the optimal policy that minimizes the expected discounted cost over time. Ultimately, we determine the optimal intervention strategy for each state and calculate the corresponding optimal expected discounted cost for the community, starting with 10 infected individuals.
</reasoning>

```python
import numpy as np

# Parameters
discount_factor = 0.99
infection_cost_per_person = 10
disruption_cost_per_restricted_person = 2
treatment_cost_per_infected_person = 1.5

# State space and action space
max_population = 100
initial_infected = 10
contact_restrictions = [0, 0.25, 0.5]
clinical_intensities = [0, 1, 2]

# Transition probabilities for recovery
recovery_probabilities = {0: 0.10, 1: 0.20, 2: 0.35}

# Initialize value function
V = np.zeros(max_population + 1)

# Value iteration
def value_iteration():
    global V
    epsilon = 1e-6
    while True:
        delta = 0
        new_V = np.copy(V)
        for i in range(max_population + 1):
            min_cost = float('inf')
            for q in contact_restrictions:
                for m in clinical_intensities:
                    # Immediate cost
                    immediate_cost = infection_cost_per_person * i + disruption_cost_per_restricted_person * (100 * q) + treatment_cost_per_infected_person * m * i

                    # Expected future cost
                    expected_future_cost = 0
                    if i < max_population:
                        # Probability of imported infection
                        imported_infection_prob = 0.05

                        # Probability of no imported infection
                        no_imported_infection_prob = 0.95

                        # Calculate expected future cost considering imported and internal infections
                        for k in range(max_population - i + 1):
                            prob_k_new_infections = (np.math.comb(100 - i, k) *
                                                    (0.002 * (1 - q) * i)**k *
                                                    (1 - 0.002 * (1 - q) * i)**(100 - i - k))

                            for r in range(i + 1):
                                prob_r_recoveries = (np.math.comb(i, r) *
                                                    recovery_probabilities[m]**r *
                                                    (1 - recovery_probabilities[m])**(i - r))

                                next_state_no_import = min(max_population, max(0, i + k - r))
                                next_state_with_import = min(max_population, max(0, i + k - r + 1))

                                expected_future_cost += (prob_k_new_infections * prob_r_recoveries *
                                                         (no_imported_infection_prob * V[next_state_no_import] +
                                                          imported_infection_prob * V[next_state_with_import]))

                    total_cost = immediate_cost + discount_factor * expected_future_cost
                    min_cost = min(min_cost, total_cost)

            new_V[i] = min_cost

        delta = max(delta, np.max(np.abs(new_V - V)))
        V = new_V
        if delta < epsilon:
            break

value_iteration()

# Output the optimal expected discounted cost for the initial state
print(V[initial_infected])
```
\end{dplmoutput}
\textbf{GPT-4o Output}
\begin{dplmoutput}
To solve this problem, we can use dynamic programming to find the optimal intervention strategy that minimizes the expected discounted sum of weekly costs over an infinite horizon. The problem can be modeled as a Markov Decision Process (MDP) with the following components:

- **States**: The number of infected individuals, \(i\), ranging from 0 to 100.
- **Actions**: The pair of intervention intensities \((q, m)\), where \(q \in \{0, 0.25, 0.5\}\) and \(m \in \{0, 1, 2\}\).
- **Transition Probabilities**: The probability of moving from one state to another given an action, which depends on the probabilities of infection and recovery.
- **Rewards (or Costs)**: The negative of the weekly cost, which is \(10i + 2(100q) + 1.5mi\).

The goal is to find the policy that minimizes the expected discounted cost. We will use value iteration to solve this MDP.

Here's the Python code to compute the optimal expected discounted cost:

```python
import numpy as np

# Parameters
N = 100
discount_factor = 0.99
infection_burden_cost = 10
disruption_cost_per_person = 2
treatment_cost_per_person = 1.5
imported_infection_prob = 0.05

# Possible actions
contact_restriction_levels = [0, 0.25, 0.5]
clinical_intensity_levels = [0, 1, 2]

# Recovery probabilities
recovery_probs = [0.10, 0.20, 0.35]

# Initialize value function
V = np.zeros(N + 1)

# Value iteration
tolerance = 1e-6
while True:
    delta = 0
    new_V = np.copy(V)
    for i in range(N + 1):
        min_cost = float('inf')
        for q in contact_restriction_levels:
            for m in clinical_intensity_levels:
                # Calculate expected cost
                infection_cost = infection_burden_cost * i
                disruption_cost = disruption_cost_per_person * 100 * q
                treatment_cost = treatment_cost_per_person * m * i
                immediate_cost = infection_cost + disruption_cost + treatment_cost

                # Calculate expected number of infections and recoveries
                infection_prob = 0.002 * (1 - q) * i
                expected_new_infections = (N - i) * infection_prob
                expected_recoveries = i * recovery_probs[m]

                # Calculate expected next state
                expected_i_next = i + expected_new_infections - expected_recoveries
                expected_i_next = min(max(expected_i_next, 0), N)

                # Consider imported infection
                if i < N:
                    expected_i_next = (1 - imported_infection_prob) * expected_i_next + imported_infection_prob * min(expected_i_next + 1, N)

                # Calculate total expected cost
                total_expected_cost = immediate_cost + discount_factor * V[int(round(expected_i_next))]

                # Update minimum cost
                min_cost = min(min_cost, total_expected_cost)

        new_V[i] = min_cost
        delta = max(delta, abs(new_V[i] - V[i]))

    V = new_V
    if delta < tolerance:
        break

# Print the optimal expected discounted cost starting from 10 infected individuals
print(round(V[10], 2))
```

This code uses value iteration to compute the optimal expected discounted cost starting from 10 infected individuals. The value function \(V[i]\) represents the minimum expected discounted cost starting from state \(i\). The code iteratively updates the value function until convergence, and then prints the optimal expected discounted cost for the initial state of 10 infected individuals.
\end{dplmoutput}
\textbf{Analysis:}
\

This problem is challenging primarily due to the transition dynamics. There are two stochastic mechanisms that increase infections and two control mechanisms that reduce infections, making the state transition structure complex. $\DPLM$’s solution is largely correct, but it fails to properly handle the boundary state when the number of infections reaches the maximum population.

For GPT-4o, instead of modeling the full stochastic transition, it computes the expected number of infections and recoveries and then directly substitutes these expected values into the value function. This approach is incorrect, as it ignores the proper expectation over next-period value and collapses the transition uncertainty prematurely.

\subsubsection{Problem 2: Kidney Transplantation}
\begin{example}
A transplant center is managing the long-term treatment of a 65-year-old white male patient with end-stage renal disease who is receiving maintenance dialysis and is eligible for cadaveric kidney transplantation. Time is measured in discrete monthly periods. At the beginning of each month, the patient is in exactly one of five clinical situations: (i) alive on dialysis and eligible to receive an offer, (ii) alive on dialysis but temporarily ineligible to receive an offer, (iii) alive with successful transplantion , (iv) alive with failed transplantion, or (v) dead. In each month that the patient is alive, he accrues quality-adjusted life according to a quality-of-life weight that depends only on the current clinical situation, given by $QOL = (0.65, 0.60, 0.90, 0.60, 0)$ for the five situations listed above, respectively. Future months are discounted at an annual rate of 5\%, implemented on a monthly basis as $\delta = 0.996 = (1-0.05)^{1/12}$.
Kidney offers can occur only in months when the patient is alive on dialysis and eligible. In each eligible month, exactly one kidney offer arrives with probability $\lambda = 0.30$; with probability $1 - \lambda$, no offer arrives. When an offer arrives, the kidney is summarized by a single observed number $x \in (0,1)$. Conditional on an offer arriving, the kidney quality $x$ takes one of five discrete values $x_k$ with probabilities $p_k$, where $p_k$ denotes the probability of observing quality level $x_k$ given that a kidney offer arrives:

\begin{center}
\begin{tabular}{ccc}
\hline
Quality level $k$ & $x_k$ & $p_k$ \\
\hline
1 & 0.55 & 0.15 \\
2 & 0.65 & 0.25 \\
3 & 0.70 & 0.25 \\
4 & 0.88 & 0.20 \\
5 & 0.95 & 0.15 \\
\hline
\end{tabular}
\smallskip
\end{center}

If no offer arrives, no acceptance decision is made in that month. When an offer arrives in an eligible month, the patient observes the realized quality $x$ and must immediately decide whether to accept the kidney for transplantation or reject it and remain on dialysis. If the kidney is rejected, the month proceeds as a dialysis month. If the kidney is accepted, transplantation occurs during that month and, because immunosuppressive therapy imposes side effects that are worse when rejection risk is higher, the quality-of-life weight during the transplant month is reduced by an amount proportional to $(1-x)$. Specifically, if a kidney with quality $x$ is accepted, then the quality-adjusted life accrued during that month equals $\frac{1}{12}\bigl(0.65-\text{Imm}(1-x)\bigr)$ with $\text{Imm}=0.2$, and at the beginning of the next month the patient transitions either to the functioning-transplant situation with probability $x$ or to the failed-transplant situation with probability $1-x$. If the patient does not accept a kidney in the month (either because no offer arrives or because an offer is rejected), then the quality-adjusted life accrued during the month is $\frac{1}{12}QOL_j$ where $QOL_j$ corresponds to the clinical situation at the beginning of the month.
If the patient does not undergo transplantation in the current month, the patient's clinical situation in the following month evolves according to the transition probabilities shown in the table below.

\begin{center}
    \begin{tabular}{lccccc}
\hline
Current clinical situation & Eligible & Temp. ineligible & Functioning & Failed & Death \\
(start of month)            & dialysis &                  & transplant  & transplant & \\
\hline
Eligible on dialysis        & 0.93 & 0.03 & 0     & 0     & 0.04 \\
Temporarily ineligible      & 0.30 & 0.66 & 0     & 0     & 0.04 \\
Successful transplant       & 0    & 0    & 0.988 & 0.007 & 0.005 \\
Failed transplant           & 0    & 0    & 0     & 0.96  & 0.04 \\
Death                       & 0    & 0    & 0     & 0     & 1.00 \\
\hline
\end{tabular}

\smallskip
\textit{Monthly transition probabilities when no transplant is performed.}
\end{center}
The 65-year-old white male patient starts eligible on dialysis at the beginning of month 0. The goal is to choose a rule for when to accept an offered kidney for each observed $x$ to maximize the patient's expected discounted total quality-adjusted life years. Please determine the corresponding maximum expected discounted total QALYs.
\end{example}
\textbf{DPLM Output}
\begin{dplmoutput}
<reasoning>
To solve this medical decision-making problem involving a 65-year-old white male patient with end-stage renal disease, we employ a Markov decision process framework over an infinite time horizon to maximize the patient's expected discounted quality-adjusted life years (QALYs). The states are defined by the patient's clinical situation at the beginning of each month, which can be alive on dialysis and eligible for a transplant, alive on dialysis but temporarily ineligible, alive with a successful transplant, alive with a failed transplant, or dead. The actions involve deciding whether to accept or reject a kidney offer when the patient is eligible on dialysis. Our objective is to maximize the expected discounted QALYs, which involves weighing the quality-of-life weights associated with each state and the impact of potential transplants. The reward function is based on the quality-adjusted life accrued each month, influenced by the clinical situation and the outcome of any transplant decision. If a kidney offer is accepted, the immediate reward is reduced by a factor related to the kidney's quality, reflecting the side effects of immunosuppressive therapy. Transition probabilities dictate how the patient's condition evolves monthly, depending on whether a transplant is performed and its success. We use a value iteration method due to the infinite horizon and discounting factor of \(\delta = 0.996\). This involves iteratively updating the value function for each state by considering the expected discounted future rewards, balancing immediate QALYs against potential future outcomes. Ultimately, we derive the optimal policy, which specifies whether to accept a kidney offer based on its observed quality, and compute the maximum expected discounted QALYs for the patient starting in the eligible-on-dialysis state.
</reasoning>

To solve this problem, we will implement a value iteration algorithm based on the provided dynamic programming model. The goal is to determine the optimal policy for accepting or rejecting kidney offers based on their observed quality, and to compute the maximum expected discounted QALYs for the patient starting in the eligible-on-dialysis state.

Here is the Python code implementing the solution:

```python
import numpy as np

# Define the quality-of-life weights for each state
QOL = np.array([0.65, 0.60, 0.90, 0.60, 0.0])

# Define the discount factor
delta = 0.996

# Define the transition probabilities when no transplant is performed
P_no_transplant = np.array([
    [0.93, 0.03, 0.00, 0.00, 0.04],
    [0.30, 0.66, 0.00, 0.00, 0.04],
    [0.00, 0.00, 0.988, 0.007, 0.005],
    [0.00, 0.00, 0.00, 0.96, 0.04],
    [0.00, 0.00, 0.00, 0.00, 1.00]
])

# Define the probabilities of observing each quality level
quality_levels = np.array([0.55, 0.65, 0.70, 0.88, 0.95])
probabilities = np.array([0.15, 0.25, 0.25, 0.20, 0.15])

# Initialize the value function
V = np.zeros(5)

# Value iteration
def value_iteration(V, delta, QOL, P_no_transplant, quality_levels, probabilities, Imm=0.2, epsilon=1e-6):
    while True:
        V_new = np.copy(V)
        for s in range(5):  # For each state
            if s == 0:  # Eligible on dialysis
                # Calculate the expected value if no offer arrives (probability 0.7)
                expected_value_no_offer = 0.7 * (QOL[s] / 12 + delta * np.dot(P_no_transplant[s], V))

                # Calculate the expected value if an offer arrives (probability 0.3)
                expected_value_offer = 0
                for x, p in zip(quality_levels, probabilities):
                    # Value if accept the offer
                    accept_value = (QOL[s] - Imm * (1 - x)) / 12 + delta * (x * V[2] + (1 - x) * V[3])
                    # Value if reject the offer
                    reject_value = QOL[s] / 12 + delta * np.dot(P_no_transplant[s], V)
                    # Choose the best action
                    expected_value_offer += p * max(accept_value, reject_value)

                # Total expected value for state 0
                V_new[s] = expected_value_no_offer + 0.3 * expected_value_offer
            else:
                # For other states, no decision is made, just follow the transition
                V_new[s] = QOL[s] / 12 + delta * np.dot(P_no_transplant[s], V)

        # Check for convergence
        if np.max(np.abs(V_new - V)) < epsilon:
            break
        V = V_new

    return V

# Run value iteration
V_optimal = value_iteration(V, delta, QOL, P_no_transplant, quality_levels, probabilities)

# Output the maximum expected discounted QALYs starting from the eligible-on-dialysis state
print(V_optimal[0])
```

This code performs value iteration to find the optimal policy for maximizing the patient's expected discounted QALYs. It considers the different states, actions, and transitions as described in the problem statement. The output is the maximum expected discounted QALYs for the patient starting in the eligible-on-dialysis state.
\end{dplmoutput}
\textbf{GPT-4o Output}
\begin{dplmoutput}
To solve this problem, we can use dynamic programming to determine the optimal policy for accepting or rejecting a kidney offer based on its quality. The goal is to maximize the expected discounted total quality-adjusted life years (QALYs) for the patient.

### Problem Modeling

1. **States**: The patient can be in one of five states:
   - \( S_1 \): Eligible on dialysis
   - \( S_2 \): Temporarily ineligible
   - \( S_3 \): Functioning transplant
   - \( S_4 \): Failed transplant
   - \( S_5 \): Dead

2. **Actions**: The patient can either accept or reject a kidney offer when in state \( S_1 \).

3. **Rewards**: The reward is the quality-adjusted life accrued during the month, which depends on the state and action.

4. **Transition Probabilities**: These are given for each state and action.

5. **Discount Factor**: The monthly discount factor is \( \delta = 0.996 \).

6. **Quality Levels**: The kidney quality levels \( x_k \) and their probabilities \( p_k \) are given.

### Dynamic Programming Approach

We will use a value iteration approach to solve this problem. The value function \( V(s) \) represents the maximum expected discounted QALYs starting from state \( s \).

1. **Initialize**: Start with an initial guess for the value function \( V(s) \) for each state \( s \).

2. **Iterate**: Update the value function using the Bellman equation until convergence.

3. **Policy Extraction**: Determine the optimal action for each state based on the updated value function.

### Python Code

Here's the Python code to implement the above approach:

```python
import numpy as np

# Constants
delta = 0.996
lambda_offer = 0.30
Imm = 0.2
QOL = [0.65, 0.60, 0.90, 0.60, 0.0]
x_k = [0.55, 0.65, 0.70, 0.88, 0.95]
p_k = [0.15, 0.25, 0.25, 0.20, 0.15]

# Transition probabilities when no transplant is performed
P_no_transplant = np.array([
    [0.93, 0.03, 0, 0, 0.04],
    [0.30, 0.66, 0, 0, 0.04],
    [0, 0, 0.988, 0.007, 0.005],
    [0, 0, 0, 0.96, 0.04],
    [0, 0, 0, 0, 1.0]
])

# Initialize value function
V = np.zeros(5)

# Value iteration
tolerance = 1e-6
max_iterations = 1000
for _ in range(max_iterations):
    V_new = np.zeros(5)

    # Update for each state
    for s in range(5):
        if s == 4:  # Dead state
            V_new[s] = 0
        else:
            # Calculate expected value if no transplant is performed
            expected_no_transplant = sum(P_no_transplant[s, s_prime] * V[s_prime] for s_prime in range(5))
            expected_no_transplant = (QOL[s] / 12) + delta * expected_no_transplant

            if s == 0:  # Eligible on dialysis
                # Calculate expected value if a transplant is performed
                expected_transplant = 0
                for k in range(5):
                    x = x_k[k]
                    reward_transplant = (0.65 - Imm * (1 - x)) / 12
                    expected_transplant += p_k[k] * (reward_transplant + delta * (x * V[2] + (1 - x) * V[3]))

                # Consider the probability of offer arrival
                expected_transplant = lambda_offer * expected_transplant + (1 - lambda_offer) * expected_no_transplant

                # Choose the best action
                V_new[s] = max(expected_no_transplant, expected_transplant)
            else:
                V_new[s] = expected_no_transplant

    # Check for convergence
    if np.max(np.abs(V_new - V)) < tolerance:
        break

    V = V_new

# Print the final numerical answer
print(V[0])
```

This code uses value iteration to compute the maximum expected discounted QALYs starting from the initial state (eligible on dialysis). The result is printed as the final numerical answer.
\end{dplmoutput}
\textbf{Analysis:}
\

This problem is challenging because it involves state-dependent action spaces and action-dependent transitions. For example, a decision is required only when the patient is in the ``Eligible on dialysis'' state and a kidney arrives. Moreover, the transition dynamics differ significantly depending on whether the offer is accepted or rejected.

$\DPLM$ models these features correctly. In contrast, GPT-4o fails to recognize that the decision to accept or reject a kidney offer should depend on the kidney quality $x$. Instead, it imposes a single uniform policy across all quality levels, ignoring the heterogeneity in $x$ and thereby mis-specifying the optimal policy structure.

\subsubsection{Problem 3: Spherocytosis Treatment}
\begin{example}
A pediatric hematology clinic follows children diagnosed with mild hereditary spherocytosis (HS), a chronic hemolytic anemia associated with an increased lifetime risk of pigment gallstones. In some patients, removal of the spleen alleviates hemolysis-related symptoms. Patients receive regular annual follow-up, during which clinicians decide whether to continue conservative management or to perform surgical intervention. Alternative management strategies are evaluated based on their impact on expected lifetime quality-adjusted life years (QALYs).

At each annual visit, the patient's condition is described by gallstone status and spleen status. Gallstone disease is categorized by increasing symptom severity, with post-cholecystectomy as an absorbing gallstone state (Table 1). Death is treated as an absorbing outcome.

\begin{center}
\smallskip
\begin{tabular}{ll}
\hline
Label & Interpretation \\
\hline
$G_0$ & No gallstones \\
$G_1$ & Asymptomatic gallstones \\
$G_2$ & Symptomatic gallstones with occasional biliary colic \\
$G_3$ & Symptomatic gallstones with recurrent colic \\
$G_4$ & Gallbladder removed (Post-cholecystectomy) \\
$D$ & Death \\
\hline
\end{tabular}

\smallskip
\textit{Table 1. Gallstone status categories.}
\end{center}

Spleen status records whether the spleen is present or has been removed. Once the spleen is removed, it remains absent permanently.

\begin{center}
\smallskip
\begin{tabular}{ll}
\hline
Label & Interpretation \\
\hline
$S_0$ & Spleen present \\
$S_1$ & Spleen absent (post-splenectomy) \\
\hline
\end{tabular}

\smallskip
\textit{Table 2. Spleen status categories.}
\end{center}

During each year of follow-up, patients are also subject to mortality from nonoperative causes. Background mortality depends on age and increases over the life course.

\begin{center}
\begin{tabular}{lc}
\hline
Age band (years) & Probability of death \\
\hline
6 to 14 & 0.0002 \\
15 to 24 & 0.0004 \\
25 to 34 & 0.0008 \\
35 to 44 & 0.0016 \\
45 to 54 & 0.0040 \\
55 to 64 & 0.0100 \\
65 to 74 & 0.0250 \\
75 to 84 & 0.0600 \\
85 to 101 & 0.1400 \\
\hline
\end{tabular}

\smallskip
\textit{Table 3. Annual background mortality probabilities by age band.}
\end{center}

During each year of follow-up, the patient first faces age-related background mortality.
If the patient survives background mortality for that year, clinicians then choose one of four management actions for the coming year: continued observation without surgery, splenectomy alone, cholecystectomy alone, or combined splenectomy and cholecystectomy.

If a surgical action is selected, the patient is subsequently exposed to an additional procedure-specific operative mortality risk during that year. Background mortality and operative mortality are assumed to be independent, and operative mortality applies only conditional on survival from background mortality.

\begin{center}
\begin{tabular}{lc}
\hline
Procedure & Probability of operative death \\
\hline
splenectomy alone & 0.0010 \\
cholecystectomy alone & 0.0005 \\
combined & 0.0013 \\
\hline
\end{tabular}

\smallskip
\textit{Table 4. Operative mortality probabilities.}
\end{center}

If death occurs, the patient dies during that year and no further QALYs are accrued. If the patient survives, the direct clinical effects of surgery are reflected at the start of the following year. Cholecystectomy permanently places the patient in the post-cholecystectomy gallstone category $G_4$, splenectomy permanently changes spleen status from $S_0$ to $S_1$, and combined surgery produces both effects simultaneously. Apart from these changes, no other clinical descriptors are directly altered by surgery. The year in which surgery is performed is also associated with a temporary reduction in quality of life due to postoperative recovery.
\begin{center}
\begin{tabular}{lc}
\hline
Procedure & QALY penalty \\
\hline
splenectomy alone & 0.05 \\
cholecystectomy alone & 0.03 \\
combined & 0.07 \\
\hline
\end{tabular}

\smallskip
\textit{Table 5. One-year QALY penalty applied in the surgery year.}
\end{center}

For patients who remain alive throughout a year, health-related quality of life depends on gallstone severity and spleen status. Baseline annual QALYs reflect the burden of gallstone symptoms. Living without a spleen provides a modest ongoing improvement due to reduced hemolysis-related symptoms.
\begin{center}
\begin{tabular}{lc}
\hline
Gallstone status & Baseline QALYs \\
\hline
$G_0$ & 0.98 \\
$G_1$ & 0.96 \\
$G_2$ & 0.90 \\
$G_3$ & 0.82 \\
$G_4$ & 0.97 \\
\hline
\end{tabular}

\smallskip
\textit{Table 6. Baseline annual QALYs by gallstone status.}\\
\smallskip
\begin{tabular}{lc}
\hline
Spleen status & Add to baseline QALY \\
\hline
$S_0$ & +0.00 \\
$S_1$ & +0.02 \\
\hline
\end{tabular}

\smallskip
\textit{Table 7. Annual QALY adjustment by spleen status.}
\end{center}

Among patients who survive a given year without undergoing cholecystectomy, gallstone disease may evolve over time. The pace of this evolution depends on whether the spleen is present: ongoing hemolysis with an intact spleen is associated with higher rates of gallstone formation and progression, whereas these risks are reduced after splenectomy. Once the gallbladder has been removed, gallstone status remains permanently in the post-cholecystectomy category.

\begin{center}
\begin{tabular}{lccccc}
\hline
From \textbackslash\ To & $G_0$ & $G_1$ & $G_2$ & $G_3$ & $G_4$ \\
\hline
$G_0$ & 0.986 & 0.014 & 0.000 & 0.000 & 0.000 \\
$G_1$ & 0.000 & 0.950 & 0.045 & 0.005 & 0.000 \\
$G_2$ & 0.000 & 0.000 & 0.860 & 0.140 & 0.000 \\
$G_3$ & 0.000 & 0.000 & 0.030 & 0.970 & 0.000 \\
$G_4$ & 0.000 & 0.000 & 0.000 & 0.000 & 1.000 \\
\hline
\end{tabular}

\smallskip
\textit{Table 8. One-year gallstone transition probabilities conditional on survival (spleen present: $S_0$).}\\
\smallskip
\begin{tabular}{lccccc}
\hline
From \textbackslash\ To & $G_0$ & $G_1$ & $G_2$ & $G_3$ & $G_4$ \\
\hline
$G_0$ & 0.994 & 0.006 & 0.000 & 0.000 & 0.000 \\
$G_1$ & 0.000 & 0.960 & 0.038 & 0.002 & 0.000 \\
$G_2$ & 0.000 & 0.000 & 0.880 & 0.120 & 0.000 \\
$G_3$ & 0.000 & 0.000 & 0.040 & 0.960 & 0.000 \\
$G_4$ & 0.000 & 0.000 & 0.000 & 0.000 & 1.000 \\
\hline
\end{tabular}

\smallskip
\textit{Table 9. One-year gallstone transition probabilities conditional on survival (spleen absent: $S_1$).}
\end{center}
The patient enters follow-up at age 6 with no gallstones and an intact spleen. Her condition is reassessed annually over a finite horizon extending to age 101. The objective is to determine a management strategy that maximizes expected lifetime quality-adjusted survival.
\end{example}
\textbf{DPLM Output}
\begin{dplmoutput}
<reasoning>
To solve this medical decision-making problem involving a pediatric hematology clinic and patients with mild hereditary spherocytosis, we model the situation as a finite-horizon dynamic programming problem. The states are defined by the patient's gallstone status and spleen status, along with their age, which together capture the health condition and potential quality of life. The possible actions at each state include continuing conservative management, performing splenectomy alone, cholecystectomy alone, or a combined procedure. Our objective is to maximize the expected lifetime quality-adjusted survival, which involves calculating the quality-adjusted life years (QALYs) over the patient's lifetime. The reward function is determined by the annual QALYs, which depend on the gallstone severity and spleen status, adjusted for any surgical penalties. Transition dynamics are governed by the probabilities of gallstone progression, which vary based on spleen status, and the fixed probabilities of background and operative mortality. The process begins with the patient at age 6, having no gallstones and an intact spleen, and continues until age 101. We employ backward induction to solve this problem, starting from the final year and working backwards to the initial state, calculating the expected QALYs for each possible action and state. This involves evaluating the immediate QALYs and the expected future QALYs, considering the transition probabilities and mortality risks. Ultimately, we derive an optimal management strategy that specifies the best action at each state to maximize the patient's expected lifetime QALYs, providing a comprehensive plan for managing the patient's condition over the long term.
</reasoning>

```python
import numpy as np

# Define constants
age_start = 6
age_end = 101
ages = list(range(age_start, age_end + 1))
gallstone_states = ['G_0', 'G_1', 'G_2', 'G_3', 'G_4']
spleen_states = ['S_0', 'S_1']
actions = ['Observe', 'Splenectomy', 'Cholecystectomy', 'Combined']

# Background mortality probabilities by age band
background_mortality = {
    (6, 14): 0.0002,
    (15, 24): 0.0004,
    (25, 34): 0.0008,
    (35, 44): 0.0016,
    (45, 54): 0.0040,
    (55, 64): 0.0100,
    (65, 74): 0.0250,
    (75, 84): 0.0600,
    (85, 101): 0.1400,
}

# Operative mortality probabilities
operative_mortality = {
    'Splenectomy': 0.0010,
    'Cholecystectomy': 0.0005,
    'Combined': 0.0013,
}

# QALY penalties for surgery
qaly_penalty = {
    'Splenectomy': 0.05,
    'Cholecystectomy': 0.03,
    'Combined': 0.07,
}

# Baseline annual QALYs by gallstone status
baseline_qaly = {
    'G_0': 0.98,
    'G_1': 0.96,
    'G_2': 0.90,
    'G_3': 0.82,
    'G_4': 0.97,
}

# QALY adjustment by spleen status
spleen_adjustment = {
    'S_0': 0.00,
    'S_1': 0.02,
}

# Transition probabilities for gallstone status when spleen is present (S_0)
transition_prob_S0 = {
    'G_0': [0.986, 0.014, 0.000, 0.000, 0.000],
    'G_1': [0.000, 0.950, 0.045, 0.005, 0.000],
    'G_2': [0.000, 0.000, 0.860, 0.140, 0.000],
    'G_3': [0.000, 0.000, 0.030, 0.970, 0.000],
    'G_4': [0.000, 0.000, 0.000, 0.000, 1.000],
}

# Transition probabilities for gallstone status when spleen is absent (S_1)
transition_prob_S1 = {
    'G_0': [0.994, 0.006, 0.000, 0.000, 0.000],
    'G_1': [0.000, 0.960, 0.038, 0.002, 0.000],
    'G_2': [0.000, 0.000, 0.880, 0.120, 0.000],
    'G_3': [0.000, 0.000, 0.040, 0.960, 0.000],
    'G_4': [0.000, 0.000, 0.000, 0.000, 1.000],
}

# Initialize value function dictionary
V = {}

# Boundary condition: V(G, S, 101) = 0 for all G, S
for g in gallstone_states:
    for s in spleen_states:
        V[(g, s, age_end)] = 0

# Backward induction to fill in the value function for t < 101
for t in reversed(ages[:-1]):
    for g in gallstone_states:
        for s in spleen_states:
            # Get background mortality probability for current age
            for age_range, prob in background_mortality.items():
                if age_range[0] <= t <= age_range[1]:
                    p_death_background = prob
                    break

            # Calculate value for each action
            V_action = {}

            for action in actions:
                if action == 'Observe':
                    p_survive = 1 - p_death_background
                    qaly = baseline_qaly[g] + spleen_adjustment[s]

                    # Transition probabilities based on spleen status
                    if s == 'S_0':
                        transition_probs = transition_prob_S0[g]
                    else:
                        transition_probs = transition_prob_S1[g]

                    # Expected future value
                    expected_future_value = sum(
                        transition_probs[i] * V.get((gallstone_states[i], s, t + 1), 0)
                        for i in range(len(gallstone_states))
                    )

                    V_action[action] = p_survive * (qaly + expected_future_value)

                elif action == 'Splenectomy':
                    p_survive = (1 - p_death_background) * (1 - operative_mortality['Splenectomy'])
                    qaly = baseline_qaly[g] + spleen_adjustment['S_1'] - qaly_penalty['Splenectomy']

                    # Transition probabilities remain the same since gallstone status doesn't change immediately
                    if s == 'S_0':
                        transition_probs = transition_prob_S0[g]
                    else:
                        transition_probs = transition_prob_S1[g]

                    # Expected future value
                    expected_future_value = sum(
                        transition_probs[i] * V.get((gallstone_states[i], 'S_1', t + 1), 0)
                        for i in range(len(gallstone_states))
                    )

                    V_action[action] = p_survive * (qaly + expected_future_value)

                elif action == 'Cholecystectomy':
                    p_survive = (1 - p_death_background) * (1 - operative_mortality['Cholecystectomy'])
                    qaly = baseline_qaly['G_4'] + spleen_adjustment[s] - qaly_penalty['Cholecystectomy']

                    # Transition to G_4
                    expected_future_value = V.get(('G_4', s, t + 1), 0)

                    V_action[action] = p_survive * (qaly + expected_future_value)

                elif action == 'Combined':
                    p_survive = (1 - p_death_background) * (1 - operative_mortality['Combined'])
                    qaly = baseline_qaly['G_4'] + spleen_adjustment['S_1'] - qaly_penalty['Combined']

                    # Transition to G_4 and S_1
                    expected_future_value = V.get(('G_4', 'S_1', t + 1), 0)

                    V_action[action] = p_survive * (qaly + expected_future_value)

            # Choose the action that maximizes the value function
            V[(g, s, t)] = max(V_action.values())

# Initial state: (G_0, S_0, 6)
initial_state = ('G_0', 'S_0', age_start)
max_expected_lifetime_qaly = V[initial_state]

# Print the final result
print(max_expected_lifetime_qaly)
```
\end{dplmoutput}
\textbf{GPT-4o Output}
\begin{dplmoutput}
To solve this problem using dynamic programming, we need to model the decision-making process over time, considering the patient's age, gallstone status, and spleen status. The goal is to maximize the expected lifetime quality-adjusted life years (QALYs) by choosing the optimal management strategy at each decision point.

Here's a step-by-step outline of the approach:

1. **State Representation**: Each state is represented by a tuple `(age, gallstone_status, spleen_status)`. The gallstone status can be one of `G_0`, `G_1`, `G_2`, `G_3`, `G_4`, and `D`, while the spleen status can be `S_0` or `S_1`.

2. **Actions**: The possible actions are:
   - `observe`: Continue observation without surgery.
   - `splenectomy`: Perform splenectomy alone.
   - `cholecystectomy`: Perform cholecystectomy alone.
   - `combined`: Perform both splenectomy and cholecystectomy.

3. **Transition Probabilities**: Define the transition probabilities for gallstone progression based on spleen status and the probabilities of death due to background and operative mortality.

4. **Rewards**: Calculate the QALYs for each state-action pair, considering the baseline QALYs, spleen status adjustment, and surgery penalties.

5. **Dynamic Programming Recursion**: Use backward induction to compute the expected QALYs for each state, starting from the terminal age (101) and moving backward to the initial age (6). At each state, choose the action that maximizes the expected QALYs.

6. **Implementation**: Implement the above logic in Python to compute the optimal strategy and the maximum expected lifetime QALYs.

Here's the Python code to solve the problem:

```python
# Define constants
ages = list(range(6, 102))
gallstone_states = ['G_0', 'G_1', 'G_2', 'G_3', 'G_4', 'D']
spleen_states = ['S_0', 'S_1']
actions = ['observe', 'splenectomy', 'cholecystectomy', 'combined']

# Mortality probabilities
background_mortality = {
    (6, 14): 0.0002,
    (15, 24): 0.0004,
    (25, 34): 0.0008,
    (35, 44): 0.0016,
    (45, 54): 0.0040,
    (55, 64): 0.0100,
    (65, 74): 0.0250,
    (75, 84): 0.0600,
    (85, 101): 0.1400
}

operative_mortality = {
    'splenectomy': 0.0010,
    'cholecystectomy': 0.0005,
    'combined': 0.0013
}

# QALY penalties
surgery_penalty = {
    'splenectomy': 0.05,
    'cholecystectomy': 0.03,
    'combined': 0.07
}

# Baseline QALYs
baseline_qalys = {
    'G_0': 0.98,
    'G_1': 0.96,
    'G_2': 0.90,
    'G_3': 0.82,
    'G_4': 0.97
}

# Spleen status QALY adjustment
spleen_qaly_adjustment = {
    'S_0': 0.00,
    'S_1': 0.02
}

# Transition probabilities
gallstone_transition_S0 = {
    'G_0': [0.986, 0.014, 0.000, 0.000, 0.000],
    'G_1': [0.000, 0.950, 0.045, 0.005, 0.000],
    'G_2': [0.000, 0.000, 0.860, 0.140, 0.000],
    'G_3': [0.000, 0.000, 0.030, 0.970, 0.000],
    'G_4': [0.000, 0.000, 0.000, 0.000, 1.000]
}

gallstone_transition_S1 = {
    'G_0': [0.994, 0.006, 0.000, 0.000, 0.000],
    'G_1': [0.000, 0.960, 0.038, 0.002, 0.000],
    'G_2': [0.000, 0.000, 0.880, 0.120, 0.000],
    'G_3': [0.000, 0.000, 0.040, 0.960, 0.000],
    'G_4': [0.000, 0.000, 0.000, 0.000, 1.000]
}

# Helper function to get background mortality for a given age
def get_background_mortality(age):
    for age_range, prob in background_mortality.items():
        if age_range[0] <= age <= age_range[1]:
            return prob
    return 0.0

# Initialize the value function
V = {}
for age in ages:
    for g in gallstone_states:
        for s in spleen_states:
            V[(age, g, s)] = 0.0

# Dynamic programming recursion
for age in reversed(ages):
    for g in gallstone_states:
        for s in spleen_states:
            if g == 'D':
                continue  # No QALYs after death

            # Calculate expected QALYs for each action
            action_values = {}
            for action in actions:
                if action == 'observe':
                    # Transition probabilities
                    if s == 'S_0':
                        transition_probs = gallstone_transition_S0[g]
                    else:
                        transition_probs = gallstone_transition_S1[g]

                    # Expected QALYs
                    expected_qalys = 0.0
                    for i, next_g in enumerate(gallstone_states[:-1]):
                        next_state = (age + 1, next_g, s)
                        expected_qalys += transition_probs[i] * V[next_state]

                    # Add baseline QALYs
                    expected_qalys += baseline_qalys[g] + spleen_qaly_adjustment[s]

                else:
                    # Surgery action
                    if action == 'splenectomy':
                        operative_death_prob = operative_mortality['splenectomy']
                        next_s = 'S_1'
                        next_g = g
                    elif action == 'cholecystectomy':
                        operative_death_prob = operative_mortality['cholecystectomy']
                        next_s = s
                        next_g = 'G_4'
                    elif action == 'combined':
                        operative_death_prob = operative_mortality['combined']
                        next_s = 'S_1'
                        next_g = 'G_4'

                    # Expected QALYs
                    expected_qalys = (1 - operative_death_prob) * V[(age + 1, next_g, next_s)]
                    expected_qalys += baseline_qalys[next_g] + spleen_qaly_adjustment[next_s]
                    expected_qalys -= surgery_penalty[action]

                # Background mortality
                background_death_prob = get_background_mortality(age)
                action_values[action] = (1 - background_death_prob) * expected_qalys

            # Choose the action with the maximum expected QALYs
            V[(age, g, s)] = max(action_values.values())

# Initial state
initial_state = (6, 'G_0', 'S_0')
print(V[initial_state])
```

This code models the problem using dynamic programming and computes the maximum expected lifetime QALYs starting from the initial state `(6, 'G_0', 'S_0')`. The code only prints the final numerical answer, which is the expected QALYs for the optimal management strategy.
\end{dplmoutput}

\textbf{Analysis:}
\

This problem is challenging for several reasons. First, the context is long and includes multiple information-dense tables. In addition, the state is multi-dimensional and the rewards are action-dependent, which increases the modeling burden.

$\DPLM$ captures these features correctly. In contrast, GPT-4o exhibits two main errors. First, it makes an indexing mistake in the backward iteration algorithm. Second, it computes rewards incorrectly. In particular, when calculating QALYs, the flow reward should be conditioned on the patient surviving the operation. The correct term is
$
(1-\text{operative\_death\_prob})\big(r + V_{t+1}\big),
$
whereas GPT-4o uses
$
(1-\text{operative\_death\_prob})V_{t+1} + r,
$
which fails to discount the immediate reward by the survival probability.

\subsubsection{Problem 4: Liver Transplantation}

\begin{example}
A transplant center must determine the optimal timing of a living-donor liver transplant for a single patient. The patient is a 60-year-old woman with primary biliary cirrhosis and blood type A. A medically suitable living donor has already been identified and is available whenever transplantation is undertaken.
At the beginning of each day, clinicians observe the patient's current MELD (Model for End-Stage Liver Disease) score, an integer between 6 and 40, with higher values indicating more severe disease. If the patient is alive and has not yet undergone transplantation at the start of the day, clinicians must decide whether to proceed with transplantation on that day or to wait one more day and reassess the decision the following morning, provided the patient survives.
If clinicians choose to wait for one day and the patient survives, the patient faces a probability of death that depends on the current MELD score. For a patient with MELD score $m \in \{6,7,\dots,40\}$, the daily mortality probability is given by $p_{\mathrm{death}}(m) = \min\left\{0.20,\; 2.5\times 10^{-5}\exp\bigl(0.18(m-6)\bigr)\right\}$.
Otherwise, if the patient without undergoing transplantation survives the day, the patient obtains one day of life and the patient's MELD score evolves stochastically the next day. Specifically, the MELD score on the next day equals \(m+\Delta\), where \(\Delta \in \{-1,0,+1,+2\}\) with probabilities
\[
\mathbb{P}(\Delta=-1)=0.03,\quad
\mathbb{P}(\Delta=0)=0.72-0.008(m-6),
\]
\[
\mathbb{P}(\Delta=+1)=0.22+0.006(m-6),\quad
\mathbb{P}(\Delta=+2)=1-\sum_{\Delta\in\{-1,0,+1\}}\mathbb{P}(\Delta).
\]
After the change \(\Delta\) is realized, the next-day MELD score is constrained to remain within the interval \([6,40]\); that is, values below 6 are reset to 6 and values above 40 are reset to 40.
Alternatively, if clinicians choose to perform the transplant when the patient's MELD score is \(m\), the patient faces a MELD-dependent operative mortality probability
\[
p_{\mathrm{op}}(m)=\min\left\{0.08,\;0.001+0.00015(m-6)^2\right\}.
\]
If operative death occurs, no additional life-days are accrued. If the patient survives surgery, she receives an expected remaining lifetime, measured in days, given by
\[
L_{\mathrm{post}}(m)=\max\left\{500,\;5600-35(m-6)-1.2(m-6)^2\right\}.
\]
The patient is initially alive with MELD score 18 at the start of day 0. Determine the optimal transplantation-timing policy, and output the optimal expected life-days for this patient starting from MELD 18.
\end{example}
\textbf{$\DPLM$ Output}
\begin{dplmoutput}
<reasoning>
To solve this medical decision-making problem involving the optimal timing of a living-donor liver transplant for a patient with primary biliary cirrhosis, we approach it as a finite-horizon dynamic programming problem. The states are defined by the patient's current MELD score, which ranges from 6 to 40, reflecting the severity of the liver disease. The actions available to clinicians each day are either to proceed with the transplantation or to wait and reassess the following day. Our objective is to maximize the expected life-days for the patient, considering the trade-offs between the risks of waiting—such as the daily mortality probability based on the MELD score—and the benefits and risks of transplantation, including operative mortality and postoperative life expectancy. The reward function incorporates the life-days gained if the patient survives, along with the probabilities of transitioning between MELD scores if the patient chooses to wait. These transitions are governed by stochastic changes in the MELD score, constrained to remain within the interval [6, 40]. The initial state is a MELD score of 18 at day 0, and the terminal state is defined by the end of the patient's life. We employ backward induction to solve this problem, calculating the expected life-days for each possible state-action pair from the last potential day back to the initial day. By comparing the expected life-days for each decision at every MELD score, we derive the optimal policy that specifies whether to perform the transplant immediately or wait for a potentially more favorable condition. Ultimately, this methodology allows us to determine the optimal expected life-days for the patient starting from a MELD score of 18, guiding the clinicians in making the best possible decision regarding the timing of the transplant.
</reasoning>

To solve the given dynamic programming problem regarding the optimal timing of a living-donor liver transplant, we can implement a solution using Python. The solution will follow the outlined dynamic programming model, taking into account the state space, actions, transitions, Bellman equation, and boundary conditions. Here's the implementation:

```python
import numpy as np

# Constants
MIN_MELD = 6
MAX_MELD = 40
INITIAL_MELD = 18

# Probability functions
def p_death(m):
    return min(0.20, 2.5e-5 * np.exp(0.18 * (m - 6)))

def p_op(m):
    return min(0.08, 0.001 + 0.00015 * (m - 6)**2)

def L_post(m):
    return max(500, 5600 - 35 * (m - 6) - 1.2 * (m - 6)**2)

# Transition probabilities for MELD score changes
def transition_probabilities(m):
    delta_probs = {
        -1: 0.03,
        0: 0.72 - 0.008 * (m - 6),
        1: 0.22 + 0.006 * (m - 6),
        2: 1 - (0.03 + 0.72 - 0.008 * (m - 6) + 0.22 + 0.006 * (m - 6))
    }
    return delta_probs

# Initialize value function table
V = np.zeros(MAX_MELD + 1)

# Backward induction
for day in range(1000, -1, -1):  # Assuming a large number of days for convergence
    V_new = np.zeros_like(V)
    for m in range(MIN_MELD, MAX_MELD + 1):
        # Option 1: Wait
        wait_value = (1 - p_death(m)) * (1 + np.sum(
            transition_probabilities(m)[delta] * V[min(max(m + delta, MIN_MELD), MAX_MELD)]
            for delta in [-1, 0, 1, 2]
        ))

        # Option 2: Transplant
        transplant_value = (1 - p_op(m)) * L_post(m)

        # Choose the action that maximizes the expected life-days
        V_new[m] = max(wait_value, transplant_value)

    # Check for convergence
    if np.allclose(V, V_new, atol=1e-6):
        break

    V = V_new

# Output the optimal expected life-days starting from MELD score 18
print(V[INITIAL_MELD])
```

### Explanation:
1. **State Space:** The MELD score ranges from 6 to 40.
2. **Actions:** At each state, the action is either to wait or to transplant.
3. **Transition Probabilities:**
   - If waiting, the MELD score changes stochastically according to specified probabilities.
   - If transplanting, there's an operative mortality risk, and if survived, a fixed life expectancy is granted.
4. **Bellman Equation:** The value function is updated by considering the maximum expected life-days from either waiting or transplanting.
5. **Boundary Condition:** The value function is initialized to zero, representing no life-days if the patient is deceased.

This code uses backward induction to compute the optimal expected life-days for each MELD score, starting from a large number of days and iterating until convergence. The result for the initial MELD score of 18 is printed as the final output.
\end{dplmoutput}
\textbf{GPT-4o Output}
\begin{dplmoutput}
To solve this problem using dynamic programming, we need to model the decision process as a Markov Decision Process (MDP). The state of the system is defined by the patient's MELD score, and the actions are either to perform the transplant or to wait. The goal is to maximize the expected life-days of the patient.

Here's a step-by-step outline of the solution:

1. **State Space**: The states are the possible MELD scores, \(m \in \{6, 7, \dots, 40\}\).

2. **Actions**: For each state, the actions are:
   - `transplant`: Perform the transplant.
   - `wait`: Wait for one more day.

3. **Transition Probabilities**:
   - If `wait` is chosen, the MELD score changes according to the given probabilities for \(\Delta\).
   - If `transplant` is chosen, the patient either survives or dies based on the operative mortality probability.

4. **Rewards**:
   - If `wait` is chosen and the patient survives, the reward is 1 day of life.
   - If `transplant` is chosen and the patient survives, the reward is the expected remaining lifetime \(L_{\mathrm{post}}(m)\).

5. **Value Function**: Define \(V(m)\) as the maximum expected life-days starting from MELD score \(m\).

6. **Bellman Equation**:
   - For `wait`:
     \[
     V_{\mathrm{wait}}(m) = (1 - p_{\mathrm{death}}(m)) \left(1 + \sum_{\Delta} \mathbb{P}(\Delta) V(m + \Delta)\right)
     \]
   - For `transplant`:
     \[
     V_{\mathrm{transplant}}(m) = (1 - p_{\mathrm{op}}(m)) L_{\mathrm{post}}(m)
     \]
   - Optimal value:
     \[
     V(m) = \max(V_{\mathrm{wait}}(m), V_{\mathrm{transplant}}(m))
     \]

7. **Boundary Conditions**: If the patient dies, the value is 0.

Now, let's implement this in Python:

```python
import numpy as np

# Constants
MELD_MIN = 6
MELD_MAX = 40

# Probability of death without transplant
def p_death(m):
    return min(0.20, 2.5e-5 * np.exp(0.18 * (m - 6)))

# Probability of operative death
def p_op(m):
    return min(0.08, 0.001 + 0.00015 * (m - 6)**2)

# Expected life-days post-transplant
def L_post(m):
    return max(500, 5600 - 35 * (m - 6) - 1.2 * (m - 6)**2)

# Transition probabilities for MELD score change
def transition_probs(m):
    p_delta = {
        -1: 0.03,
        0: 0.72 - 0.008 * (m - 6),
        1: 0.22 + 0.006 * (m - 6),
    }
    p_delta[2] = 1 - sum(p_delta.values())
    return p_delta

# Initialize value function
V = np.zeros(MELD_MAX + 1)

# Dynamic programming to compute V(m)
for m in range(MELD_MAX, MELD_MIN - 1, -1):
    # Value if we wait
    p_delta = transition_probs(m)
    V_wait = (1 - p_death(m)) * (1 + sum(p * V[min(max(m + delta, MELD_MIN), MELD_MAX)] for delta, p in p_delta.items()))

    # Value if we transplant
    V_transplant = (1 - p_op(m)) * L_post(m)

    # Optimal value
    V[m] = max(V_wait, V_transplant)

# Output the optimal expected life-days starting from MELD 18
print(int(V[18]))
```

This code calculates the optimal expected life-days for the patient starting from MELD score 18 using dynamic programming.
\end{dplmoutput}
\textbf{Analysis:}
\

This problem is challenging not only because of action-dependent transitions, but also because the problem description is not explicit about the horizon. The model must infer that it is an infinite-horizon problem.

$\DPLM$ handles the structure almost correctly. Its issue only lies in the reasoning section, where it mistakenly interprets the problem as finite-horizon and states that it will use backward induction. However, in the actual implementation, it applies value iteration and produces the correct solution.

In contrast, GPT-4o makes more substantial errors. It neither uses value iteration nor backward induction, and instead traverses the state space only once, which is insufficient for solving an infinite-horizon dynamic programming problem.

\section{Full Text of Prompts}
\label{appendix_prompts}
\subsection{RAG-Based Generator}
\begin{promptbox}{promptLabelAssign}
You are an expert in Dynamic Programming, and your task is to assign appropriate labels to the given DP problem. Before labeling, first develop a clear understanding of how to model the problem so that you can verify if the current problem statement is a valid dp problem with nessecary element and know how to sort this problem. Think Step by Step.
**First you need to verify if the current problem is valid.**

A problem is considered a valid DP problem if it satisfies the following conditions:

1. **State Representation**: The problem must have a well-defined state representation that captures all relevant information for decision-making.

2. **Decision Variables (Actions)**: There should be a set of actions or decisions at each stage that influence the state transitions.

3. **State Transition Function**: The problem must define how the state evolves based on decisions taken.

4. **Objective Function**: There must be a clear objective to optimize (e.g., minimize cost, maximize reward).

5. **Optimal Substructure**: The problem should exhibit optimal substructure, meaning the optimal solution to a subproblem contributes to the optimal solution of the full problem.

6. **Sequential Decision Process**: The problem should involve a sequence of decisions where the outcome of each decision affects future decisions.

If the problem meets all of the above criteria, it is a **valid DP problem**. Otherwise, it is **invalid**, and you should provide a reason why.

If the problem is invalid, output label Invalid. Otherwise, you can continue and choose labels for the valid problem.

Here is the set of labels you can choose from:

	1.	Select exactly one option from the first group. You must choose one and only one:

	•	“Finite”

	•	“Infinite-average”

	•	“Infinite-discounted”

	2.	For the other groups, you can choose at most one label from each group:

	•	[“Deterministic”]

	•	[“Optimal Stopping Problem”]

	•	[“Stage not Indexed by Time”]

	•	[“Time-Dependent State Space”]

	•	[“Continuous or Non-Integer State Variables”]

	•	[“Unbounded State Space (Requires Truncation)”]

	•	[“Action-Dependent Transition Probability”]

Here are brief introduction of some of the label above:

    •   If next states are not deterministic and depend on random disturbances, the “Deterministic” label should not be selected.

	•	Select "Stage not Indexed by Time" if stages are based on problem-specific elements (e.g., resource allocation, spatial zones, or task dependencies), not follow chronological time.

**The Dynamic Programming Problem**

\texttt{<insert problem>}

Output Format:

Provide the selected labels in the following format:

**Labels**: ["label1", "label2", ...]

Make sure your output follows the format above.
\end{promptbox}
\begin{promptbox}{\texttt{<role-based instruction>} for \texttt{Expert in Dynamic Programming}}
    \textbf{promptCoT:} now you need to construct a Chain of Thought (CoT) of a DP problem that emulates human logical reasoning, making it easier for students and AI tools to understand and apply the problem-solving process.

    \noindent\textbf{promptM:}
you are required to model a given dynamic programming (DP) problem to aid students in recognizing critical elements when solving DP problems. Use the specified format below to ensure clarity and consistency.

    \noindent\textbf{promptC:}
    your task is to generate Python code to solve the given DP problem efficiently and effectively.
\end{promptbox}

\begin{promptbox}{\texttt{<role-based instruction>} for \texttt{Operation Research Professors}}
    \textbf{promptCoT:} to help students solve Dynamic Programming problems in a logical and structured way, your task is to generate a **Chain of Thought (CoT)** of a DP problem that emulates human logical reasoning.

    \noindent\textbf{promptM:}
your task is to model the given dynamic programming (DP) problem in a way that helps students understand its critical components. Use the specified format to ensure pedagogical clarity and accessibility.

    \noindent\textbf{promptC:}
    your task is to generate Python code to solve the given DP problem efficiently and effectively.
\end{promptbox}

\begin{promptbox}{\texttt{<role-based instruction>} for \texttt{Operations Research Student}}
    \textbf{promptCoT:} ince DP problems often have similar structures, your goal is to break down any new problem into a clear and logical Chain of Thought (CoT). This step-by-step approach helps simplify the problem, identify key components, and develop effective solutions.

    \noindent\textbf{promptM:}
 your task is to break down and model a given dynamic programming (DP) problem to practice identifying critical elements. Use the specified format to ensure clarity and accuracy.

    \noindent\textbf{promptC:}
    given a dynamic programming (DP) problem as part of your homework, write Python code to solve the problem.
\end{promptbox}

\begin{promptbox}{\texttt{<role-based instruction>} for \texttt{Researcher Specializing in Decision-making under Uncertainty}}
    \textbf{promptCoT:} given a Dynamic Programming (DP) problem, your task is to generate a Chain of Thought (CoT) that highlights the logical steps needed to solve the problem, with a focus on handling stochastic elements and optimizing decision-making processes.

    \noindent\textbf{promptM:}
your task is to break down and model the given DP problem by focusing on its stochastic elements. Use the specified format to highlight how uncertainty impacts state transitions, decisions, and objectives.

    \noindent\textbf{promptC:}
    generate Python code to solve the given DP problem, focusing on handling stochastic elements effectively.
\end{promptbox}

\begin{promptbox}{\texttt{<role-based instruction>} for \texttt{Specialist in Reinforcement Learning}}
    \textbf{promptCoT:} your goal is to construct a Chain of Thought (CoT) for solving a Dynamic Programming problem, bridging the principles of DP with their applications in reinforcement learning, and providing a structured reasoning process for both students and AI models.

    \noindent\textbf{promptM:}
your task is to break down and model the given DP problem, bridging its structure with reinforcement learning principles. Use the specified format to clearly define states, actions, transition probabilities, and reward functions, and highlight their connection to RL concepts.

    \noindent\textbf{promptC:}
generate Python code to solve the given DP problem, connecting it with your expertise in reinforcement learning principles.
\end{promptbox}
\begin{promptbox}{promptCoT}
You are \texttt{<role-based agent>}, \texttt{<role-based instruction>}
To assist you, I provide the following:

1. **An instruction set** outlining the key components of CoT.
2. **Examples** of CoT construction that demonstrate how to logically connect each element for maximum clarity and coherence.

Please review the instructions and examples carefully, paying attention to how each part of the CoT builds upon the previous one to form a seamless explanation. After becoming familiar with the approach, use it to solve the new problem provided below, maintaining the same level of detail and logical progression.

**Insctruction**
A good CoT should include information about the problem's type, state, action objectives, transitions, methodology, and result, all combined smoothly into a logical chain without using bullet points, so that the final CoT forms a cohesive and continuous narrative.

1. **Problem Classification and Description**
   Describe the type of problem, including its background and time horizon (finite or infinite).For example,To solve this inventory problem with stochastic demand over the 3-period horizon

2. **State Definition**
   Describe the state variables and their meaning, including any constraints or limits. Provide a brief explanation of why this state representation is appropriate for the problem context.

3. **Action Definition**
   Define the possible actions at each state, including their values, relevance to the problem and any constraints or limits. Clarify how actions influence state transitions or outcomes.

4. **Objective**
   State the optimization goal explicitly, such as minimizing cost or maximizing reward. Specify whether the objective involves averages, totals, or expectations, and detail its components and calculation.

5. **Reward Function**
   Describe the reward or cost associated with each state-action pair. Provide details on how the reward is calculated, including any fixed or dynamic components.

6. **Transitions**
   Clearly explain the transition dynamics, describing how states evolve based on actions. Include information about probabilities, constraints, or rules governing transitions.

7. **Boundary Conditions**
   Define the start, end, or fixed values of the process:
   - **Initial State**: The starting point of the process.
   - **Terminal State**: The state where the process ends (e.g., final time step \( T \)).
   - **Boundary Values**: Fixed values at specific points, such as \( V_T(s) = 0 \) in finite horizon problems.

8. **Methodology**
   Specify the method used to solve the problem, note methods used for solving DP problems are different in different problem types, such as backward induction for finite horizons or policy iteration/value iteration/LP methods for infinite horizons. Provide a concise description of the method's steps.

9. **Result**
   Clarify the required output, such as the optimal value, optimal policy, or other measures. If deriving the optimal policy, explain how it is represented or extracted.

The template you can refer to is as follows:
To solve the [DP problem type with background and time horizon], we consider states as .... The actions are... Our objective is to .. which includes ... In each step, [how transitions are carried out]. To address this problem, we employ [xxx method]. Specifically, we ... Ultimately, we attain [certain result required in the question].

Also here are few examples for you to understand how to generate a fluent CoT:

\texttt{<insert\_examples>}

**New Problem:**

*Question:*
\texttt{<insert\_case>}

*Generate CoT:*

Following the examples above as a guide, provide a smooth and logical description of CoT for the new problem with the same brevity like the CoT in examples with is a fluent paragraph.  Remind strive for a balance between demonstrating reasoning and maintaining logical transitions, ensuring your explanation is both detailed and easy to follow.

\#\# Your output should be as follows:

\#\#\# CoT:
[Your CoT]

Output the CoT with the formulation above

\end{promptbox}
\begin{promptbox}{promptM}
You are \texttt{<role-based agent>}, \texttt{<role-based instruction>}
Also,here are some examples which is similar to the new porblem that can help you better model the problem.

\texttt{<insert\_examples>}

**YOUR QUESTION AND CHAIN OF THOUGHT:**

**Question:** \texttt{<insert\_case>}

**CoT:** \texttt{<insert\_CoT>}

**YOUR MODELING OUTPUT FORMAT:**

- **State:** [Provide your state representation in math formula, explaining the components and their significance, note the range and specific constraints of state]

- **Action:** [Detail the actions that can be taken, noting how they impact the problem, note the range and specific constraints of action]

- **Transition:** [Explain the transition mechanism, including any randomness or probabilities involved.]

- **Bellman Equation:** [Write down the Bellman equation for this problem.]

- **Boundary Condition:** [Specify the conditions that apply at the boundary or final stage of the problem.]

**Note:** Be consistent and precise in using this format to help convey your understanding of the dynamic programming problem and its solution approach.
\end{promptbox}
\begin{promptbox}{promptC}
You are \texttt{<role-based agent>}, \texttt{<role-based instrcution>}.
I'll give you the Chain of Thought (CoT) reasoning process and it's critical modeling components: State space, Action, Transition, Bellman Equation, and Boundary Condition of this problem.
When you thinking, please follow the provided CoT and modeling step by step to generate a correct code solving the complex DP problem.
I'll give you some examples to help you better solve the new problem which you can learn from.

Here are some examples similar to the new problems in its structure and solving method.

**Examples for Reference**
\texttt{<insert examples>}

**The Dynamic Programming Problem**

\texttt{<insert problem>}

**Chain of Thought** \texttt{<CoT>}

**Dynamic Programming Model:**

- **State:** \texttt{<State>}

- **Action:** \texttt{<Action>}

- **Transition:** \texttt{<Transition>}

- **Bellman Equation:** \texttt{<Bellman Equation>}

- **Boundary Condition:** \texttt{<Boundary Condition>}

Note:
- Implement the dynamic programming solution strictly following the above model.
- Ensure the code adheres to the given DP model and checks for index boundary violations. If violations are detected, modify the implementation to handle them correctly.
- The program's final output should only consist of the final numerical result.
\end{promptbox}
\begin{promptbox}{promptRefineCode}
As a dynamic programming and python coding expert, your expertise is needed to modify and refine the following  python codes to ensure solvability with  python compiler in solving a dynamic programming problem.

    This model addresses a practical application in DP. However, limitations in the original Python code prevent it from finding a solution. Please leverage the original code and your dynamic programming expertise to refine the Python implementation.
      Challenges to solvability may include:
          1. Index error that Mismatch between array index and state index
          2. Index error that Boundary issues in the state space
          3. Runtime Error that Too large state/action space
          4. Runtime Error that model the dp problem in a wrong way such that there is redundant or overly complex state/action space

The original**Problem Description:**
\texttt{<Question>}

The previous Python Code:**
\texttt{<mc\_result>}

**Error Encountered:**

After running the code, the following error occurred:
\texttt{<error>}

Please think step by step as dynamic programming and python coding expert, revise both the python scripts and the mathematical model to guarantee solvability with the python compiler,
The output should include an executable python script with solvable solution, make sure your output in the following format

\#\# Revised Solution Code:

[Your improved, executable code here]
\end{promptbox}
\subsection{Forward Generation}
\begin{promptbox}{promptForwardP}
You are an expert in Dynamic Programming (DP). Based on following industry scenario, convert a orignal DP problem into a new one.
I'll give you some example DP problems, your question should match the complexity of the reference example.

 Note:
- Ensure there is only one final question.
- You must provide a final question that can be answered with a single numerical result. Eg: find the optimal value for a specific initial state; the average optimal objective over the entire process;  the optimal action for a given initial state (must be numeric, such as how many items to purchase).
- All variables in the problem must have explicit numerical values as examples. All variables must have explicit numerical values so that students do not need to make any assumptions. The problem should be understandable and solvable for an average university student.
- Make sure you problem is based on the given new scenario and converted from original problem. Do not create scenario by yourself.
- The converted states and objectives must align with the provided industry scenario.
- The states and objectives in the original case may not directly apply to the new scenario. Refer to the distinct characteristics of the new category to ensure alignment and consistency.
- The example provided within the category can serve as a useful reference, but you should not completely follow it.
Here are the Industry Scenario, original problem and examples:

\#\#\# Industry Scenario and Introduction
\texttt{<category>}

\#\#\# Problem Requiring New Scenario
\texttt{<insert\_case>}

\#\#\# DP examples
\texttt{<insert\_few\_shot\_examples>}

\#\#\# Your output should be as follows:

\#\# question:
[Your question]
\end{promptbox}
\subsection{Backward Generation}
\begin{promptbox}{Pertub}
You are an expert in Dynamic Programming. I will provide a code that solves a specific DP problem. Your task is to modify the code by replacing the given parameter values with new, reasonable values.

**The Dynamic Programming Code**
\texttt{<insert code>}

Note:
- Do not change anything in the code except the parameter values.
- Implicit values (e.g., transition probabilities) must also be modified. Some values are not explicitly defined as parameters but can be inferred from the code.
For example, in:

max(dp[year][capital], 0.3 * dp[year + 1][new\_capital] + 0.7 * dp[year + 1][capital - invest])

The transition probabilities 0.3 and 0.7 are inferred and should be modified.
You should change all values wherever possible.

- The new values must be reasonable—they should be based on the original values but cannot be exactly the same. Here are specific guidelines to determine reasonableness:
	1.	Probabilities must be valid: Transition probabilities starting from one state should sum to 1.
	2.	Budget constraints should be logical: The total budget should be greater than the cost of a single item to ensure feasibility.
	3.	State and action constraints must make sense: If a state variable represents a quantity (e.g., inventory, capital, or energy level), it should remain within realistic bounds (e.g., non-negative values).
    4.  Any multiplier applied to the future value term in the Bellman equation (e.g., $cost = g(x, u) + \gamma * J[k + 1][x']$) should lie within the interval **(0, 1]**, as it represents a **discount factor**.
    5.  If the original code uses discrete increments for state or action variables (e.g., in steps of 10), you must ensure that all related components—including the state space, action space, and disturbances/external inputs (such as demand or supply)—use the same increment to avoid invalid index during transition. In converse, if you change the increment of the disturbances/external inputs, you must also ensure that the increments of the state and action spaces remain compatible when original space is not continuous integer set.

Output Format:

- The program's final output should only consist of the final numerical result.

As a dynamic programming and Python coding expert, your expertise is required to refine and correct the following Python code to ensure it can be successfully executed by the Python compiler and produce a valid solution to a dynamic programming problem.

The current implementation was obtained by modifying a reference solution to a specific DP problem. The intent was to replace the original parameter values with new, reasonable ones. However, limitations introduced during the modification have rendered the code unsolvable.

You are asked to identify and fix these issues by leveraging both the original version and your understanding of dynamic programming design principles.

Common challenges that may cause unsolvability:

1. **Index Errors due to Inconsistent Increments**
   In some cases, the state space, action space, and disturbances (external inputs such as demands and supply) are defined over non-continuous values and rely on fixed increments. It is critical to ensure that, during every transition step, the **next state always lies within the defined state space**.
   A simple and effective strategy is to ensure that all elements—state, action, and disturbance (such as demands and supply) —use the same increment, and all their possible values are integer multiples of that increment. This prevents invalid transitions and indexing issues.

2. **Manually Set Truncation Bounds**
   In problems with potentially unbounded state spaces (e.g., investment problems where capital grows), we often manually set an upper or lower bound to **truncate** the space for computational feasibility.
   A **reasonable bound** is one that the process cannot realistically reach within the time horizon `T`.
   When modifying the model, such bounds should be **updated consistently with the new parameters**, and must remain **integers**. Avoid arbitrarily changing constants unless they are directly tied to the structure of the state space.

3. With both truncation and increment:
   When both **increment** and **truncation bounds** are applied, the truncation effectively defines a new, bounded state space (and possibly action space).
   In this case, you must ensure that **every possible next state** resulting from any combination of (state, action, disturbance) lies **within the truncated state space** which requires the same as 1.

Use the provided original and modified code, analyze the causes of failure, and deliver an improved Python implementation that:
- Eliminates the errors introduced during modification.
- Maintains logical consistency of the DP model in the original code

**The Original Code:**
\texttt{<Code>}

**Modified Code That Needs Refinement:**
\texttt{<New\_Code>}

**Error Encountered:**
After running the modified code, the following error occurred:
\texttt{<error>}

Please think step by step as both a dynamic programming and Python coding expert. First, identify the issue caused by the previous modification. Then, revise the modified code accordingly.
Your output must include a corrected and **executable** Python script, with a valid solution and without runtime errors. Please use the following format:

```markdown
[Analysis]
<Step-by-step reasoning identifying the issue and solution>

\#\# Revised Code:
```python
\# Your corrected Python code here
\end{promptbox}
\begin{promptbox}{promptBackwardP}
You are an expert in Dynamic Programming. I will provide a code implementation that solves a specific DP problem, along with an specific industry scenario. Your task is to construct a concise, textbook-style DP problem based on that scenario, such that solving the problem yields an optimal value or action that exactly matches the output of the provided code.

Also, I’ll provide several example DP problems. Your problem must match the narrative structure, language style, and level of detail of these reference examples.

**The Dynamic Programming Code**
\texttt{<insert code>}

**The Industry Scenario**
- Scenario: \texttt{<insert\_scenario>}

- Examples: \texttt{<insert\_industry\_examples>}

- Characteristics: \texttt{<insert\_industry\_characteristics>}

**Textbook-style DP Exmaples**

\texttt{<insert\_few\_shot\_examples>}

Note:
- Your DP problem must have the exact same dynamic programming formulation as the given code. You must explicitly state all necessary information required to solve the problem in your question, including state transitions

and any probability distributions. Do not assume the reader has any predefined information.
- Do not modify any parameter values. If I solve your problem using the information you provide, the final answer must match exactly with the output of my code.

- Don't use bullet points or enumerating cases. Use concise representations (e.g., tables, matrices) to express transition probabilities, rewards, and constraints when applicable, instead of enumerating all cases.

- The final question must match the output of the code precisely. This output may be the optimal value for a given initial state or the optimal action, expressed as a numerical value or index, under specific state.

- Your problem must be grounded in the given industry scenario , but the provided examples and characteristics are only for reference. Dont's use the same detailed background as the example.

- Present the problem as a short, self-contained story. Make the numbers and constraints feel like a natural part of the setting, as in a textbook or exam question.

- No variables presetation (such as V[t][s]) and solving method (eg. backward induction loops) should appear in your generated problem directly.

- The variable names in the code may resemble a specific scenario (e.g., investment,allocation etc), but you must **ignore their semantic meaning entirely** and focus only on the code’s mathematical DP structure and state transitions.

Output Format:
Present your full problem at the end of your response with following format:

\#\# Problem:

[your full problem]
\end{promptbox}
\begin{promptbox}{promptCoTReflect}
You are dynamic programming and python coding expert. Your task is to analysis an incorrect code solving one DP problem according to correct code and provide a correct CoT for it.

To assist you, I provide the following:

1. **An instruction set** outlining the key components of CoT.

2. **Examples** of CoT construction that demonstrate how to logically connect each element for maximum clarity and coherence.

I previously asked you to solve a DP problem, but your code was incorrect. Now, I will provide you with the correct version of the code.

First, identify all the mistakes in your original code by comparing it to the correct one.
Second, write a detailed Chain of Thought (CoT) reasoning process. At the beginning of the CoT (Step 0), explicitly state the mistakes you made, explain why they occurred, and describe how to avoid them in the future. Then, proceed with a complete and well-structured CoT for solving the problem correctly.

If no code comparison is provided, you should skip step 0 and still write a full CoT based on your own reasoning process.

**Previous Code**

\texttt{<insert\_previous\_code>}

**Correct Code**

\texttt{<insert\_correct\_code>}

Please review the instructions and examples carefully, paying attention to how each part of the CoT builds upon the previous one to form a seamless explanation. After becoming familiar with the approach, use it to solve the new problem provided below, maintaining the same level of detail and logical progression.

**Insctruction**
A good CoT should include information about the problem's type, state, action objectives, transitions, methodology, and result, all combined smoothly into a logical chain without using bullet points, so that the final CoT forms a cohesive and continuous narrative.

0. **Previous Mistakes and Why they were made**

    Clearly identify errors made in your earlier solution. Explain not just what the mistakes were, but why they happened (e.g., misunderstanding of state transitions, incorrect base case, etc.). Then reflect on how to avoid such errors in the future. Be honest and specific in this step.

1. **Problem Classification and Description**

   Describe the type of problem, including its background and time horizon (finite or infinite).For example,To solve this inventory problem with stochastic demand over the 3-period horizon

2. **State Definition**

   Describe the state variables and their meaning, including any constraints or limits. Provide a brief explanation of why this state representation is appropriate for the problem context.

3. **Action Definition**

   Define the possible actions at each state, including their values, relevance to the problem and any constraints or limits. Clarify how actions influence state transitions or outcomes.

4. **Objective**

   State the optimization goal explicitly, such as minimizing cost or maximizing reward. Specify whether the objective involves averages, totals, or expectations, and detail its components and calculation.

5. **Reward Function**

   Describe the reward or cost associated with each state-action pair. Provide details on how the reward is calculated, including any fixed or dynamic components.

6. **Transitions**

   Clearly explain the transition dynamics, describing how states evolve based on actions. Include information about probabilities, constraints, or rules governing transitions.

7. **Boundary Conditions**

   Define the start, end, or fixed values of the process:
   - **Initial State**: The starting point of the process.
   - **Terminal State**: The state where the process ends (e.g., final time step \( T \)).
   - **Boundary Values**: Fixed values at specific points, such as \( V_T(s) = 0 \) in finite horizon problems.

8. **Methodology**

   Specify the method used to solve the problem, note methods used for solving DP problems are different in different problem types, such as backward induction for finite horizons or policy iteration/value iteration/LP methods for infinite horizons. Provide a concise description of the method's steps.

9. **Result**

   Clarify the required output, such as the optimal value, optimal policy, or other measures. If deriving the optimal policy, explain how it is represented or extracted.

The template you can refer to is as follows:

I made the following mistakes:

1.	[Mistake 1] — this happened because [brief explanation].

2.	[Mistake 2] — this was due to [brief explanation].

(Omit the above section if no code comparison is provided.)

To solve the [DP problem type with background and time horizon], we consider states as .... The actions are... Our objective is to .. which includes ... In each step, [how transitions are carried out]. To address this problem, we employ [xxx method]. Specifically, we ... Ultimately, we attain [certain result required in the question].

Also here are few examples for you to understand how to generate a fluent CoT:

\texttt{<insert examples>}

**New Problem:**

*Question:*

\texttt{<insert\_case>}

*Generate CoT:*

Following the examples above as a guide, provide a smooth and logical description of CoT (including the mistakes you made) for the new problem with the same brevity like the CoT in examples with is a fluent paragraph.  Remind strive for a balance between demonstrating reasoning and maintaining logical transitions, ensuring your explanation is both detailed and easy to follow.

\#\# Your output should be as follows:

\#\#\# CoT:

[Your CoT]

Output the CoT with the formulation above
\end{promptbox}
\end{document}